%% file: sm_manuscript_combined_v2.tex
\documentclass[12pt]{article}
\usepackage{amsmath}
\usepackage{amsfonts}
\usepackage{changepage}

\usepackage{setspace}
\usepackage{titling}
\usepackage{hyperref}
\usepackage{lipsum}
\usepackage{graphicx}
\usepackage{subcaption}

\usepackage[labelfont=bf]{caption}
\usepackage{cleveref}
\crefformat{figure}{Figure~#2#1#3}
\crefformat{table}{Table~#2#1#3}
\Crefformat{table}{Table~#2#1#3}

\usepackage{booktabs}
\usepackage{multirow}
\usepackage{array}
\usepackage{longtable}
\usepackage{times}
\usepackage[sort]{natbib}
\usepackage{scicite}

\usepackage{xcolor}
\newenvironment{rev}{\par\bgroup\color{blue}}{\par\egroup}

\usepackage[utf8]{inputenc}
\usepackage[T1]{fontenc}
\usepackage{listings}
\usepackage{lscape}
\usepackage{rotating}
\usepackage{enumitem}
\usepackage{float}
\begin{document}

\input{Body/main_body_v2.tex}

\clearpage
\appendix
\setcounter{section}{0}
\renewcommand{\thefigure}{A\arabic{figure}}
\renewcommand{\thetable}{A\arabic{table}}
\setcounter{figure}{0}
\setcounter{table}{0}
\pagenumbering{arabic}
\setcounter{page}{1}

\pagebreak

\setcounter{figure}{0}

\renewcommand{\thefigure}{A\arabic{figure}}
\renewcommand{\thetable}{A\arabic{table}}

\input{Body/appendix_body_v2.tex}

\end{document}

%% file: Body/main_body_v2.tex
\begin{titlepage}

\title{AI-Augmented Surveys: Leveraging Large Language Models and Surveys for Opinion Prediction}
\author{Junsol Kim \\ Department of Sociology \\ University of Chicago \\ Chicago, IL \and Byungkyu Lee\thanks{Please direct your correspondence to \href{mailto:bklee@nyu.edu}{bklee@nyu.edu}.} \\ Department of Sociology \\ New York University \\ New York, NY}
\date{}
\maketitle

\begin{abstract}

Nationally representative surveys track public opinion, yet they ask only a limited set of questions each year, limiting its potential to capture historical changes. To fill this gap, we develop a large language model (LLM)-based framework for predicting missing responses in repeated cross-sectional surveys by incorporating embeddings for questions, respondents, and survey periods. We introduce two new applications of LLMs to survey research: retrodiction (predicting year-level missing opinions) and unasked opinion prediction (predicting entirely missing opinions). Using data from the 1972-2021 General Social Surveys, our LLM-based models perform strongly in retrodicting masked GSS opinions through cross-validation and public opinions measured by other organizations in years when the GSS did not ask them. These capabilities enable us to recover missing trends and pinpoint when public attitudes changed, such as the rising support for same-sex marriage. However, performance remains modest for unasked opinion prediction. We show when our models outperform established benchmarks, examine which opinions and and respondents are more predictable, and evaluate whether our approach reduces LLMs' tendency to homogenize predicted responses. Our study demonstrates that LLMs and surveys can mutually enhance each other: LLMs broaden survey potential, while surveys calibrate LLMs for simulating human opinions.

\end{abstract}

\thispagestyle{empty}

\vspace*{\fill}

\end{titlepage}

\vspace{0.5cm}

\clearpage
\setlength{\droptitle}{-5em}

\singlespacing

\noindent
Predicting opinion trends on a range of social issues, from climate change to gay marriage, is crucial for making informed decisions, tracking social changes, and understanding the dynamics of opinion formation~\citep{brooksSocialPolicyResponsiveness2006,bursteinImpactPublicOpinion2003}. Recently, numerous breakthroughs have been made to infer and predict people's opinions and preferences from their written records, such as books in the past, internet search, and public sentiments in social media ~\citep{beauchampPredictingInterpolatingStateLevel2017a,grimmerTextDataNew2022,mooreRapidlyDecliningRemarkability2019,oconnorTweetsPollsLinking2010,stephens-davidowitzEverybodyLiesBig2017}. However, using digital trace data for predicting public opinion presents a substantial challenge, as these ``proxy'' measures cannot be deemed reliable without validating them against other ``ground truth'' benchmarks, like surveys~\citep{beauchampPredictingInterpolatingStateLevel2017a,ferraroUtilityHealthData1999}. Even if digital trace data can closely track public opinion trends, its unobtrusive and anonymous nature prompts questions about its ability to truly represent the diverse voices of the population, particularly considering the skewed representation of demographic groups in digital traces~\citep{cesarePromisesPitfallsUsing2018}. 

Surveys have been an essential tool to measure and predict opinions in academic and market research for a long time. Among others, the General Social Survey (GSS)--a nationally representative survey with exceptional quality--has been widely used to track Americans' cultural beliefs and public opinion since the 1970s over a half-century~\citep{marsdenTrackingUSSocial2020}. However, one of the practical challenges in survey research is that the number of questions that can be asked in a single survey is limited due to respondent fatigue and limited resources~\citep{berinskyMeasuringPublicOpinion2017,couperNewDevelopmentsSurvey2017}. This is especially the case for repeated cross-sectional, nationally representative surveys like the GSS, where most survey items are collected only once or twice. New questions about major social shifts are likely to be added to these surveys after they have attracted considerable public interest, leading to a lag that limits the surveys' effectiveness in identifying turning points and continuities in historical social changes. While surveys can provide a representative measure of public opinion, their range in capturing the full spectrum of public views over time is limited.

How can we predict a broad spectrum of public opinion in the dynamic social world without compromising on accuracy and representativeness while simultaneously addressing the limitations of both digital trace and survey data? This paper investigates whether we can address these challenges by training a model based on large language model (LLM) features (i.e., feature-based transfer learning) to predict survey responses in nationally representative surveys. Recent studies have already suggested the possibility of using LLMs trained on massive amounts of text data on the internet to predict public opinion by leveraging the remarkable capability of LLMs in mimicking human responses~\citep{aherUsingLargeLanguage2023,argyleOutOneMany2023,chuLanguageModelsTrained2023,dillionCanAILanguage2023,hamalainenEvaluatingLargeLanguage2023,hortonLargeLanguageModels2023,jiangCommunityLMProbingPartisan2022,metafundamentalairesearchdiplomacyteamfairHumanlevelPlayGame2022,schramowskiLargePretrainedLanguage2022,mei2024turing, park2024generativea}. However, the assumption behind these attempts is that LLMs can directly simulate the population-level \textit{average} responses to a survey question, presuming the high levels of alignment of LLMs with the specific population. This assumption has recently been challenged~\citep{santurkarWhoseOpinionsLanguage2023, bisbee2024synthetic}, which is not surprising given well-documented biases and the skewed representation of demographic groups in the digital trace data used to train LLMs~\citep{cesarePromisesPitfallsUsing2018}. Instead of building a one-size-fits-all LLM predicting the typical opinion, we propose a novel method to \textit{personalize} and \textit{contextualize} existing LLMs to be aligned to individual-specific, heterogeneous beliefs based on their responses to other questions in specific periods. Our study demonstrates that training a model based on LLM features with surveys can augment the capacity of surveys to predict public opinion by accurately predicting each survey participant's answers to new questions over long periods of time in nationally representative surveys.

\section*{Imputation, Retrodiction, and Unasked Opinion Prediction}

This paper presents three key challenges in predicting survey responses while emphasizing the opportunities that emerge from addressing each of these challenges. First, \cref{fig:figure1}, Panel A illustrates a common situation in surveys where some respondents would fail to answer or skip specific questions due to refusal and attrition. Although this task has been thoroughly investigated by traditional missing data imputation models~\citep{buurenMiceMultivariateImputation2011,honakerAmeliaIIProgram2011,rubinInferenceMissingData1976}, popular multiple imputation techniques, including Amelia and MICE, do not perform well in cases of imputing responses in sparse data sets~\citep{senguptaSparseDataReconstruction2023}. Increasingly sparse survey data due to high attrition (e.g., online surveys) or complex designs (e.g., split-ballot design) may be tackled more effectively with an accurate prediction of missing responses. We use the term ``missing data imputation'' to refer to predicting response-level missing data.

\input{Figures/figure1_three_problems_block.tex}

Panel B presents a scenario that arises in repeated cross-sectional surveys, where certain questions were not asked in some periods, resulting in year-level missing data. By predicting responses for the missing years, we can ``retrodict'' trends and patterns that would have emerged if the data had been collected consistently every year. For example, the question of whether same-sex couples have the right to marry one another has been asked since 2008 in the GSS. How do we know Americans' public opinion on same-sex marriage in the 1970s? Can we precisely pinpoint when public attitudes toward same-sex marriage started to shift? Developing a device to retrodict missing responses can open an entirely new opportunity for understanding historical changes, given that survey questions addressing specific issues tend to be introduced after societies become aware of social changes concerning those issues~\citep{behrTelevisionNewsRealWorld1985,downsEcologyIssueattentionCycle1972,hilgartnerRiseFallSocial1988}. Additionally, survey designers can utilize this device to choose which questions they ask in a survey since it enables them to focus on less predictable questions or conversely those expected to shift. We use the term ``retrodiction'' to refer to predicting year-level missing data.

Panel C presents a scenario where the goal is to predict individuals' responses to a question that has never been asked in the existing survey data. This unasked opinion prediction task with LLMs has been proposed by recent studies, motivated by their abilities to understand questions and generate human-like responses~\citep{argyleOutOneMany2023,chuLanguageModelsTrained2023,jiangCommunityLMProbingPartisan2022,santurkarWhoseOpinionsLanguage2023}. Considering the limited number of questions that can be practically included in a survey, developing a device that predicts unasked personal opinions will offer unprecedented opportunities for social science communities, businesses, organizations, and policy-makers. For example, researchers could use this device to predict people's preferences over beverages that have never been measured in the existing survey data~\citep{dellapostaWhyLiberalsDrink2015,brand2023using}. Or, it could allow researchers to study people's opinions on sensitive issues without directly asking them, particularly when asking sensitive questions may affect response quality and non-response error~\citep{yanConsequencesAskingSensitive2021}. Thus, achieving high accuracy in this task suggests the potential to infinitely expand the number of variables we can predict, unlocking unparalleled opportunities. We use the term ``unasked opinion prediction'' to refer to predicting responses to a question without any prior survey responses about the question in the training data.

\section*{Frontiers and Challenges in Opinion Prediction with Machine Learning and LLM}

Social science research has recently developed and employed machine learning techniques to predict individual attitudes and behaviors by capturing these underlying correlations among beliefs and opinions in survey data~\citep{senguptaSparseDataReconstruction2023,salganikMeasuringPredictabilityLife2020,molina2019machine,lall2022midas}. However, these techniques depend on a critical assumption for accurately predicting public opinion. That is, survey data contain all essential variables necessary for predicting missing responses to a particular question. However, due to the limited number of questions that can practically be included in a survey, it is not always feasible to ask every other questions needed to recover missing responses on a specific question~\citep{lundberg2024origins}. For example, we know that it is crucial to know respondents' LGBT status for predicting their support for same-sex marriage, but the GSS started to ask this question only after 2008. Without such key features, models might struggle to predict support for same-sex marriage accurately. Even with a comprehensive set of variables, it is very hard to predict responses that have been entirely missing for a year or those that have not been asked before (see Panels B and C in \cref{fig:figure1}).

On the other hand, recent studies suggest that LLMs could serve as a next-generation solution to some of the challenges faced by survey research, given their remarkable ability to generate responses that mimic human-like language, thoughts, and behaviors~\citep{argyleOutOneMany2023,santurkarWhoseOpinionsLanguage2023,schramowskiLargePretrainedLanguage2022, jansen_employing_2023}. Given that LLMs are trained by a wide array of text data, including data with Q\&A (questions \& answers) formats,
it makes sense to use LLMs to predict the next token (i.e., answers) to the prompt (i.e., questions). For example, LLMs can choose the answer to the question, ``Do you agree with legalizing same-sex marriage?'' in a way that their responses would be well aligned with human responses. Moreover, LLMs can encode a diverse set of information from extensive corpora generated by a diverse set of people across different periods, which enable them to learn the changing meaning of words and culture over time~\citep{farrell2025largeb}. 

Currently, a dominant approach to employing LLMs for predicting opinions in survey data is a prompt-based approach, initially proposed by Argyle and colleagues\citep{argyleOutOneMany2023}. In this approach, researchers curate prompts to contain an individual's demographic information, such as age, race, gender, political attitude, party affiliation, income, and education level and use them to simulate a persona before asking survey questions. It is akin to making LLMs ``role-play" as a survey participant with a specific set of demographic attributes and respond to a survey question~\citep{wang2023rolellm}. Emerging research shows that prompt-based LLMs can simulate survey responses and realistic personas, including ANES-based voting behaviors and social media agents used to study how news feed algorithms shape content quality \citep{argyleOutOneMany2023,tornberg2023simulating}.

However, recent work suggests that prompt-based LLMs may fail to accurately simulate personal and public opinion. First, even after prompting with demographic information, studies found that vanilla LLMs have limitations in accurately and equally representing subpopulations~\citep{santurkarWhoseOpinionsLanguage2023,abidPersistentAntiMuslimBias2021}. While conditioning LLMs with empirical data generally enhanced overall performance, the performance of in-context learning varies across different demographic groups~\citep{simmons2024assessing}. Second, LLMs may simulate a typical opinion in a particular demographic group, but it is difficult to predict a specific individual's complex and nuanced opinions. As a result, LLMs tend to generate a much narrower distribution of responses, particularly within groups, compared to those in the real surveys~\citep{argyleOutOneMany2023,gordonJuryLearningIntegrating2022,kirkPersonalisationBoundsRisk2023, bisbee2024synthetic, zhang2025generative}. 

Third, there have been concerns about the composition of demographic groups that contribute to the pre-training data of LLMs. Since the majority of the pre-training data comes from the internet, it is possible that LLMs would learn biased stereotypes against a particular demographic group~\citep{simmons2024assessing}. While it is true that these stereotypes may reflect real-world opinions, the risk lies in the potential for LLMs to over-represent or amplify such biases due to imbalances or over-reliance on certain data sources. This can result in unfair or inaccurate predictions that disproportionately affect marginalized groups. Also, as more recent text data are part of the pre-training data, LLMs' predictions could be less accurate for the past opinions~\citep{longpre2023pretrainer,gonzalez-gallardoYesCanChatGPT2023}. Fourth, LLMs’ responses could be affected by ordering and labeling biases, which may hamper the reliability of LLMs in consistently predicting the opinion~\citep{dominguez2023questioning, bisbee2024synthetic}. In sum, although the data used to train LLMs come from diverse individuals and periods, which gives them the potential to simulate public opinions from diverse populations over time, their response patterns would not reliably represent any specific individual from any particular period in the data. This limitation reduces their practical significance~\citep{santurkarWhoseOpinionsLanguage2023}. 

 \section*{Model Architectures Based on LLMs and Nationally Representative Surveys}
 
Against this background, we propose a new methodological framework by training a model based on LLM features to predict individuals' survey responses using the General Social Survey (GSS), a nationally representative survey of Americans' opinions since 1972~\citep{davernGeneralSocialSurveys2021,marsdenTrackingUSSocial2020}.
Our framework addresses two key limitations in existing approaches. First, it overcomes the shortcomings of traditional statistical methods that often treat variables as abstract entities without considering their semantic meaning. Second, it improves upon current LLMs by explicitly accounting for individual and temporal heterogeneities, which are typically not built into their core architecture. Specifically, we exploit the overlooked fact that surveys collect data through a series of texts with the same Q\&A format that can be used to train LLM feature-based models.
By incorporating the content of survey questions during the training process, our models can capture the textual nuances of survey questions and infer how individuals interpret the meaning of questions differently across time based on their response patterns. Our models assume that the meaning of survey questions and the correlations among variables for these questions can vary across individuals and change over time. This enables us to address missing data problems, as shown in Panels B and C of \cref{fig:figure1}, more effectively.
\input{Figures/figure2_architecture_block.tex}
We present an overview of our approach to personalizing LLM-based models to predict public opinion in \cref{fig:figure2}. Our approach first predicts individuals' opinions and then aggregates them at the population level using survey weights to account for sample selection bias (Panel A). Assuming accurate prediction of opinions across individuals and accounting for sample selection bias through survey weighting, our models can make predictions deemed representative of the population. However, the standard architecture of LLMs does not account for individual variability in responses, making it challenging to personalize predictions that are suited to specific individuals' distinctive beliefs and opinions in a specific period. Therefore, we need to customize the architecture of LLMs to make personalized predictions over time. 

We incorporate the three most important neural embeddings 
for predicting opinions -- survey question embedding, respondent embedding, and period embedding -- that capture latent characteristics of survey questions, individual heterogeneities, and survey periods, respectively (Panel B). Similar to word embeddings that position similar words close together, these neural embeddings represent similarities in the meanings of survey questions, individual beliefs, and temporal contexts in high-dimensional vector spaces. Finally, our models use these latent features to predict the most plausible answer to a specific question for each individual at a given moment. This novel architecture enables our models to recognize that survey responses to the same question can vary across individuals and over time.

Our setup follows the \emph{feature-based transfer learning} paradigm: the LLM backbone (e.g., self-attention layers) is held frozen, and task-specific modules built on top of the LLM backbone are trained to predict GSS responses, allowing the LLM-based model to adapt to a specific task like opinion prediction without substantially disturbing its general knowledge or incurring large computational costs. \input{Snippets/snippet_param_efficient_def.tex} Literature on ULMFiT~\citep{howard2018ulmfit} and BERT~\citep{devlin2019bert} broadly defined such methods as fine-tuning: adapting pretrained language models to downstream tasks by training task-specific parameters while keeping the backbone LLM frozen. Subsequent approaches, including adapter tuning~\citep{houlsby2019adapter}, prompt tuning~\citep{lester2021prompt}, LoRA~\citep{hu2022lora}, and BitFit~\citep{benzaken2022bitfit}, likewise freeze the backbone LLM and train only a small auxiliary module attached to the model. This differs from full fine-tuning, in which the backbone LLM weights are also updated. This approach reduces training costs, mitigates overfitting on the relatively small GSS corpus, and preserves the pretrained model’s broader linguistic and factual knowledge, which is especially important for unasked opinion prediction.

\section*{Data and Methods}

\subsection*{Data}

In this study, we use the GSS, a nationally representative survey in the United States, for fine-tuning LLMs on surveys, aiming to predict how a given individual would respond to a specific question in a given time period. 
We retrieve the text content of GSS survey questions from GSS data explorer. \footnote{\url{https://gssdataexplorer.norc.org/variables/vfilter}}.

Survey responses are commonly recorded in various formats, ranging from binary options (e.g., yes/no, agree/disagree) to ordinal Likert scales (e.g., strongly agree, agree, neutral, disagree, strongly disagree). In this study, we use binary and binarized responses to predict which side the respondent supports (i.e., agree vs. disagree) rather than the intensity (i.e., strongly agree vs. agree). We dichotomize ordinal responses by assigning a value of 1 to positive responses (e.g., agree, yes, true, likely) and 0 to negative responses (e.g., disagree, no, false, unlikely). See \cref{tab:binary_transformation} for the top 50 response options. For instance, positive responses such as ``strongly agree'' and ``agree'' were coded as 1, while negative responses such as ``strongly disagree'' and ``disagree'' were coded as 0.

Among all $7{,}136$ variables retrieved from the GSS Data Explorer API, we keep discrete categorical variables with two or more valid response labels. We then annotate each variable with a question type and whether they can be binarized using Gemini-2.5-Pro (temperature $= 0.2$); the full annotation prompts are shown in \hyperref[appendix:raw_prompts]{Appendix~A}. Specifically, Gemini-2.5-Pro assigns each variable to one of nine question types: Behavioral, Attitudinal/Opinion, Demographic, Objective/Knowledge/Factual, Questions about Other People, Open-ended, Derived/Computed/Condition, Metadata/Paradata, and Miscellaneous.

We then keep only three substantive question types relevant to opinion and behavior prediction---Attitudinal/Opinion ($68.15\%$), Behavioral ($26.85\%$), and Objective/Knowledge/Factual ($5.00\%$); see \cref{tab:question_distribution}---that are also flagged as binarizable. For each variable flagged as binarizable, Gemini-2.5-Pro then produces the actual response-option binarization, mapping every valid response code to 1 (positive) or 0 (negative) (see the binarization prompt in \hyperref[appendix:binarization_prompt]{Appendix~A.1}). Finally, the authors validate and correct the LLM-produced category and binarization labels. \cref{fig:figure_a_variable_selection} summarizes the variable selection flowchart. Finally, the dataset consists of $68{,}846$ individuals' responses to $3{,}699$ binarized questions collected over $33$ repeated cross-sectional waves between $1972$ and $2021$.

\subsection*{Overview of the Model Training Process}
In this section, we provide an overview of the training process used to build a model for predicting GSS responses. We employ a \emph{feature-based transfer learning} approach: the LLM backbone is held frozen, and only the task-specific modules attached on top of the backbone are updated on the GSS data. Panel B of \cref{fig:figure2} summarizes the nature and architecture of three embeddings for survey questions, respondents, and time periods. These embeddings are designed to capture the semantic structure of questions, the latent belief systems of respondents, and the temporal dynamics of public opinion, respectively (see \cref{fig:figure_a_architecture_detail} and \hyperref[appendix:architecture_detail]{Appendix~D} for full architectural detail).

\textit{Question embedding.} To obtain the question embedding, we begin with sentence-level embeddings from LLMs pre-trained on vast text corpora; these enable us to encode the meaning of survey questions, such as ``Do you agree or disagree that homosexual couples have the right to marry one another?'' and map these questions into a latent vector space~\citep{jurafskySpeechLanguageProcessing2023, taorirohanAlpacaStrongReplicable2023}. We pass each survey question through Alpaca-7B (frozen) and project the last-token sentence-level vector through a small trainable feed-forward layer to obtain the question embedding \(s \in \mathbb{R}^{n}\). Examples of free-text Alpaca-7B responses to GSS questions are reported in \cref{tab:alpaca_text_examples}, and the most-similar GSS items in the embedding space are listed in \cref{tab:similar_questions}.

\textit{Respondent embedding.} A breakthrough we made for personalizing LLMs is to incorporate respondent embeddings to account for individuals' heterogeneous responses to survey questions based on heterogeneous belief systems~\citep{baldassarriNeitherIdeologuesAgnostics2014,milbauerAligningMultidimensionalWorldviews2021}. Recent approaches like in-context tuning utilize various prompts to steer LLMs to represent group-level opinions~\citep{brownLanguageModelsAre2020}, but they fail to account for unique individual perspectives because these methods can only predict the commonly held opinions of a ``typical'' group~\citep{argyleOutOneMany2023,santurkarWhoseOpinionsLanguage2023}. We refer to this vector, \(b \in \mathbb{R}^{n}\), as the ``respondent embedding'' of individuals. Unlike the question embedding, respondent embeddings are not derived from Alpaca-7B; instead, we initially assign random latent features for each individual and optimize them during the training process on GSS data, such that two individuals closely located in this embedding space are either more likely to have a similar set of beliefs or to have similar latent correlational structure of beliefs~\citep{gordonJuryLearningIntegrating2022}.

\textit{Period embedding.} We incorporate period embeddings to consider temporal changes in the meaning of questions and individuals' belief systems~\citep{jooOutSyncOut2020,ruleLexicalShiftsSubstantive2015} as the third layer of neural embeddings. This accounts for the impact of temporal factors on survey responses, such as the gradual shift towards more progressive beliefs over time~\citep{baldassarriWasThereCulture2020} and the effects of specific events during a particular period, such as macroeconomic changes, presidential elections, and the COVID-19 pandemic. We represent the historical features of survey periods as an $n$-dimensional embedding vector \(p \in \mathbb{R}^{n}\), which we refer to as the ``period embedding'' of surveys. Like the respondent embedding, the period embedding is not derived from Alpaca-7B; instead, we initially assign random latent features for each period, which are optimized on GSS data such that two adjacent periods characterized by similar response patterns are located close to each other in the latent space during the training process. A $t$-SNE visualization of all three learned embedding spaces is shown in \cref{fig:figure3}.

\textit{Higher-order interactions through Deep Cross Networks.} The three embeddings are concatenated into $x_0 \in \mathbb{R}^{3n}$ and passed through a stack of \emph{cross layers}~\citep{gordonJuryLearningIntegrating2022, wangDCNV2Improved2021},
\[x_{l + 1} = x_{0}\, \odot\, (W_{l}x_{l} + b_{l}) + x_{l},\]
which capture multiplicative interactions among question, respondent, and period features (e.g., how an individual's ideology and the ongoing pandemic jointly shape responses to a vaccine-mandate question). Each cross layer adds one polynomial order of feature interaction; with $k$ cross layers, the model captures interactions up to order $k+1$. The cross-layer stack is followed by feed-forward dense layers and a sigmoid head producing $P(\text{Agree}) \in [0,1]$. 
This design has been employed by prior work that aims to model how different individuals react to the same text differently---for example, whether a particular individual perceives a given text as toxic or not---by jointly modeling text and respondent latent features~\citep{gordonJuryLearningIntegrating2022}.

To generate the predicted probability that each individual provides a positive response to a particular question, we employ a sigmoid function, which generates a probability that falls within the range of 0 to 1, as logistic regression models do. To estimate public opinion, we aggregate these probabilities with survey weights. We find that averaging the sigmoid-transformed individual predictions can lead to an underestimation or overestimation of the true population proportion, especially when the true probabilities approach extreme values near $0$ or $1$. This bias arises due to the non-linearity of the sigmoid function, defined as $sigmoid(x) = \frac{1}{1 + e^{-x}}$, and the implications of Jensen's Inequality, which asserts that $sigmoid(\mathbb{E}[X]) \neq \mathbb{E}[\sigma(X)]$. To correct this bias, we adjust the aggregated outcome by modeling the target observation using the equation $sigmoid(\beta_0 + \beta_1 \cdot \text{aggregated prediction})$. This approach accounts for the non-linear transformation at the aggregate level, ensuring that the expected average agreement is accurately estimated without the bias introduced by directly averaging individual predictions. Finally, we predict counterfactual trends using aggregated responses from retrodiction models. To estimate the direction and magnitude of over-time trends, we employ a Generalized Additive Model (GAM), a flexible, non-parametric regression technique capable of modeling complex, non-linear trends~\citep{hastieGeneralizedAdditiveModels1992}.

\subsection*{Model Training}

Our approach to personalizing LLMs is model-agnostic, meaning we can generate personalized responses using any LLM. For example, we can employ either decoder-only transformer models with billions of parameters (e.g., GPT-4, Alpaca, GPT-J) or encoder-only transformer models (e.g., BERT, RoBERTa), given that their hidden layers are accessible. Given the limited availability and reproducibility issues of private LLMs (e.g., GPT-4) despite their impressive performance~\citep{aiyappaCanWeTrust2023}, we opt for three widely-used open-source alternatives that demonstrate competitive performance in previous natural language processing benchmark tests: Alpaca-7B,
GPT-J-6B, and RoBERTa-large~\citep{liuRobertaRobustlyOptimized2019,taorirohanAlpacaStrongReplicable2023,wangbenGPTJ6BBillionParameter2021}.

All embeddings and the DCN are jointly trained on GSS data using HuggingFace and TensorFlow Recommenders, with the LLM backbone frozen, the Adam optimizer (learning rate $= 2\mathrm{e}{-5}$), and binary cross-entropy loss. Hyperparameter choices (embedding dimension 50, three cross-layers and three dense layers of size 150) and other details are presented in \hyperref[appendix:methodological_details]{Appendix~E}.

We found that using demographic features during the training process did not significantly affect the models' performance, as shown in \cref{fig:figure_a_demographic_cv}. Based on these results, we exclude socio-demographic variables in the training data but include measures of political ideology (liberal, moderate, conservative) and party affiliation (Democrat, Republican, Independent, Others) that are known to be one of the most important predictors of opinion formation.

\subsection*{Model Evaluation}

We evaluate the model's performance in predicting opinions at the individual and population levels by conducting 10-fold cross-validation and measuring their accuracy in restoring missing data. \cref{fig:cv_scheme_a} shows examples of the 10-fold cross-validation scheme. 
In each fold, the model is trained on 90\% of the data and tested on the remaining 10\%, with held-out responses excluded entirely from training. We repeat this process across all folds to assess performance across respondents and opinion items.

For the response-level missing opinion (Panel A, \cref{fig:cv_scheme_a}), we randomly assign  combinations of survey year, question, and respondent ID into ten folds. In each iteration, one fold is excluded from training, and the model predicts the held-out responses. Repeating this process across all folds yields predictions for every response in the dataset.

To evaluate \emph{retrodiction} (i.e., predicting year-level missing opinions), we use a modified 10-fold cross-validation procedure. Unlike the first setting, where questions are asked but some respondents do not answer them, this setup mimics periods in which questions are not asked at all. Specifically, we randomly remove approximately 10\% of (survey question, year) pairs from the training data (Panel B, \cref{fig:cv_scheme_a}). These pairs are assigned into ten folds, and in each iteration the model is trained on the remaining data and tasked with predicting the full set of responses for the excluded question-year pairs. This process is repeated across all ten folds.

Within the retrodiction task, each held-out (variable, year) pair falls into one of three scenarios (illustrated in \cref{fig:regime_schematic}), defined by the position of the held-out year relative to the observed years of the same variable in the training data. \emph{Backcasting} describes cases in which the held-out year comes before all observed years of that variable in the training data---for example, when the held-out year is 1990 but the variable is only observed from 2008 onward in the training set, requiring the model to predict responses backward from later observations. \emph{Forecasting} is the converse case, in which the held-out year comes after all observed years---for example, when the held-out year is 2010 but the variable is only observed up to 1990 in the training data. \emph{Interpolation} describes held-out years that fall between observed years of the same variable---for example, when the held-out year is 2002 and the variable is observed in both 2000 and 2004 in the training data. We report performance separately for each case and additionally evaluate how prediction quality changes with temporal distance from the nearest observed training year (1, 2--3, 4--5, 6--7, and 8+ years). Specifically, within each subgroup, compute the Spearman correlation between Alpaca-predicted and GSS-observed mean responses.

To evaluate ``unasked opinion prediction'' (i.e., predicting entirely missing opinions), we randomly eliminate around 10\% of survey questions entirely from the training data and use the model to predict them (see \cref{fig:cv_scheme_a} Panel C). Survey questions are randomly assigned into ten folds, and in each iteration the model predicts responses for questions it never observed during training.

\subsubsection*{Evaluation Metrics}

For personal opinion prediction we report the Area Under the ROC Curve (AUC), which captures how well the model ranks positive over negative responses without requiring an arbitrary threshold and is robust to class imbalance; Accuracy and F1 yield similar results (\hyperref[appendix:methodological_details]{Appendix~E})~\citep{salganikMeasuringPredictabilityLife2020,savcisens2024using}. For public opinion prediction we report three complementary metrics: the per-variable Pearson correlation $r_{i}$ between observed and predicted mean responses across years; the per-variable mean absolute error MAE$_{i}$; and the percent prediction accuracy at margins of error such as 3\% and 5\% (\hyperref[appendix:methodological_details]{Appendix~E}).

\subsubsection*{Performance Comparison Against Three Missing Mechanisms}

Previous research on imputing missing data has explored three different assumptions about how missing data occurs: Missing Completely at Random (MCAR), Missing at Random (MAR), and Missing Not at Random (MNAR)~\citep{senguptaSparseDataReconstruction2023, gelman2007data}. Since our evaluation process only assumes MCAR, we also simulate the missing data based on the other two assumptions, MAR and MNAR, and examine whether our models show similarly high accuracy in such missing scenarios. See \hyperref[appendix:mcar_mar_mnar]{Appendix~D} for details on simulating the three missing mechanism scenarios.

\subsubsection*{Sensitivity Analysis for Retrodiction}

In our retrodiction analyses, one potential concern is information leakage, whereby highly correlated questions in the training data inadvertently provide direct clues about the target variable, inflating model performance. We address this issue through two complementary sensitivity analyses. First, we recognize that GSS items are typically organized into thematic modules or multi-item scales measuring specific latent constructs, and survey designers tend to add or remove entire modules rather than individual items (see \cref{tab:top20_modules_updated}). For example, survey questions on climate change beliefs might be paired with questions about government environmental policies in the same module. Then, predicting a missing question may become artificially easier if other closely related questions from the same module remain in the dataset. To account for this, we conduct a sensitivity analysis by dropping entire modules instead of individual items at random. We construct a year-module matrix listing all modules included in each survey year. We then randomly remove 10\% of modules and all corresponding items, repeating this process 10 times using 10-fold cross-validation. This approach simulates a more realistic scenario where entire modules are added or removed by survey designers. Second, we investigate whether our model's predictive strength relies excessively on a small subset of highly correlated questions, potentially limiting its robustness and generalizability. If model performance substantially declines upon removing these closely correlated items, predictions may reflect narrow relationships rather than broader underlying belief systems. This scenario is especially relevant when such strongly correlated questions are unavailable in real-world survey contexts due to design constraints or incomplete data.
\subsection*{External validation using surveys from Roper Center archive}

In years for which the GSS did not field a given question, our model's predicted trajectory cannot be verified against GSS ground truth, since none exists. To address this concern, we construct an external validation pipeline using the Roper Center iPOLL archive (\url{https://ropercenter.cornell.edu/ipoll}), which provides independent observations of opinions in years for which the GSS did not field the corresponding question. Restricting the iPOLL archive to U.S.\ national-adult studies, the corpus contains $17{,}828$ unique surveys and $643{,}283$ individual questions (about $36.1$ questions per survey on average) fielded by organizations such as CBS, Gallup, Pew, the Los Angeles Times, and many others, with coverage dating back to 1935.

We followed four steps to identify Roper surveys that measured the same questions as the GSS, in the same or different years. First, we compiled a list of surveys in the Roper Center archive whose target population is U.S.\ adults. Second, to identify GSS questions appearing in those surveys, we generated text embeddings for all GSS questions and all questions from the Roper surveys (hereafter, ``Roper questions'') using EmbeddingGemma-300M, and computed pairwise cosine similarities between each GSS question and each Roper question. We retain question pairs with cosine similarity of at least $0.85$. Third, we used Gemini-2.5-Pro to verify whether each retained pair (i.e., one GSS question and one Roper question) measures the same construct---i.e., that both questions ask the same thing---rather than merely sharing a topic, and retained only those pairs that Gemini identified as measuring the same construct with confidence of at least $0.85$ (see the verification prompt in \hyperref[appendix:roper_verify_prompt]{Appendix~A.3}). Finally, we binarized the response options from the Roper surveys (hereafter, ``Roper responses'') to match the coding of the corresponding GSS variables (see the binarization prompt in \hyperref[appendix:roper_binarize_prompt]{Appendix~A.4}).%

\subsection*{Alternative Benchmarks}

We compare our method against existing benchmarks. For personal opinion prediction we use four approaches. First, we use matrix factorization (MF), a machine-learning method shown to fill in missing survey responses on par or better than traditional imputation models when the data are sparse. Second, we use text-aware matrix factorization, which adds question-text features, such as TF-IDF vectors or embedding vectors of survey questions, to MF. Third, we use Multivariate Imputation by Chained Equations (MICE), the standard imputation method in survey methodology. Finally, we use out-of-shelf LLMs (GPT-4o, Gemini-2.5-flash) prompted to predict survey responses by asking them to assume a survey participant with a particular persona, under three conditions: (i) only the respondent's demographics are given; (ii) demographics plus responses to 10 random GSS items; (iii) demographics plus responses to the 10 GSS items most correlated with the target variable. For public opinion prediction, we use four different time-series regression models, which fit each variable's mean response $y$ (e.g., $60\%$) as a function of survey year $t$ (e.g., $2010$): linear OLS, logistic GLM (with logit link), polynomial OLS (with a quadratic year term), and logistic polynomial GLM. Implementation details for all benchmarks are provided in \hyperref[appendix:alt_benchmarks]{Appendix~E}.%

\section*{Results}

\subsection*{Model Performance for Personal Opinion Prediction}

\cref{tab:modelcomparison} shows that Alpaca-7B achieves the strongest overall performance across all three tasks among the three LLMs, the MF model, and the MICE benchmark. While performance differences are relatively small for response-level missing data imputation, they widen for retrodiction and become largest for unasked opinion prediction. \cref{fig:figure5}, Panel A displays the performance of the best model (i.e., Alpaca-7B) for individual-level predictions across three tasks. 

Our top-performing model succeeds in the missing imputation task (AUC $= 0.868$), though the MF also shows a similar level of performance (AUC $= 0.856$). Given that the MF presumes that data are missing completely at random (MCAR)---a stronger assumption than the missing at random (MAR) principle that underlies standard multiple imputation frameworks---it is essential to examine how our models operate across various missing data mechanisms. Using the simulated GSS data based on three different mechanisms (MCAR, MAR, and MNAR: Missing Not At Random), \cref{fig:figure_a_missing_mech} shows that our model performs better under MCAR, MAR, and MNAR than both MF and MICE benchmarks. These findings indicate that our model performs better in inferring answers for not only randomly skipped responses, as seen in split-ballot designs, but also for non-random systematic refusal. 

Our model performs strongly in the retrodiction task (AUC $= 0.862$), where it predicts entirely missing responses for specific years, substantially outperforming both the matrix factorization (AUC $= 0.807$) and MICE (AUC $= 0.727$) benchmarks. We further decompose retrodiction into three subtasks : \textit{backcasting} (held-out year before all training years for that variable), \textit{interpolation} (between training years), and \textit{forecasting} (after all training years). At the individual level, performance is high and similar across all three: AUC is $0.842$ for backcasting, $0.868$ for interpolation, and $0.852$ for forecasting; accuracy is $0.757$, $0.782$, and $0.766$; and F1 is $0.780$, $0.784$, and $0.782$, respectively (\cref{tab:retro_interp_forecast}). \cref{tab:retro_distance_metrics} reports the full individual-level breakdown by subtask and distance to the nearest training year.

\input{Figures/figure_modelperformance_block.tex}

How can our model generate highly accurate predictions based on sparse survey data? We conduct a permutation analysis to assess the relative importance of features~\citep{wangDCNV2Improved2021}. Specifically, we randomly shuffle one embedding at a time while keeping all other embeddings unchanged and measure their respective contributions to predictive performance (i.e., AUC, Accuracy, F1-score)~\citep{breiman2001random,mi2021permutation,altmann2010permutation,kaneko2022cross,merrick2019randomized}. Here, the decline in performance after permutation reflects the importance of each embedding in predicting survey responses. \cref{tab:metrics} shows that question embeddings and respondent embeddings are the most important features. Randomly permuting the question embeddings significantly lowers the AUC for retrodiction from $0.857$ to $0.509$, bringing the model's predictive performance close to random chance. Similarly, shuffling respondent embeddings reduces the AUC from $0.857$ to $0.720$. In contrast, period embeddings were found to be less important: shuffling period embeddings led to only a minor decrease in AUC from $0.857$ to $0.849$.

In case of unasked opinion prediction, the sheer scale of LLMs, with billions of parameters, suggests that it may be possible to predict entirely missing opinions, especially given that some research indicates the feasibility of simulating individual survey responses only using LLMs~\citep{argyleOutOneMany2023, dillionCanAILanguage2023}. However, we find that the performance in this task is lower compared to the two aforementioned tasks. For instance, Alpaca-7B, with seven billion parameters, achieved an AUC of $0.701$, and GPT-J-6B, with six billion parameters, achieved an AUC of $0.661$. However, it is important to note that bigger LLMs still outperform smaller models like RoBERTa-large (AUC $= 0.573$), which is consistent with the finding that upscaled LLMs significantly enhance task-agnostic, zero-shot, or few-shot performance without fine-tuning~\citep{brownLanguageModelsAre2020}.

To better understand the large AUC gap between retrodiction and unasked opinion prediction ($0.862$ vs.\ $0.701$), we evaluate model performance under varying levels of missingness, ranging from 10\% to 90\%. \cref{fig:figure_a_missing_prop} shows that predictive performance generally declines as less training data is available, although this pattern is less consistent for unasked opinion prediction.

\subsection*{Model Performance for Public Opinion Prediction}

Predicting public opinion is a separate challenge from predicting personal opinion, as it requires accounting for varying probabilities of sample selection when we aggregate personal opinions. For instance, if the sampling weights for Black respondents are higher than those for White respondents, biases in our estimates---when weighted by sampling weights---will become greater when the predictive accuracy is lower among Black individuals. Yet, \cref{fig:figure5} Panels B-D reveal that the performances of public opinion prediction largely mirror individual-level results. Missing data imputation and retrodiction models that rely on existing human responses show very high correlations between the observed proportions of positive responses and the predicted proportions for opinions measured in each survey year ($r > 0.98$), indicating that our predictions can be reliably used for trend estimation and correlational analysis. In contrast, our model shows a relatively lower correlation ($r = 0.68$) in unasked opinion prediction. This result suggests that researchers should be cautious when using LLMs to completely replace trend estimation and correlational analysis in social science research or high-stakes decision-making.

Since correlation metrics convey the overall alignment of predictions with true values but may overstate the practical usability of methods, we also assess prediction accuracy using prediction intervals. Specifically, we analyze the percentage of observed values falling within different ranges of predicted values. \cref{tab:model_prediction_accuracy} shows that for imputation, the model demonstrates strong accuracy, with $74.0\%$ of observed values falling within $\pm 3\%$ of predicted values and $91.9\%$ within $\pm 5\%$. Retrodiction also shows robust performance, with $50.6\%$ of observed values within $\pm 3\%$ and $73.6\%$ within $\pm 5\%$. In contrast, unasked opinion predictions exhibit considerably lower accuracy. Only $17.5\%$ of observed values fall within $\pm 5\%$ of predicted values, and even at a broader $\pm 10\%$ threshold, accuracy improves only marginally to $34.3\%$. The lower performance of unasked opinion prediction models for public opinion prediction highlights how difficult it is to make a precise prediction at the aggregate levels, even with a decent size of correlations and relatively high individual-level accuracy.

Our retrodiction models perform strongly across all three subtasks at the population level: interpolation ($N=6{,}051$) achieves $\rho=.99$ ($\text{MAE}=0.034$), while backcasting ($N=1{,}401$) and forecasting ($N=1{,}406$) yield $\rho=.97$ ($\text{MAE}=0.046$) and $\rho=.97$ ($\text{MAE}=0.049$), respectively (\cref{fig:retro_interp_forecast}). One may have a concern that predictions for periods further outside the observation window are much harder than predictions a year or two away. \cref{fig:distance_nearest_year_dummy} shows that this is not the case. For held-out years \emph{one} year away from the training window, Spearman $\rho$ is $0.97$ for backcasting, $0.99$ for interpolation, and $0.98$ for forecasting. At 2--3 years away, $\rho$ is $0.97$, $0.99$, and $0.97$ respectively. At 4--5 years away, $\rho$ is $0.98$, $0.99$, and $0.98$. At 6--7 years away, $\rho$ is $0.98$, $0.97$, and $0.96$. And even at $8+$ years away from the training window, $\rho$ stays at $0.96$ for backcasting, $0.97$ for interpolation, and $0.97$ for forecasting. In other words, opinions far from the training window can still be predicted reasonably well---not quite as well as nearby years, but the performance decline is small. 

To address concerns that highly correlated training questions could leak information about held-out variables and inflate performance, we conduct two sensitivity analyses as we describe. First, when entire modules were removed, the model maintained strong correspondence between observed and predicted aggregate responses, producing a Pearson's $r = 0.937$ and MAE $= 0.053$; predicted mean responses fell within $3$ percentage points of the observed values for $52.1\%$ of (variable, year) cells and within $5$ percentage points for $72.8\%$ of cells (\cref{fig:figure_a_module_removed}). Second, we systematically removed highly correlated GSS items (\cref{tab:samesex_50}), such as \textit{homosex1} (correlation: $0.645$) and \textit{homosex} (correlation: $0.628$) for the same-sex marriage item (\textit{marhomo1}) from the training data and evaluated the model's performance. Even after removing the top 10---and even the top 50---most correlated questions, the model's AUC drops only marginally and remains well above chance (\cref{fig:figure_a_samesex_exclude}), suggesting that the model leverages broad belief-system patterns rather than narrow correlated-item shortcuts. 

\subsection*{External validation using surveys from Roper Center archive}

Since the GSS provides no gound truth for years in which a question was not fielded, we constructed an external validation pipeline using matched national-adult surveys from the Roper Center iPOLL archive (see Methods). For each matched (variable, year) pair, we compared the mean response among GSS respondents and the corresponding Roper surveys that asked the same question in the same year. The two sources are strongly correlated but not perfect due to the differences in sampling and survey design. Across $N = 579$ matched (variable, year) pairs, the two sources show strong corcordance (Spearman $\rho = 0.82$, $\text{MAE} = 0.112$). We treat this GSS--Roper concordance as the empirical ``ceiling'' for models trained on GSS data and evaluated against external Roper observations.

Remarkably, our Alpaca-7B model, trained on GSS responses, comes close to this empirical ceiling in its retrodiction task. The model predicts Roper observations for survey questions both in years where the GSS observes the same variable and in years where it does not. We split the evaluation into two cases (\cref{fig:roper_3panel}): (a) how well Alpaca-7B predicts mean responses in Roper surveys in years when the GSS does ask the same variable, and (b) how well Alpaca-7B predicts mean responses in Roper surveys in years when the GSS does \emph{not} ask the variable. Performance is strong in both settings: in case (a), we obtain Spearman $\rho = 0.81$ with $\text{MAE} = 0.113$ ($N = 579$), and in case (b), we obtain $\rho = 0.79$ with $\text{MAE} = 0.120$ ($N = 624$). In other words, even in years where the GSS provides no direct measurement of a given question, Alpaca-7B's predictions of Roper observations remain nearly as accurate as when such direct GSS measurements are available, and are close to the empirical ceiling defined by GSS--Roper real observations.

\input{Figures/figure_roper_3panel_block.tex}

\subsection*{Retrodicting Counter-factual Trends}

Building on this Roper-based external validation, we present concrete examples in which Alpaca-7B predicts opinions for years in which the GSS itself did not field the corresponding question. \cref{fig:figure6} shows the predicted and observed trajectories for four example GSS variables, chosen to span distinct sparsity patterns of measurement trends. The full list of matched Roper surveys behind these panels is provided in \cref{tab:counterfactual_roper_surveys}.

\input{Figures/figure_counterfactual_block.tex}

As shown in Panel A, the question \texttt{marhomo1} (``Homosexual couples should have the right to marry one another''---Strongly agree/Agree vs.\ Disagree/Strongly disagree) is observed in the GSS only in seven biennial waves (2006--2018), so the entire 1972--2005 window---more than three decades---is missing from the GSS and must be backcast. The Roper iPOLL archive provides U.S.\ national-adult observations across 1992--2015 ($N=16$ surveys), most of which fall in years where the GSS does \emph{not} field the question. Alpaca-7B's predicted backcasted trajectory closely tracks the matched Roper observations across this period, picking up the gradual rise in support for same-sex marriage that the contemporaneous Roper polls also document. Because the GSS has also asked a similarly worded version of this question that includes the word ``should,'' we checked how the predicted response shifts with wording: the ``should'' variant yields a slightly lower predicted level of agreement than the original (\cref{tab:samesex_framing}).

As shown in Panel B, the question \texttt{busing} (``favor the busing of Black and white school children'') is the inverse case---a forecasting setting. The GSS fielded the question across 17 waves between 1972 and 1996 and then dropped it for nearly three decades. In 2019, a Gallup survey (archived in Roper iPOLL) re-fielded a closely matched busing question to assess how attitudes had evolved over that period. Alpaca-7B's predicted 2019 response lines up closely with the Gallup observation (orange diamond in Panel B), suggesting that public opinion on busing slightly increased after the GSS stopped asking the question.

As shown in Panels C and D, the questions \texttt{concong} (“how much confidence do you have in the U.S.\ Congress”---A great deal vs.\ Only some/Hardly any) and \texttt{cohabit} (“lived with spouse before marriage”---Yes vs.\ No) illustrate cases where the GSS contains only a small number of observed years (\texttt{concong} in 1991, 1998, 2008, 2018; \texttt{cohabit} in 1988 and 1994 alone). Across both panels, Alpaca-7B's predicted trajectories approximately track the matched Roper observations: in Panel C the smoothed Alpaca curve captures the broad decline in congressional confidence visible in the observed mean responses in Roper surveys across 1992--2019, and in Panel D it approximates the gradual rise in pre-marital cohabitation reported in the 1995--2002 Roper polls. These cases demonstrate that the model can produce plausible counterfactual trajectories even for variables with very sparse GSS coverage.

\subsection*{Benchmarks Against Alternative Methods}

We benchmark our Alpaca-7B model against the five alternative methods introduced in Methods (Alternative Benchmarks). The first four---matrix factorization (MF), text-aware matrix factorization (MF~+~TF-IDF, MF~+~SentenceBERT), Multivariate Imputation by Chained Equations (MICE), and prompted out-of-shelf LLMs (GPT-4o, Gemini-2.5-flash)---are evaluated on personal opinion prediction; the last, per-variable time-series regression in four variants (linear OLS, logistic GLM, polynomial OLS, logistic polynomial GLM), is evaluated on public opinion prediction.

\textit{Personal opinion prediction.} On the retrodiction task, our Alpaca-7B model (AUC $=0.862$) substantially outperforms MF ($0.807$) and MICE ($0.727$). Adding question text to MF does not eliminate the gap: two text-aware MF variants---MF~+~TF-IDF and MF~+~SentenceBERT (\citet{singh2008collective}; \cref{tab:hybrid_mf})---achieve AUCs of $0.853$ and $0.854$ for missing data imputation and $0.723$ for retrodiction in both cases. These models perform slightly worse than the original text-blind MF ($0.856$ for imputation and $0.728$ for retrodiction), potentially because they lack model architectures that effectively process question text, such as the self-attention layers in LLMs. Both text-aware MF variants also perform substantially worse than Alpaca-7B ($0.868$ for imputation and $0.829$ for retrodiction).

We also benchmark our Alpaca-7B-based model against prompted out-of-shelf LLMs (GPT-4o and Gemini-2.5-flash) on the 2018 GSS data under three prompting strategies: demographics-only, random-$k{=}10$ context, and correlation-top-$k{=}10$ context (\hyperref[appendix:alt_benchmarks]{Appendix~E}; \cref{tab:frontier_llm}). For missing data imputation, our model (Acc~$=0.787$, F1~$=0.813$) outperforms the best out-of-shelf LLM configuration, GPT-4o with correlation-top-$k{=}10$ (Acc~$=0.703$, F1~$=0.714$), as well as Gemini-2.5-flash with the same prompt (Acc~$=0.701$, F1~$=0.712$). For retrodiction, our model (Acc~$=0.723$, F1~$=0.764$) again surpasses both GPT4o and Gemini-2.5-flash. On unasked opinion prediction, the two approaches are roughly comparable: Alpaca-7B reaches Acc~$=0.672$, F1~$=0.725$, whereas GPT-4o with correlation-top-$k{=}10$ reaches Acc~$=0.703$, F1~$=0.714$, giving the out-of-shelf LLM a slight advantage in accuracy and our model a slight advantage in F1. Taken together, these results show that our method outperforms prompted out-of-shelf LLMs on missing data imputation and retrodiction, and is on par on unasked opinion prediction, despite using far fewer parameters (7B vs.\ hundreds of billions).

\textit{Public opinion prediction.} For public opinion prediction, we compare against two families of population-level baselines (Methods, Alternative Benchmarks): matrix factorization (MF), and per-variable time-series regression in four variants---linear OLS, logistic GLM, polynomial OLS, and logistic polynomial GLM---which fit each variable's observed mean response. We compare these baselines across three dimensions. First, we assess performance under varying data sparsity, defined by how often a question appears in the training data (once, twice, or more than twice). This is critical, as most GSS questions appear only once (60.0\%) or twice (17.2\%), making accurate prediction under sparse conditions especially important. Second, we examine temporal distance, comparing predictions made for years $\geq 3$ versus $<3$ years from observed data, to test the model's ability to extrapolate across time. Finally, we evaluate performance under volatility, distinguishing between high- and low-volatility settings based on whether the year-to-year variance in responses exceeds the mean---capturing how well the model handles stable versus rapidly shifting opinion trends.

Pooled across all (variable, year) pairs, our Alpaca-7B model achieves the strongest performance among models that work across the full set of GSS variables (Spearman $\rho = 0.982$, MAE $= 0.038$), substantially better than the matrix-factorization baseline ($\rho = 0.831$, MAE $= 0.108$) and comparable to the strongest per-variable regression baseline (Logistic OLS: $\rho = 0.984$, MAE $= 0.030$). The full breakdown across all subgroups and baselines is reported in \cref{tab:perfbygroup}. 

The advantage of our method is most pronounced in the hard cases. In highly sparse cases where a question appears only once in the training data ($\text{year} = 1$; $N_{\text{var}} = 525$), our method achieves strong performance ($\rho = 0.968$, MAE $= 0.051$), while the per-variable regression baselines are not applicable (they require at least two training years) and matrix factorization performs substantially worse ($\rho = 0.809$, MAE $= 0.215$). When two training years are available ($\text{year} = 2$), our model still leads ($\rho = 0.965$, MAE $= 0.047$) ahead of the strongest regression variant ($\rho = 0.948$, MAE $= 0.055$) and matrix factorization. With more than two years, the gap between our model and the regression variants narrows, though matrix factorization remains less effective. For predictions at greater temporal distances (nearest observed year $\geq 3$ years; $N_{\text{var}} = 929$), our approach achieves $\rho = 0.971$, MAE $= 0.047$, outperforming the strongest regression variant ($\rho = 0.958$, MAE $= 0.054$) and significantly exceeding matrix factorization ($\rho = 0.796$, MAE $= 0.169$). When predicting for periods closer to observed data ($<3$ years; $N_{\text{var}} = 680$), performance remains high ($\rho = 0.985$, MAE $= 0.035$), matching the regression baselines and again clearly surpassing matrix factorization ($\rho = 0.883$, MAE $= 0.087$). In high-volatility scenarios ($N_{\text{var}} = 419$), our method consistently outperforms all benchmarks ($\rho = 0.960$, MAE $= 0.045$), surpassing the strongest regression variant ($\rho = 0.958$, MAE $= 0.044$) and notably exceeding matrix factorization ($\rho = 0.737$, MAE $= 0.109$). Together, these results highlight our method's strengths in challenging contexts---such as rapidly fluctuating public opinion, predictions spanning longer time intervals, and scenarios with limited data---where traditional approaches typically fall short.%

\subsection*{Evaluation of Predicting Between-and Within-Group Variances}

Following~\citet{bisbee2024synthetic}, we evaluate whether the model captures \emph{between-group} and \emph{within-group} response variance across gender, race, age, education, and political-stance subgroups (full subgroup definitions and SD calculation in \hyperref[appendix:sensitivity]{Appendix~F}). The model captures between-group variance well, particularly for political stance and education (\cref{fig:between_group_variance}). For within-group variance, predicted and observed subgroup standard deviations are strongly correlated across categories; nevertheless, the model underestimates the absolute level of within-group variance across most demographic categories (\cref{fig:within_group_variance}).  

\subsection*{Individual-level and Opinion-level Heterogeneity of Model Accuracy}

Despite the strong performance of our models in predicting both personal and public opinion, it is crucial to examine whether this predictive accuracy is distributed evenly across different groups of individuals and types of opinions.
\cref{fig:figure9} shows the performance of our Alpaca-7B based model across various subgroups, including sex, age, period, race, region, education, income, and political ideology, across three different prediction tasks. We estimate OLS regression models to assess between-group gaps in individual-level AUCs with robust standard errors (full model specification, including controls, in \hyperref[appendix:sensitivity]{Appendix~F}).

\input{Figures/figure_individualauc_block.tex}

First, opinions of individuals with higher socioeconomic status (SES), as measured by their levels of education and income, are more predictable than those with lower SES across all three tasks. For retrodiction specifically, individuals with a graduate degree are more predictable than those without a high school degree, as indicated by a $0.028$ higher AUC, whereas those in the highest income bracket are more predictable than those in the lowest, with a $0.013$ higher AUC. Second, our model's prediction is less accurate for racial minorities than for White respondents across all three tasks. Third, our models can predict strong partisans more accurately than Independents across all three tasks. Lastly, individual-level AUC shows a small but significant increase in more recent waves for missing data imputation ($+0.17$ percentage points per decade) and unasked opinion prediction ($+0.14$ percentage points per decade), but is relatively flat for retrodiction ($+0.03$ percentage points per decade).

\cref{fig:figure10} shows which opinions are more or less predictable using the regression analysis of opinion-level AUCs. Focusing on retrodiction (Panel B), the strongest factor is the extent to which opinions are correlated with a seven-point political ideology (Likert: 1 = extremely liberal, 4 = moderate, 7 = extremely conservative). One might also question whether our models exhibited a different performance in 2021 when the survey response rate was $17\%$, the lowest in the history of GSS (the average response rate in the GSS was $72\%$), and the survey mode was altered due to the COVID-19 pandemic. The opinion-level regression provides little evidence for such a 2021 anomaly: missing data imputation actually shows a slightly \emph{higher} 2021 AUC ($+0.030$, $p<.05$), while retrodiction and unasked opinion prediction show no significant difference between 2021 and other years ($-0.004$ and $-0.015$, both $p>.3$). This also reduces our concerns about data leakage. Since all LLMs have been trained with CommonCrawl data collected until September 2021 or earlier, people's survey responses from 2021 would not be directly used to pre-train LLMs. The slightly higher (not lower) imputation accuracy in 2021 suggests that the model's performance is not driven by data leakage.

\input{Figures/figure_opinionauc_block.tex}
Interestingly, the association between ideological polarization and opinion-level AUC is positive but modest in magnitude: across all binarized GSS variables, the Spearman rank correlation between $|\rho(\text{polviews})|$ and Alpaca-7B retrodiction AUC is $0.225$, accounting for approximately $5\%$ of the variance in AUC (\cref{fig:polviews_rank_scatter}). In other words, our model also predicts many non-political domains accurately; the top-40 most ideologically correlated and bottom-40 least correlated GSS variables, each paired with their retrodiction AUC, are reported in \cref{tab:polviews_rank} (full discussion of predictable versus unpredictable items by ideology is in \hyperref[appendix:sensitivity]{Appendix~F}).

Opinion-level retrodiction AUC is also negatively associated with the proportion of non-responses (do-not-know, refusal, inapplicable), and positive associations with training sample size, response rate, and split-ballot design appear primarily in the missing data imputation panel rather than in retrodiction; the binarization procedure does not systematically disadvantage items with odd-numbered response categories (\cref{fig:mean_auc_response_categories}; \hyperref[appendix:sensitivity]{Appendix~F}).

Opinion-level AUC is U-shaped in the positive-response rate, with the highest accuracy at the extreme ends of the opinion spectrum and a modest decrease in the intermediate range (\cref{fig:auc_positive_responses}; \hyperref[appendix:sensitivity]{Appendix~F}). Additionally, the model shows better predictive performance for opinions that are closely located to others in the embedding space, as measured by the average cosine similarity with all other opinions, although the effect size is small.

\section*{Discussion}

In social sciences, accurately predicting public opinion remains a challenge because it is a collective representation of diverse personal opinions, and most individuals do not always hold consistent and coherent beliefs towards various issues~\citep{baldassarriNeitherIdeologuesAgnostics2014,zallerSimpleTheorySurvey1992,conversep.NatureBeliefSystems1964}. While public opinion is generally stable over time with individuals holding firmly to their beliefs~\citep{kileyMeasuringStabilityChange2020}, some attitudes (e.g., same-sex marriage) undergo dramatic shifts~\citep{baunachChangingSameSexMarriage2012}. Against this background, recent studies employing LLMs show limited successes in predicting public opinion accurately and raise questions of demographic representativeness~\citep{santurkarWhoseOpinionsLanguage2023, bisbee2024synthetic, boelaert2024how}. With a flexible methodological framework to tackle these challenges, we show that the LLM-based models trained using survey responses, combined with respondent and period embeddings, can augment survey-based applications such as missing data imputation and retrodiction. 

How can social scientists and decision-makers benefit from our AI-augmented survey methodology? We demonstrate that training LLM-based models with nationally representative surveys enables us to effectively predict a wide range of public opinion while reducing the loss in accuracy or representativeness, tackling the challenges inherent in both digital trace and survey data. The practical applicability of our model for missing data imputation arises from its consistently high accuracy, irrespective of the extent of missing data and different missing mechanisms. Recall that we show more training data increase model performance for missing data imputation and retrodiction (see \cref{fig:figure_a_missing_prop}), but one may wonder how many questions are needed to develop models with reasonably high AUCs.  \cref{fig:figure_a_n_questions} shows that even only asking 50 questions achieved a fair performance (AUC = 0.83), and the performance gain peaked at asking 200 questions (AUC = 0.88). This capability can be useful when survey designs are expected to be impacted by significant attrition in the current era of declining response rates~\citep{senguptaSparseDataReconstruction2023}. Our approach can also help national polls maximize the number of questions. For instance, rather than asking the same ten questions to a thousand participants, pollsters can disseminate twenty questions among the same thousand participants, each answering ten questions, and employ the model to infer individual responses to the remaining ten unasked questions. 

Our model with high accuracy in the retrodiction tasks can help us identify a turning point by looking into past trends -- when they started to shift\footnote{A natural concern is whether counterfactual trends can be reliably inferred from opinions observed in only a small number of survey years. However, \cref{fig:fig_figure_years_vs_auc} shows that predictive performance remains stable regardless of how many years an opinion appears in the training data during fine-tuning}.
\input{Snippets/snippet_busing_detail.tex}

Finally, despite its modest performance in unasked opinion prediction without any human response to entirely new questions compared to other tasks, our best model using Alpaca-7B still shows an AUC of $0.701$ in predicting personal opinions. This performance is comparable to or higher than the performance of recent LLMs without fine-tuning in zero-shot prediction tasks in social science, such as sentiment analysis and ideology detection~\citep{ziemsCanLargeLanguage2023}. We may benefit from this capability to inspire scientific discovery, such as assisting in the selection of relevant survey questions and discovering meaningful hypotheses that involve currently unobserved data~\citep{holmDefenseBlackBox2019}. Nevertheless, at the current level of LLMs' performance, their predictions should not be employed directly for high-stakes decisions that pose tangible risks to individuals. The ethical implications and potential for unintended consequences are too significant to ignore.

What can we learn from these results with regard to the nature of personal and public opinion? The remarkable performance of our predictive models suggests that the notion of personal opinion may not be as personal as it seems. The predictability of personal opinions highlights the inherently social nature of human beings, suggesting that our opinions are embedded in the social contexts that we belong to. This may not be surprising given that LLMs, trained on vast amounts of human-generated text, encode a wide spectrum of human attitudes, as humans utilize technological devices to voice their opinions into the socio-technical reservoir, which in turn, through algorithmic confounding, constrains and shapes what they perceive, observe, and generate~\citep{latourReassemblingSocialIntroduction2007}. 

\input{Snippets/snippet_demographic_predictability.tex}
To ascertain the origins of accuracy gaps across varying demographic groups, we compare the regression outcomes between LLMs and matrix factorization models. Given that matrix factorization models do not use any textual information, biases from matrix factorization models may indicate the extent of biases attributable to group differences in belief systems, as opposed to a biased pretraining corpus. \cref{fig:alpaca_vs_mf_individualauc} displays similar patterns between the two models, suggesting that these gaps are likely a result of different levels of belief organization across different demographic groups. The full per-subgroup comparisons for missing data imputation and retrodiction are reported in \cref{tab:subgroup_groupauc_imputation,tab:subgroup_groupauc_retrodiction}.

Our study brings several ethical concerns into focus concerning the use of LLMs to predict personal and public opinion. A major concern lies in the realm of privacy and surveillance. This paper shows a possibility that LLMs may be able to accurately estimate personal opinions that respondents might not be willing to share or may have chosen not to answer.
Therefore, it is crucial to engage in discussions about how we can maintain respondent privacy and data protection while concurrently enhancing the accountability and responsibility of generative AIs~\citep{shevlane2023model}. 

Parallel to concerns of privacy and surveillance, ethical considerations regarding individual autonomy and demographic representation are of equal importance. Despite its accuracy, predicting a person's opinion without their consent can be seen as a potential infringement on their autonomy. This concern widens when viewed from a societal perspective, particularly in contexts of democracy since autonomous opinion formation is a fundamental component of democratic processes~\citep{bursteinImpactPublicOpinion2003,shapiroPublicOpinionAmerican2011}. Our model's lower accuracy for individuals with low socioeconomic status, racial minorities, and non-partisan affiliations can exacerbate the demographic representation issue.

\subsubsection*{Future Directions and Extensions}

The current study demonstrates the potential of LLM-based models trained using surveys for simulating unasked survey responses, yet much remains to be explored. As the landscape of public opinion research continues to evolve, it is crucial to investigate how these models perform in broader contexts and under varying data conditions. Here, we discuss some of the current challenges and future directions of the proposed AI-augmented survey approach.

\input{Snippets/snippet_binary_scope.tex}

\input{Snippets/snippet_unasked_validation_limit.tex}

Third, our internal and external cross-validation results show high accuracy and strong correlations between observed and predicted responses, increasing confidence in our ability to retrodict counterfactual opinion trends. More broadly, the existence of multiple nationally representative surveys in the United States creates an important opportunity to build systematic benchmark datasets for public opinion prediction~\citep{enns2024needa}. Such benchmarks would not only provide standardized evaluations across surveys, but also supply valuable training data for fine-tuning future models. For example, researchers could test whether models fine-tuned on one survey (e.g., the GSS) can accurately predict missing opinions in another (e.g., the ANES), and vice versa.

Fourth, while our method is designed to be model-agnostic, indicating its compatibility with various LLMs for interpreting survey questions, the effectiveness of fine-tuning strategies remains uncertain when applying non-English LLMs or extending our approach to different national contexts, such as in the case of World Value Surveys. Recently, Atari and colleagues \citep{atari_which_2023} show that vanilla LLMs' answers to questions in the World Value Surveys closely align with responses typically observed in Western, Educated, Industrialized, Rich, and Democratic (WEIRD) countries with the highest correlations with the United States, and the efforts to align LLMs with human values could unintentionally increase the WEIRD bias~\citep{ryan2024unintended}. Future studies should examine whether fine-tuning LLMs with cross-cultural surveys, such as the World Value Survey, could address this bias, and whether LLMs could be applied for cross-cultural opinion prediction. 

Fifth, future research should examine how well this framework generalizes beyond nationally representative surveys like the GSS. While the GSS offers high-quality data for fine-tuning and evaluation, many applications rely on smaller, localized, or non-representative surveys. One promising direction is to fine-tune models on multiple datasets containing GSS-like questions, enabling them to learn broader relationships among attitudes, demographics, and historical contexts that extend beyond a single survey instrument. Such a hybrid approach could combine the strengths of high-quality repeated surveys with the breadth of ad hoc online datasets. However, important challenges remain, including limited generalizability from small or infrequent surveys and variation in semantic similarity and correlational structures across datasets. However, the following issues need to be addressed: first, it is unclear how well the model can generalize when applied to non-representative surveys conducted only once or twice with small sample sizes; second, the effectiveness of combining datasets depends on the degree of semantic similarity between questions and the underlying correlational structures of opinions across surveys. 

\input{Snippets/snippet_full_finetuning_tradeoff_letter}

Finally, the model’s underestimation of within-group variance highlights the need for further refinement to better capture group-level variations and nuanced group-specific trends. This limitation highlights the challenge of ``uniformity'' in LLM-generated responses, where models produce overly consistent outputs that fail to reflect real-world diversity. To address this, strategies like ``persona jittering,'' which introduces subtle variations in model prompts, could enhance response variability~\citep{kozlowski2024simulating}. A more advanced approach could involve a dynamic life-state model that simulates changing contexts, such as career progressions~\citep{savcisens2024using,vafa2022career}. By integrating such adaptive mechanisms, LLMs could better capture the complexity of human behavior, enabling more realistic survey simulations and social interaction modeling.

We anticipate that these limitations may soon be addressed as the scale of LLMs expands and more scholarly focus is directed toward enhancing the integration of LLMs with social surveys for opinion prediction. We believe that our research marks a foundational step and has shown promising potential for the future of social science research using the LLMs. With the rapid advancement in LLM-related applications, more and more survey researchers may consider using the AI-augmented survey approach or similar kinds. We make our code and data available at [suppressed for peer review] to facilitate the replication and extension of our novel approach.

\singlespacing

\bibliography{asrbib}
\bibliographystyle{asr}

\clearpage
\noindent

\noindent
\textbf{Acknowledgments:} This work was supported by the National Science Foundation (\#2335815). This work was completed in part with resources provided by the University of Chicago's Research Computing Center and in part by Lilly Endowment, Inc., through its support for the Indiana University Pervasive Technology Institute. We thank Peter Bearman, Delia Baldassarri, Jonathan Bach, Bart Bonikowski, Philipp Brandt, Siwei Cheng, Sarah Cowan, Dalton Conley, Yuting Chen, Ryan Dai, Paul DiMaggio, James Evans, Gil Eyal, Ryan Hagen, Mark Hoffman, Mike Hout, Theodora Hurley, Robert Max Jackson, Wontak Joo, Donghyun Kang, Hyunku Kwon, Keunbok Lee, So Yoon Lee, Seungwon Lee, John Levi Martin, Kinga Makovi, Lina Moe, Ravaris Moore, Austin Kozlowski, Sebastian Ortega, Barum Park, Bernice Pescosolido, Brian Powell, Alix Rule, Diana Sandoval Siman, David Stark, Daniel Tadmon, Josh Whitford, Tytus Wilam, Yoosik Youm, Linda Zhao, Simone Zhang, the members of Knowledge Lab at the University of Chicago, and the members of Networks in Context Lab for their helpful comments. This work was presented at the 9th IC2S2 conference, the 2023 American Sociological Association Annual Meeting, Korea Inequality Research Network Symposium, the CODES seminar at Columbia University, the Inequality Workshop at NYU, the Generative SI and Sociology Workshop at Yale University, AI-Data Intersection for Social Good at Yonsei University, at the Stone Center on Socio-Economic Inequality at the CUNY Graduate Center, the Department of Technology, Operations, and Statistics Seminar Series at NYU Stern School, Research Seminar in QMSS at Columbia University, the Thought Summit on the Future of Survey Science at Cornell University, and Artificially Intelligent Social Science Workshop at Nuffield College -- University of Oxford.\vspace{1em}

\noindent
\textbf{Data and materials availability:} All data and code necessary for replicating our analyses, as well as counterfactual public-opinion trends predicted by our method, will be made available upon publication of this paper.\vspace{1em}

%% file: Figures/figure1_three_problems_block.tex
\begin{figure}[H]
  \begin{center}
      \centering
     \includegraphics[width=1\columnwidth]{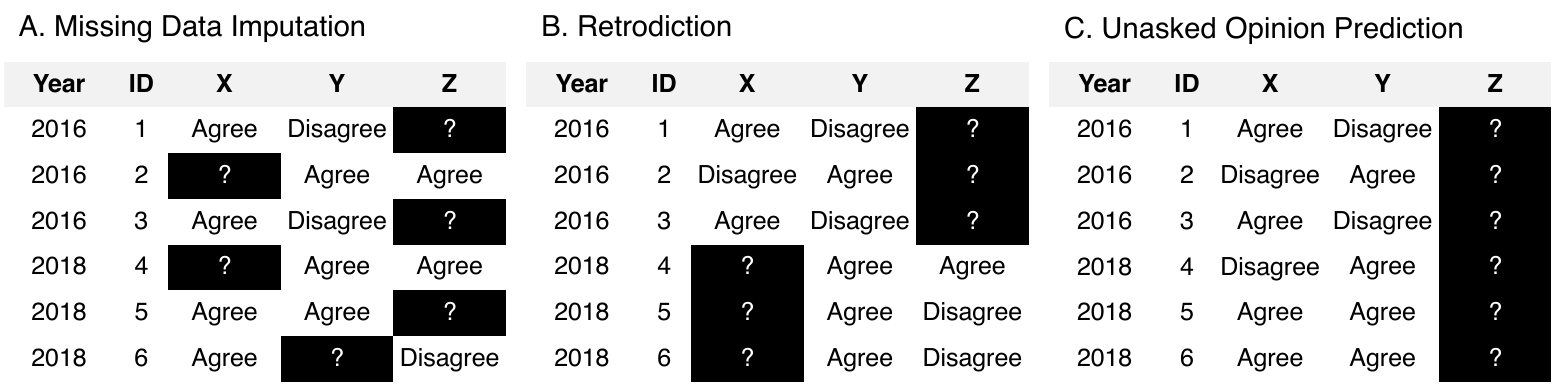}
  \caption{\textbf{Three types of missing problems in survey research.} Panels A-C illustrate three typical missing data challenges in survey research. Each row indicates an individual subject in a social survey across different periods, and each column (i.e., X, Y, and Z) indicates public opinion variables that we aim to measure. The machine learning task in each situation is to predict the unobserved values {[}?{]} in the black cells using the observed values in the white cells.} 
  \label{fig:figure1}
  \end{center}
\end{figure}

%% file: Figures/figure2_architecture_block.tex
\begin{figure}[H]
  \begin{center}
      \centering
     \includegraphics[width=1\columnwidth]{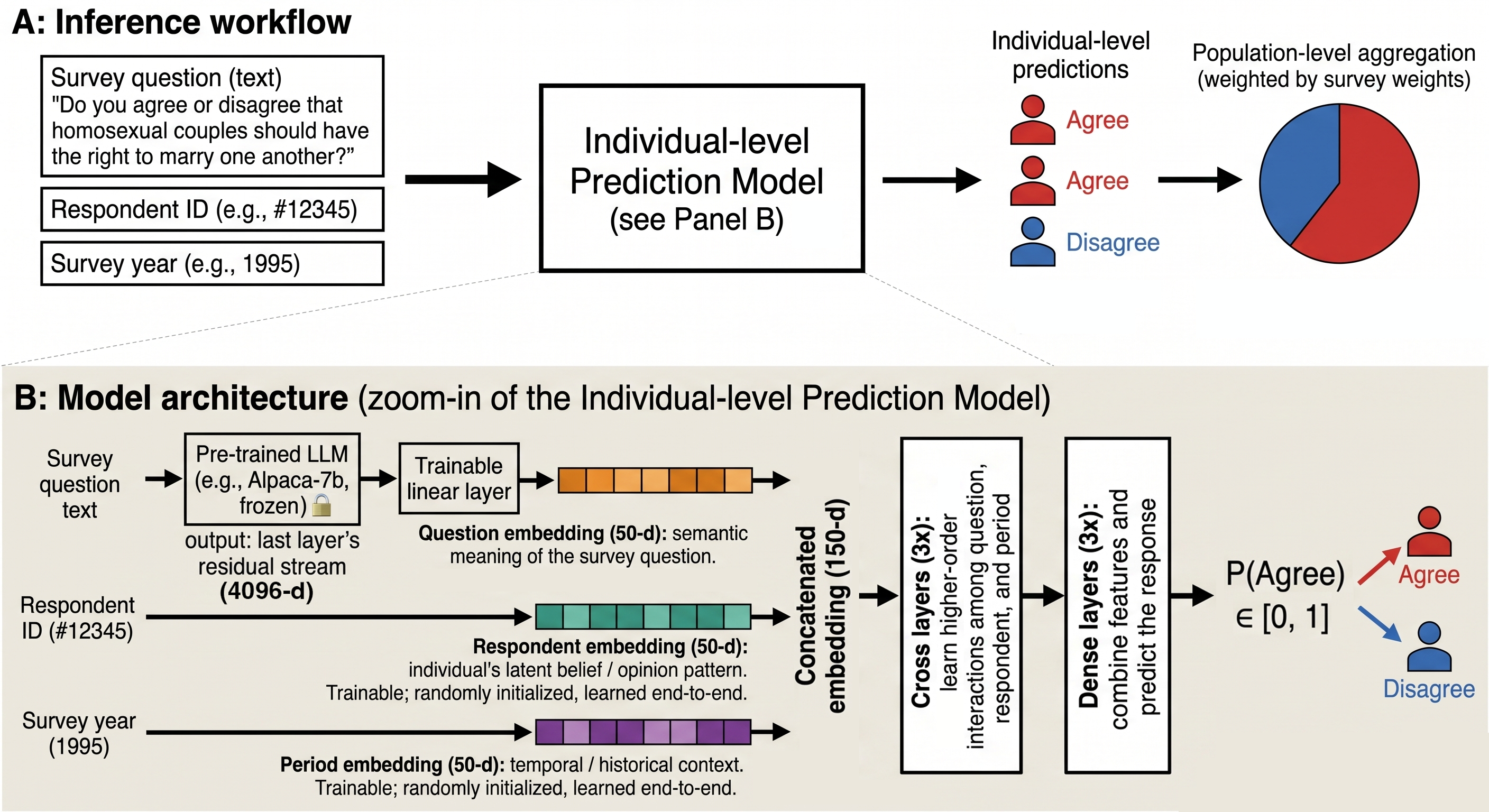}
  \caption{\textbf{Architecture overview.} Panel~A presents the end-to-end inference workflow: survey-question text, respondent ID, and survey year are fed into the individual-level prediction model, and the resulting per-respondent predictions are aggregated using survey weights to produce a population-level estimate. Panel~B provides a closer view of the prediction model: A language model like Alpaca-7B (frozen, indicated by the lock icon) encodes only the question text, after which a small trainable linear projection maps it to a 50-dimensional vector; respondent and period embeddings are randomly initialized and learned end-to-end (and crucially, do \emph{not} pass through the frozen Alpaca-7B backbone). The three 50-dimensional embeddings are then concatenated and passed through three Cross layers (modeling higher-order interactions among respondent, question, and year) and three Dense layers, yielding $P(\text{Agree}) \in [0,1]$.}
  \label{fig:figure2}
  \end{center}
\end{figure}

%% file: Snippets/snippet_param_efficient_def.tex
In our implementation, these trainable components are the respondent and period embeddings, the lightweight projection layer applied to the LLM-derived question embedding, and the Deep Cross Network with its output head, all optimized by gradient descent on the GSS data.

%% file: Figures/figure_modelperformance_block.tex
\begin{figure}[H]
  \begin{center}
      \centering
     \includegraphics[width=0.95\columnwidth]{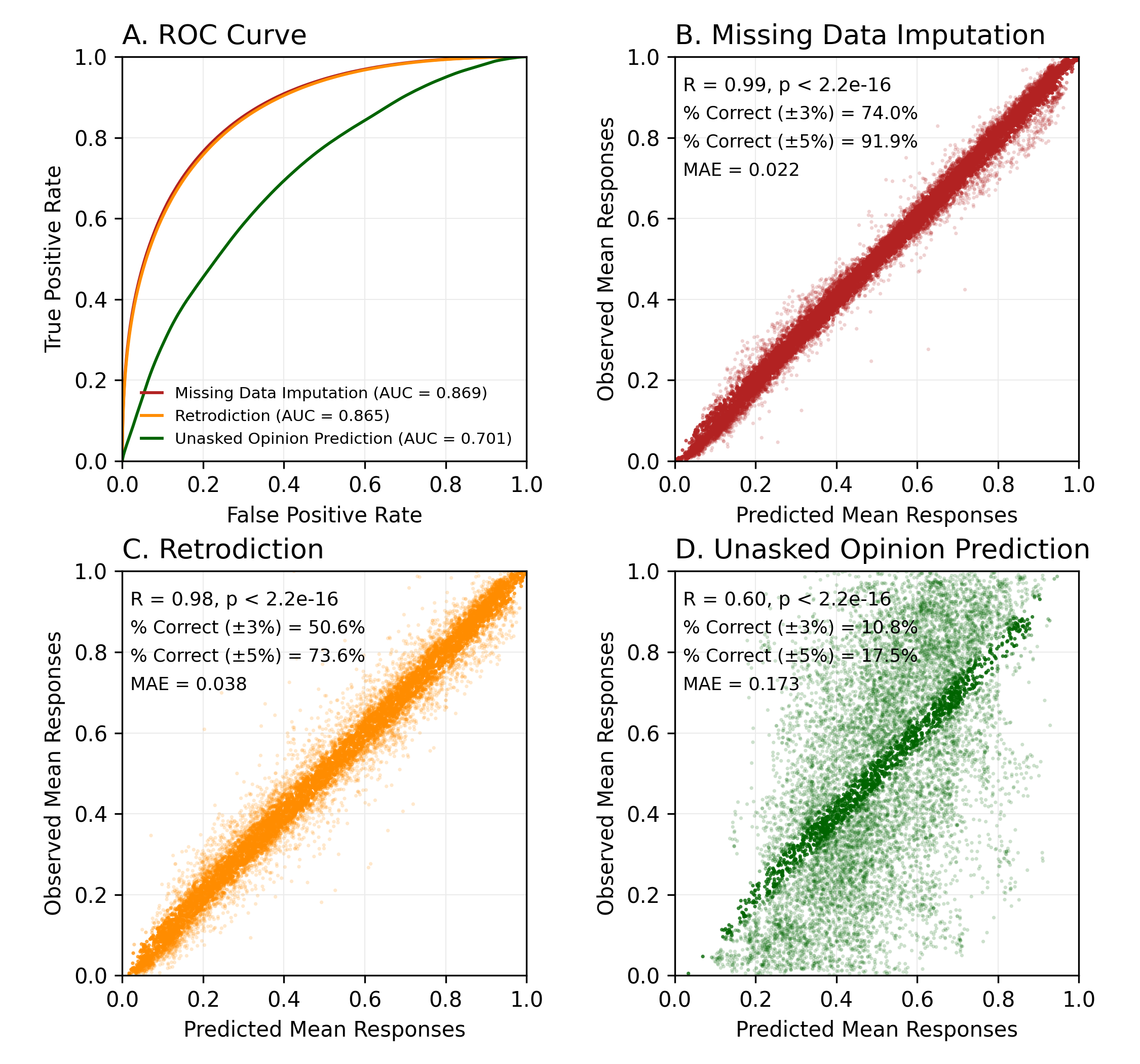}
  \caption{\textbf{Model performance for predicting three types of missing responses at individual and aggregate levels.} Panel A displays the Receiver Operating Characteristic (ROC) curve, indicating how well the Alpaca-7b based model can predict missing responses at an individual level. We also denote the AUC (Area Under Curve) values, i.e., the probability of the model ranking a randomly selected positive response over a randomly selected negative response. Panels B--D depict the relationship between the observed proportion of those who agree in each survey year and the predicted mean response (i.e., proportion of agreement) for the same opinion. Each panel also annotates the Pearson correlation $R$, the percentage of year-variable estimates whose prediction falls within $\pm 3\%$ (``\% Correct $\pm 3\%$'') and $\pm 5\%$ (``\% Correct $\pm 5\%$'') of the observed value, and the mean absolute error (MAE). }
  \label{fig:figure5}
  \end{center}
\end{figure}

%% file: Figures/figure_roper_3panel_block.tex
\begin{figure}[H]
  \begin{center}
    \includegraphics[width=0.95\columnwidth]{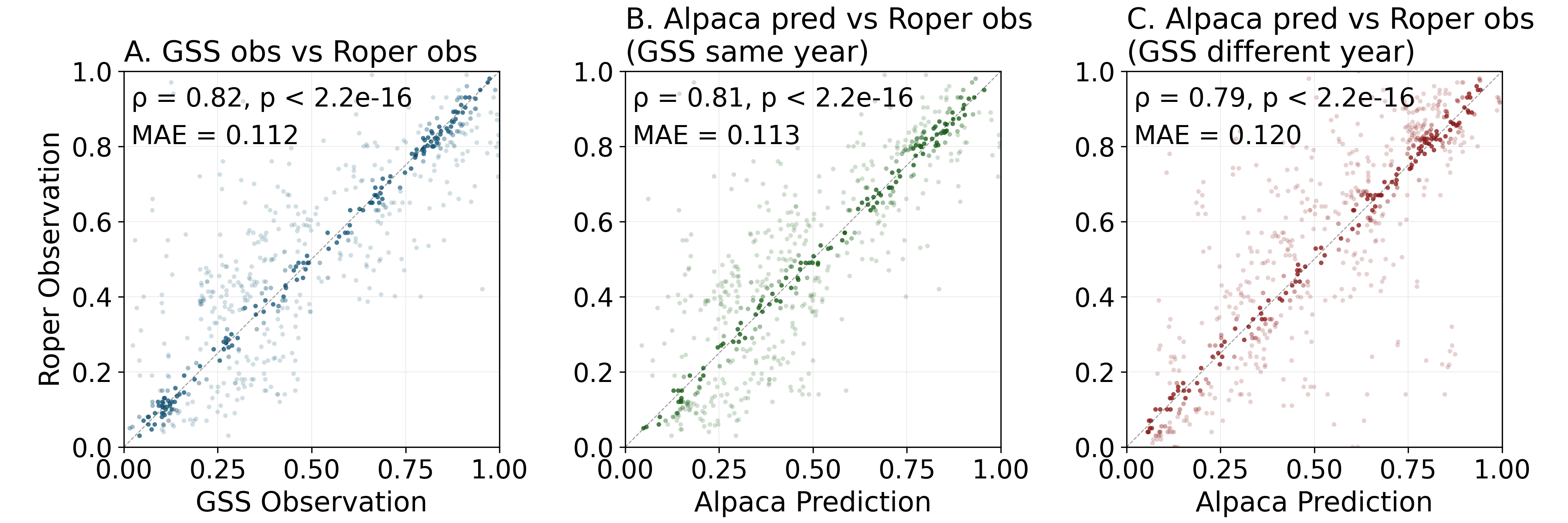}
    \caption{\textbf{External validation against the Roper Center iPOLL archive.} GSS items are matched to Roper Center items via cosine similarity in the question-embedding space and verified by Gemini-2.5-Pro as same-question pairs. Panel~A compares observed GSS means to observed Roper means at the same (variable, year). Panel~B compares Alpaca-7B retrodictions to observed Roper means at the same (variable, year). Panel~C compares Alpaca-7B predictions to observed Roper means at (variable, year) pairs where the GSS did \emph{not} field the item, so the prediction has no GSS training signal at that year. Each panel reports Spearman $\rho$ and MAE.}
    \label{fig:roper_3panel}
  \end{center}
\end{figure}

%% file: Figures/figure_counterfactual_block.tex
\begin{figure}[H]
  \begin{center}
      \centering
     \includegraphics[width=0.9\columnwidth]{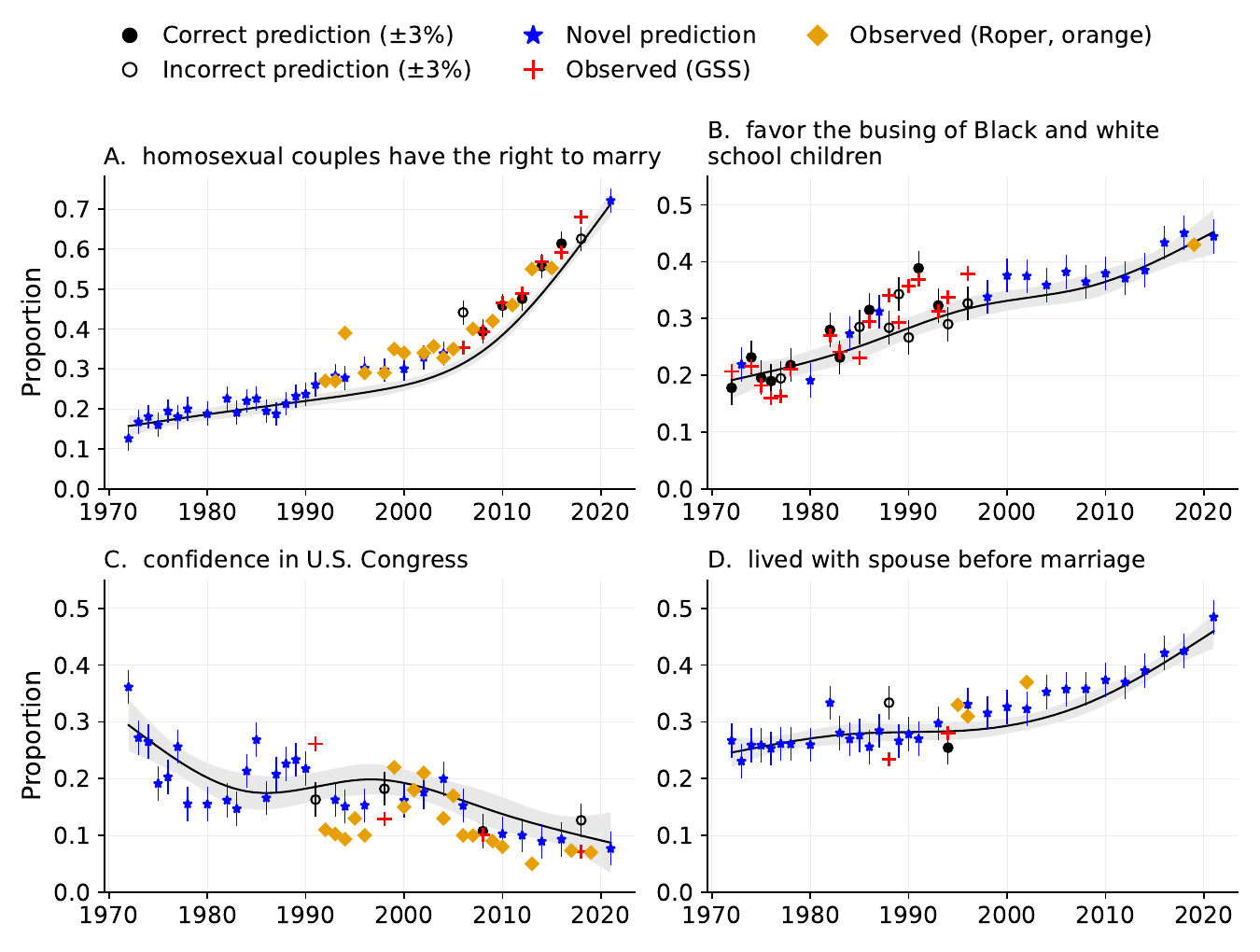}
  \caption{\textbf{Counter-factual trends predicted by our model in the GSS 1972--2021, with Roper-Center external validation.} Panels A--D show Alpaca-7b predictions over time for four GSS variables, smoothed by a generalized additive model. Filled circles mark predictions within $\pm 3\%$ of the observed GSS mean (correct); open circles mark predictions outside $\pm 3\%$ (incorrect); blue stars mark predictions for years in which the GSS did not field the question (novel predictions). Red plus signs overlay observed GSS means; orange diamonds overlay matched national-adult observations from the Roper Center iPOLL archive (see Methods for the matching pipeline). Panels: \textbf{A}: \textit{homosexual couples have the right to marry} (\texttt{marhomo1}); \textbf{B}: \textit{favor the busing of Black and white school children} (\texttt{busing}); \textbf{C}: \textit{confidence in U.S.\ Congress} (\texttt{concong}); \textbf{D}: \textit{lived with spouse before marriage} (\texttt{cohabit}).}
  \label{fig:figure6}
  \end{center}
\end{figure}

%% file: Figures/figure_individualauc_block.tex
\begin{figure}[H]
  \begin{center}
      \centering
     \includegraphics[width=1\columnwidth]{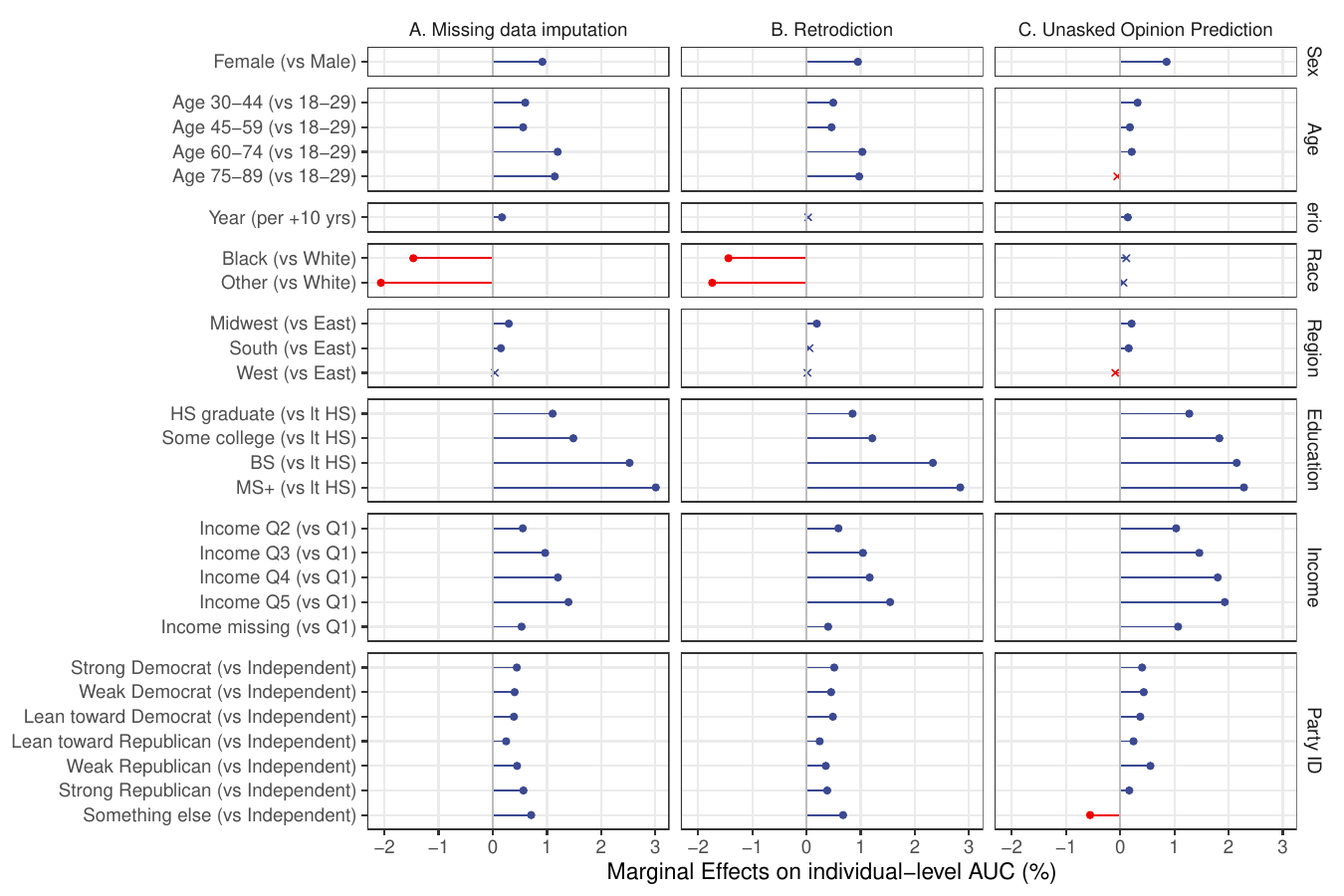}
    \caption{\textbf{Coefficient plots from OLS regression models predicting individual-level AUC across three different types of missing response prediction.} A higher AUC value indicates greater model accuracy for individuals. Here, each dot represents the expected difference of AUC (i.e., average marginal effects) against the reference group within each subgroup. Red bars indicate that the AUC for a particular group is below the AUC of the reference group, and blue bars indicate that the AUC for a particular group is above the AUC of the reference group. A filled dot refers to a statistically significant difference and an X refers to a statistically insignificant difference, based on robust standard errors ($p < 0.05$).}
  \label{fig:figure9}
  \end{center}
\end{figure}

%% file: Figures/figure_opinionauc_block.tex
\begin{figure}[H]
  \begin{center}
      \centering
     \includegraphics[width=1\columnwidth]{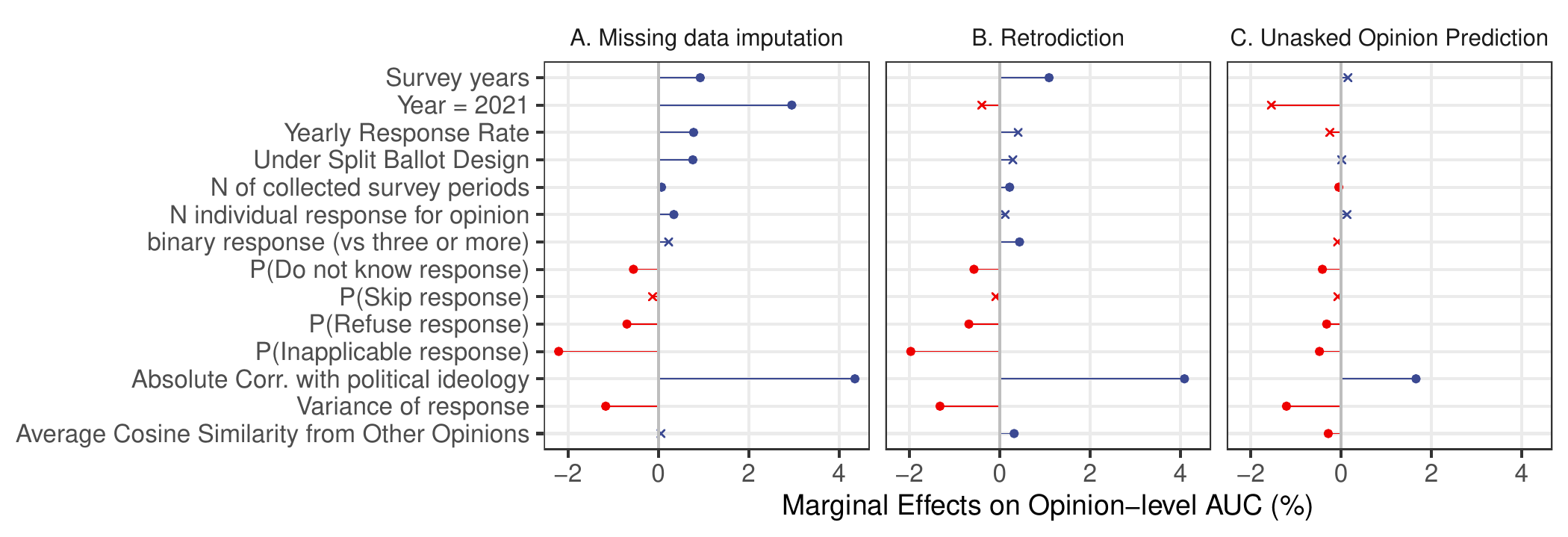}
  \caption{\textbf{Coefficient plots from OLS regression models predicting opinion-level AUC across three different types of missing response prediction.} A higher AUC value indicates greater model accuracy for opinions each year. Each dot represents the average marginal effects of each variable. To compare effect sizes across different continuous covariates, we standardize them in the regression analysis. Red bars indicate that the variable is negatively correlated with opinion-level AUC, and blue bars indicate that it is positively correlated. A filled dot refers to a statistically significant difference and an X refers to a statistically insignificant difference, based on robust standard errors ($p < 0.05$).}
  \label{fig:figure10}
  \end{center}
\end{figure}

%% file: Snippets/snippet_busing_detail.tex
A particularly illustrative case is the busing-policy item. The GSS asked respondents whether they favored busing Black and white schoolchildren between school districts only through the early 1990s, after which the question was discontinued. Our counterfactual reconstruction predicts that opposition to busing would have remained broadly stable after the early 1990s rather than changing sharply. This prediction is validated by an independent Gallup poll (archived in the Roper Center iPOLL database) that re-fielded a closely matched busing question in 2019 and produced an aggregate response very close to our 2019 estimate; the full set of matched external surveys is reported in \cref{tab:counterfactual_roper_surveys}.

%% file: Snippets/snippet_demographic_predictability.tex
The higher predictability of individuals with higher SES and stronger partisanship aligns with theories of political belief systems, as these respondents tend to hold coherent belief structures with highly correlated opinions~\citep{zallerSimpleTheorySurvey1992,conversep.NatureBeliefSystems1964}. More broadly, recent work shows that agreement about the cultural order is stratified by social position: high-SES and majority-group respondents converge more closely on shared evaluative orderings of occupations, statuses, and cultural objects than low-SES and marginalized respondents~\citep{lynn2016position,lynn2024intersectional,shi2025beyond}. Higher predictability of high-SES, White, and partisan respondents might reflect that they sit closer to a tightly shared cultural schema. The lower predictability of Black respondents might reflect two compounding factors that are hard to fully separate: their underrepresentation in the GSS data (roughly 14\% per year) and the documented under-representation of Black voices in LLM pretraining corpora~\citep{dimaggioDigitalInequalityUnequal2004,nadeemStereoSetMeasuringStereotypical2020}. 

%% file: Snippets/snippet_binary_scope.tex
First, our current models dichotomize survey responses into positive and negative categories to provide an intuitive representation of opinions. However, many survey responses are inherently more complex, involving ordinal scales and nominal categories. Accordingly, the binary setting here represents a particularly favorable case for our framework, and should not be interpreted as evidence of ordinal- or multi-class performance. The current approach is best suited to public opinion questions organized along a binary dimension such as support for same-sex marriage, abortion, or immigration policy.  More fundamentally, even when a binary model accurately distinguishes “for” versus “against,” its continuous outputs do not necessarily capture variation in response intensity (e.g., strongly agree vs.\ agree) \citep{welbl2021detoxifying, pavlopoulos2021semeval}. Future work should therefore extend this framework to ordinal and multi-class prediction through alternative classification architectures or decoders.

%% file: Snippets/snippet_unasked_validation_limit.tex
Second, our validation of unasked opinion prediction is limited to questions semantically related to items the GSS has fielded at some point in its history. It does not establish how well the model generalizes to genuinely novel topics that the GSS has never measured (e.g., attitudes toward K-pop), for which no ground truth exists in our data. The reported accuracy should therefore be interpreted as evidence of generalization across held-out but related items, rather than as a guarantee of performance on entirely new domains. At the same time, modern LLMs are pretrained on large-scale digital traces, including web searches and social-media discourse, which may help them infer opinions on novel topics. LLMs may carry implicit knowledge of how the public has \emph{talked} about topics that no probability sample like GSS has ever asked about. Whether such pretraining signals can be effectively combined with survey-grounded models to produce reliable predictions on genuinely new variables—and how to validate those predictions—remains an important direction for future research.

%% file: Snippets/snippet_full_finetuning_tradeoff_letter.tex
Sixth, an important direction for future research is to evaluate the trade-offs of fully fine-tuning the LLM backbone rather than freezing the LLM embeddings, as we do here. On the one hand, full fine-tuning may adapt the backbone representations to the survey prediction task and potentially improve accuracy on items where frozen embeddings are suboptimal. However, it is substantially more computationally expensive, risks overfitting on relatively small survey datasets like the GSS, and may weaken the pretrained knowledge that supports out-of-distribution generalization in unasked opinion prediction \citep{kumar2022finetuning}. As a preliminary step, we conducted a small-scale LoRA-based fine-tuning experiment, which produced little gain in AUC and accuracy and slightly lower F1 scores across all prediction tasks (see Appendix~I). A more systematic analysis of these trade-offs remains for future work.

%% file: Body/appendix_body_v2.tex
\section*{Appendix A. LLM prompts used in the construction of the dataset and benchmarks}\label{appendix:raw_prompts}

This appendix presents the LLM prompts used in our pipeline. All prompts were submitted via the Gemini batch API (Gemini-2.5-Pro) at temperature $T = 0.2$ except where noted. The prompted out-of-shelf LLM benchmark used GPT-4o and Gemini-2.5-flash with $T = 0$.

\subsection*{A.1. GSS binarization prompt}\label{appendix:binarization_prompt}

The following system prompt is sent for every candidate variable. \texttt{<labels>} expands to the list of code labels for that variable.

\begingroup\par\noindent
\begin{verbatim}
You are a survey data analyst. Your task is to determine
whether a categorical survey variable can be meaningfully
binarized into two groups (coded as 1 and 0), and if so,
produce the mapping.

## Rules
1. can_binarize = 1 if the response categories can be split
   into two substantively meaningful, non-trivial groups
   along a single dimension (e.g., "yes vs no",
   "working vs not working", "agree vs disagree",
   "married vs not married"). Both groups must contain at
   least one category.  Otherwise, can_binarize = 0.
2. When can_binarize = 1, assign:
   - 1 to the "positive / active / yes / agree / present /
     more" side
   - 0 to the "negative / inactive / no / disagree /
     absent / less" side
   - "Neither" / "nor" / neutral / middle-ground responses
     -> always 0
   Use the question wording and variable label to decide
   which side is which.
3. When can_binarize = 0, fill all keys in `binarized` with 0.
4. Variable type constraints
   Ordinal scales CAN be binarized.
   Nominal categorical variables WITHOUT ordinal or binary
   structure CANNOT be binarized (e.g., "Which state do you
   live in?").  Continuous variables (e.g., age) also
   CANNOT be binarized.
5. Missing responses such as "refused to answer" or
   "don't know" should be coded as null and not binarized.

Variable name: <var>
Variable label: <label>
Question: <question>

Code labels (use these EXACTLY as keys in `binarized`):
- <label_1>
- <label_2>
- ...

Return JSON with `can_binarize` (0 or 1) and `binarized`
(dict with every label above as a key, each value 0, 1,
or null).
\end{verbatim}
\endgroup\par

\subsection*{A.2. GSS categorization prompt (nine-category question type)}

\begingroup\par\noindent
\begin{verbatim}
You are a survey methodology expert. Your task is to
categorize a given survey question into one of the
following nine categories.

## Categories
1. Behavioral Questions: Questions related to actions or
   activities performed by the respondent.
2. Attitudinal/Opinion Questions: Questions that capture
   the respondent's beliefs, attitudes, or opinions.
3. Demographic Questions: Questions that gather information
   about the respondent's demographic characteristics, such
   as age, gender, or education level.
4. Objective/Knowledge/Factual Questions: Questions
   assessing the respondent's knowledge or factual
   information.
5. Questions about Other People: Questions about
   relationships, household rosters, or other individuals
   connected to the respondent.
6. Open-ended Questions: Questions that allow the
   respondent to provide responses in free-text form.
7. Derived/Computed/Condition Questions: Questions derived
   or calculated based on other responses or dependent on
   specific conditions.
8. Metadata/Paradata: Questions or information related to
   the survey administration process.
9. Miscellaneous Questions: Questions that do not fit
   neatly into the other categories.

Variable name: <var>
Variable label: <label>
Question: <question>

Return JSON: {"category": "<one of the nine categories>"}
\end{verbatim}
\endgroup\par

\subsection*{A.3. Roper-Center same-question verification prompt}\label{appendix:roper_verify_prompt}

After cosine-similarity matching of GSS and Roper-Center questions, every retained pair is verified by Gemini-2.5-Pro using the following prompt. Pairs are kept only if the model returns ``Yes'' with confidence at least $0.85$.

\begingroup\par\noindent
\begin{verbatim}
Task:
You are given two survey questions. Determine whether
a positive response (e.g., "yes", "agree") to Survey
Question 1 would generally imply a positive response
to Survey Question 2.

Output Format:
Return ONLY a valid JSON object in the following form:
{"answer": "Yes" or "No",
 "confidence": <float between 0.0 and 1.0>}

Survey Question 1: "<gss_question>"
Survey Question 2: "<roper_question>"
\end{verbatim}
\endgroup\par

\subsection*{A.4. Roper-Center response-binarization prompt}\label{appendix:roper_binarize_prompt}

For each verified GSS--Roper pair, we map Roper response options to the GSS binary coding using:

\begingroup\par\noindent
\begin{verbatim}
Task:
You are given a GSS survey question with its binarization
mapping, and a matched Roper survey question with response
options and their percentages.

Map each Roper response to 1 (positive/agree) or 0
(negative/disagree) following the GSS binarization logic.

GSS Question: "<gss_question>"
GSS Binarization: <gss_binarization_mapping>

Roper Question: "<roper_question>"
Roper Responses: <list of (response_text, percentage)>

Output Format:
Return ONLY a valid JSON object:
{"mapping": {"<response_text>": 0 or 1 or null, ...},
 "roper_yes_pct": <sum of percentages mapped to 1>}
\end{verbatim}
\endgroup\par

\subsection*{A.5. Out-of-shelf LLM benchmark prompt (correlation-top-$k$ context)}

The following prompt is sent to GPT-4o and Gemini-2.5-flash for the prompted-baseline benchmark on the 2018 GSS data, with the ten in-context items chosen as the GSS items most correlated with the target on the training pool. Demographics-only and random-$k$ variants drop or replace the context items accordingly. All calls use temperature $T = 0$.

\begingroup\par\noindent
\begin{verbatim}
[System]
Assume you are a survey participant with the given
description. You are also given the participant's
answers to other questions.  Using this information,
predict the participant's answer to the final survey
question.  Also predict the confidence of your answer
on a scale of 0 to 100.  Answer in JSON.

[User]
I am answering the survey question in 2018.
Ideologically, I describe myself as moderate.
Politically, I am Democrat.  Racially, I am White.
I am female.  Financially, our annual family income
is 60000$.  In terms of my age, I am 45.  I was born
in 1973.  Regarding education, I have attained a
Bachelor's degree.  I live in East North Central.
Regarding the size of my area, I reside in suburbs of
medium-sized cities.  My ethnic background is German.
I am currently working full time.  I am married.
Religiously, I identify as Protestant.

Question: <Q1>.  Response Options: 1: <pos labels>
                                 / 0: <neg labels>.
Answer: <A1>

... [eight more (question, answer) context pairs,
selected as the 10 GSS items most correlated with the
target on the training pool] ...

Question: <target question>.  Response Options:
  1: <pos labels> / 0: <neg labels>.  Answer:
\end{verbatim}
\endgroup\par

\clearpage
\section*{Appendix B. Model Architecture Detail}\label{appendix:architecture_detail}
\label{sec:architecture_detail}

\noindent \subsection*{Question Embedding}
To obtain question embedding, we begin with sentence-level embeddings from LLMs pre-trained on vast text corpora. These enable us to encode the meaning of survey questions, such as ``Do you agree or disagree that homosexual couples have the right to marry one another?'' and map these questions into a latent vector space~\citep{jurafskySpeechLanguageProcessing2023}. To determine the optimal model for our task, we have conducted extensive experiments with three LLMs with varying architectures and parameters (Alpaca-7B, GPT-J-6B, RoBERTa-large)~\citep{liuRobertaRobustlyOptimized2019,taorirohanAlpacaStrongReplicable2023,wangbenGPTJ6BBillionParameter2021}. Specifically, we pass survey questions into the LLMs and extract the last token's embedding for decoder-only models (Alpaca and GPT-J) and a special token embedding (or pooler output) for encoder-only models (RoBERTa-large). Note that LLMs have different dimensions of embeddings. For instance, Alpaca-7B has 4096 dimensions, while RoBERTa-large has 1024 dimensions. We use a single trainable feed-forward layer (the projection layer) to map the LLM's last-token output to a 50-dimensional question embedding, regardless of the original LLM's dimensionality. This 50-dimensional projection is the only LLM-side trainable component; the backbone weights of Alpaca-7B remain frozen. We employ the following prompt, based on Alpaca's instruction data template~\citep{taorirohanAlpacaStrongReplicable2023}, to extract question embeddings from pre-trained LLMs: ``Below is an instruction that describes a task. Write a response that appropriately completes the request. \#\#\# Instruction: {[}SURVEY QUESTION{]} \#\#\# Response: '' To ensure that the question embedding can accurately capture the meaning of survey questions, we generate text responses using this embedding to the question, and assess whether the responses seem to align with what humans might say. \cref{tab:alpaca_text_examples} demonstrates that the pre-trained LLMs accurately understand the semantic meaning of survey questions and generate human-like answers even before GSS-based training.

Next, we train the sentence embeddings using actual survey responses to better reflect survey contexts. During this process, the embedding layers' weights are updated to make an accurate prediction of a binary response to all questions in the training data (0 or 1). The sentence embeddings are contextualized such that two questions with similar response patterns are more closely mapped within the embedding space. This enables the trained models to learn meanings of survey questions shared by respondents and predict appropriate responses. As a result, the models can interpret the meaning of questions, such as ``Do you agree or disagree with the following statement? Homosexual couples have the right to marry one another.'' and map them into other seemingly unrelated questions such as ``Some people think the use of marijuana should be made legal. Other people think marijuana use should not be made legal. Which do you favor?'' (also see \cref{tab:similar_questions}, which presents survey questions that are most similar to the four questions used as an illustration in our paper in the embedding space).

The trained embeddings can capture the similarity in meanings of questions as well as their response patterns in surveys is possible because the trained models learn its meaning not only based on the text alone, but also based on the expected patterns in survey responses in a way that the predictability of responses is maximized. We refer to this embedding vector, \(s\  \in \ R^{n}\)\, as the ``question embedding'' of survey questions.

\subsection*{Respondent Embedding}

A breakthrough we made for personalizing LLMs is to incorporate respondent embeddings to account for individuals' heterogeneous responses to survey questions based on heterogeneous belief systems~\citep{baldassarriNeitherIdeologuesAgnostics2014,milbauerAligningMultidimensionalWorldviews2021}. Recent approaches like in-context tuning utilize various prompts to steer LLMs to represent group-level opinions~\citep{brownLanguageModelsAre2020}, but they fail to account for unique individual perspectives because these methods can only predict the commonly held opinions of a ``typical'' group~\citep{argyleOutOneMany2023,santurkarWhoseOpinionsLanguage2023}. Traditional machine learning models have primarily focused on producing a single output for a given input. For instance, when an input sentence like ``the weather is so good'' is given, models for sentiment analysis are designed to generate a single output, such as ``positive.'' However, this method may not generate reliable responses for controversial issues. For example, people's responses to a question about their stance on gun control will depend on their pre-existing beliefs on other issues such as religious beliefs or political ideology.

To address these challenges, we add an additional layer of neural embedding in our model architecture---respondent embedding. We build on recent studies that develop a methodological framework to account for individual heterogeneity, particularly focusing on instances where individuals differ in their interpretations regarding the true labels in a given task~\citep{gordonJuryLearningIntegrating2022,parkDetectingCommunitySensitive2021}. These models assign labels on an individual basis, instead of applying a uniform label to everyone. As such, we assume that each individual's beliefs are represented as an N-dimensional embedding vector. We refer to this vector, \(b\  \in \ R^{n}\), as the ``respondent embedding'' of individuals. To generate respondent embeddings, we initially assign random latent features for each individual and optimize them during the training process such that two individuals closely located in this embedding space are either more likely to have a similar set of beliefs or to have similar latent correlational structure of beliefs~\citep{gordonJuryLearningIntegrating2022}. These two properties emerge from the Deep Cross Network (DCN) architecture detailed below.

\subsection*{Period Embedding}

The final component necessary for contextualizing LLMs is to account for temporal variations in the interpretation of survey questions and individual belief systems. While some scholars argue that people's beliefs are generally stable, attributing fluctuations in opinions to short-term changes or measurement errors~\citep{kileyMeasuringStabilityChange2020,ansolabehereStrengthIssuesUsing2008}, others contend that social transformations and life events can indeed lead to lasting shifts in people's opinions~\citep{lerschChangePersonalCulture2023}. Numerous studies demonstrate that both personal and public opinions evolve in response to salient social events and the framing of issues, whether these changes occur in the short term or reflect longer-term shifts~\citep{barrie2021sect, legewie2013terrorist, flores2018can}. Importantly, the meanings of questions are deeply shaped by their temporal context, which in turn influences survey responses. For example, when individuals are asked to name those with whom they discuss ``important matters,'' the term ``important matters'' may be interpreted as political issues if the survey is conducted during a heated presidential election debate period~\citep{lee2017important}. However, existing LLMs struggle to effectively account for these temporal heterogeneities, which limits their ability to accurately capture the dynamic nature of opinions and language over time.

Against this challenge, we incorporate period embeddings to consider temporal changes in the meaning of questions and individuals' belief systems~\citep{jooOutSyncOut2020,ruleLexicalShiftsSubstantive2015} as the third layer of neural embeddings. This accounts for the impact of temporal factors on survey responses, such as the gradual shift towards more progressive beliefs over time~\citep{baldassarriWasThereCulture2020} and the effects of specific events during a particular period, such as the macroeconomic changes, presidential elections, and the COVID-19 pandemic. We represent the historical features of survey periods as an N-dimensional embedding vector, which we refer to as the ``period embeddings'' of surveys, denoted by \(p\  \in \ R^{n}\). To generate period embeddings, we initially assign random latent features for each period, which are optimized such that two adjacent periods characterized by similar response patterns are located close to each other in the latent space during the training process. \cref{fig:figure3} shows two-dimensional $t$-SNE projections of the three embedding spaces, with points colored by $k$-means clusters for the question and respondent panels and labelled by survey year for the period panel.

\subsection*{Higher Order Interactions through Deep Cross Networks}

Finally, we incorporate the higher-order interactions between the three embeddings related to survey question meanings, individual beliefs, and historical contexts by employing a deep learning architecture called the ``Deep Cross Network (DCN)''~\citep{gordonJuryLearningIntegrating2022,wangDCNV2Improved2021}. The DCN comprises multiple cross layers designed to capture feature interactions~\citep{wangDCNV2Improved2021}. The ``cross-layer'' component enables the DCN to account for heterogeneous belief structures~\citep{baldassarriNeitherIdeologuesAgnostics2014}, where individuals can share a similar ``belief system'' not only by holding similar opinions, but also by exhibiting a consistent pattern in how their opinions relate to one another. First, the DCN learns interactions between question embeddings and individual embeddings, clustering respondents who respond similarly to semantically related questions. Second, the DCN models the interactions across different dimensions within individual embeddings, enabling certain dimensions of individual embeddings to act as ``moderators'' that shape how other dimensions influence predicted responses. For instance, as~\citep{baldassarriNeitherIdeologuesAgnostics2014} suggest, individuals' socio-economic status (SES) can moderate how people organize their beliefs, meaning that the influence of ideology on a particular attitude may vary depending on an individual's SES. Such interactions can be captured by the stack of cross-layers. The DCN contains feed-forward dense layers and a classifier head, which is responsible for producing the final prediction.

In the following example, we demonstrate how the DCN operates. Let's assume that \(s_{1}\) captures whether a question is about vaccine hesitancy, \(b_{1}\) captures whether an individual holds conservative ideology, and \(p_{1}\) captures the extent to which the COVID-19 pandemic is ongoing. If an individual's response to a vaccine mandate is influenced by their conservative ideology and the ongoing COVID-19 pandemic, we will need to consider the interaction \(s_{1} \times b_{1} \times p_{1}\) to predict the individual's response. To capture these interactions, we define cross layers in the DCN as follows. We begin by concatenating the question embedding of the survey question (\(s\)), the respondent embedding (\(b\)), and the temporal embedding (\(p\)) into a single vector, denoted as \(x_{0}\). The (l+1)th cross layer can then be defined as follows:

\[x_{l + 1} = x_{0}\  \odot \ (W_{l}x_{l} + b_{l}) + x_{l}\]

where \(x_{l}\  \in \ R^{3n}\) and \(x_{l + 1}\  \in \ R^{3n}\) are the input and output of the cross layers, respectively. \(W_{l}\  \in \ R^{3n \times 3n}\) and \(b_{l}\  \in \ R^{3n}\) are the learned weights and biases of the cross-layer. Each element in \(W_{l}\) captures the relative contribution of the interaction between features in the prediction. In the DCN with \emph{k} cross layers, the model includes all feature interactions up to a maximum polynomial order of \emph{k + 1}.

By concatenating multiple cross layers, it is possible to consider more complex interactions. For instance, the relationship between attitudes toward vaccine mandate and being a liberal might depend on another latent dimension, such as one's attitudes toward scientific knowledge. Complex feature crossings like \(s_{1} \times b_{1} \times b_{2} \times p_{1}\) can be captured by employing multiple cross layers. By considering complex interactions between the meaning of survey questions, individual beliefs, and survey periods, we can avoid the assumption that the meaning of opinions is identical for everyone at different times. This assumption has been challenged by previous research that shows the heterogeneous perception of cultural meanings across different socio-demographic backgrounds~\citep{goldbergMappingSharedUnderstandings2011,baldassarriNeitherIdeologuesAgnostics2014} and cognitive limitations~\citep{martinLifeBeachYou2010a}.

It is important to note that the DCN allows the model to make direct use of the text of their responses to all (non-left-out) survey items, rather than solely relying on the similarity of their responses to other respondents. As the respondent embeddings $b_i$ are trained via back-propagation, they are optimized to interact with features derived from question texts (i.e., question embeddings $s_i$) and contextual features (i.e., period embeddings $p_i$). In other words, $b_i$ is not learned in isolation or solely through similarity to other respondents' responses, as in methods like matrix factorization or PCA.

\cref{fig:figure_a_architecture_detail} provides additional information regarding the input and output dimensions of each layer. We first encode survey questions into a question embedding using Alpaca-7B, GPT-J-6B, or RoBERTa-large, while individual ID and survey year are respectively encoded into respondent and period embeddings. These embeddings are then concatenated and used as inputs to the DCN, which then captures the higher-order interactions between them and generates the predictions. During the training process, all embeddings and the DCN are jointly trained to predict an individual's answer to a specific question in a certain year.

\clearpage
\section*{Appendix C. Methodological details on training, aggregation, evaluation metrics}\label{appendix:methodological_details}
\label{sec:methodological_details}

\noindent \subsection*{Hyperparameters and training}

We first encode three inputs---survey questions, individual IDs, and survey years---into question, respondent, and period embeddings, and then concatenate and use them as inputs to the DCN to capture higher-order interactions within and across them. To implement our model architecture, we utilize Huggingface's API for incorporating LLMs (Alpaca-7B, GPT-J-6B, RoBERTa-large) and TensorFlow Recommenders (TFRS) for deploying the DCN. During the training process with the GSS data, all model components are jointly trained with the DCN. We freeze the pre-trained parameters of LLMs, except for an adaptable, feed-forward layer to convert the original contextual embedding to an N-dimensional trainable question embedding, to avoid the known over-fitting issues when jointly trained with other embeddings~\citep{gordonJuryLearningIntegrating2022}, which also helps keep computational costs at a manageable level. From our own experiments with RoBERTa-large, we found that models without freezing the parameters did not meaningfully improve the performance. We train our models using the Adam optimizer with a learning rate of $2\mathrm{e}{-5}$, and a binary cross-entropy is used as the loss function. We use a batch size of 128, and we limit the maximum sequence length to 150 tokens for the RoBERTa-large model. For the DCN-specific architecture, we experiment with various hyperparameters and choose a fixed embedding dimension of 50, three cross-layers of size 150, three feed-forward dense layers of size 150, and a final output layer of size 1 based on the previous literature and a grid search~\citep{gordonJuryLearningIntegrating2022}. The number of training epochs is determined based on the performance in validation data.

Our model architecture shares a similar design principle with other NLP models that aim to predict how different individuals label texts differently, such as the jury learning model~\citep{gordonJuryLearningIntegrating2022}. Simultaneously, our model architecture shares a similar goal with a group of models that use latent features to pinpoint key dimensions of beliefs underlying various opinions, such as principal component analysis (PCA) and the NOMINATE algorithm~\citep{jooOutSyncOut2020}. It is important to highlight that our model architecture does not impose any specific missing data mechanisms, such as Missing at Random (MAR) or Missing Completely at Random (MCAR). Instead, our model operates on the assumption that these latent factors influence responses, a concept similar to other machine learning models employed for deducing missing survey responses, like matrix factorization~\citep{senguptaSparseDataReconstruction2023}. A key difference is that our models assume multiple latent features, including sentence embeddings, interact together to shape responses, while the matrix factorization model does not.

\subsection*{Sigmoid aggregation and bias correction}

To generate the predicted probability that each individual provides a positive response to a particular question, we employ a sigmoid function, which generates the probability that falls within the range of 0 to 1 as logistic regression models do. To estimate public opinion, we aggregate these probabilities with survey weights. We find that averaging the sigmoid-transformed individual predictions can lead to an underestimation or overestimation of the true population proportion, especially when the true probabilities approach extreme values near $0$ or $1$. This bias arises due to the non-linearity of the sigmoid function, defined as $\sigma(x) = 1/(1 + e^{-x})$, and the implications of Jensen's Inequality, which asserts that $\sigma(\mathbb{E}[X]) \neq \mathbb{E}[\sigma(X)]$. To correct this bias, we adjust the aggregated outcome by modeling the target observation using the equation $\sigma(\beta_0 + \beta_1 \cdot \text{aggregated prediction})$. This approach accounts for the non-linear transformation at the aggregate level, ensuring that the expected mean response is accurately estimated without the bias introduced by directly averaging individual predictions. Finally, we predict counterfactual trends using aggregated responses from retrodiction models. To estimate the direction and magnitude of over-time trends, we employ a Generalized Additive Model (GAM), a flexible, non-parametric regression technique capable of modeling complex, non-linear relationships between variables~\citep{hastieGeneralizedAdditiveModels1992}.

\subsection*{Evaluation metrics}

\subsubsection*{Personal opinion prediction}

We evaluate model performance using the \textit{Area Under the Curve (AUC)}, a metric that measures the ability of the model to distinguish between positive and negative responses. The AUC is computed as the area under the Receiver Operating Characteristic (ROC) curve, which plots the \textit{True Positive Rate (TPR)} against the \textit{False Positive Rate (FPR)} across various classification thresholds:
\[
\text{AUC} = \int_{0}^{1} \text{TPR}(c) \, d(\text{FPR}(c)),
\]
where $\text{TPR}(c) = \text{TP}/(\text{TP}+\text{FN})$ (sensitivity / recall), $\text{FPR}(c) = \text{FP}/(\text{FP}+\text{TN})$ (the proportion of negatives incorrectly classified as positives), and $c$ is the classification threshold. AUC quantifies the probability that the model will correctly rank a randomly selected positive response higher than a randomly selected negative response. The AUC value ranges from 0.5 (no discriminative ability) to 1.0 (perfect classification). Unlike other metrics, AUC does not require setting an arbitrary threshold for binarizing predicted probabilities, making it particularly advantageous for tasks with imbalanced response distributions. We also assess accuracy and F1 score and find similar results, but we do not present them in the main manuscript because they require an arbitrary threshold for binarizing continuous predicted probabilities, which can be problematic when responses are imbalanced. The selection of metrics is informed by prior work on the predictability of behaviors and life outcomes~\citep{salganikMeasuringPredictabilityLife2020,savcisens2024using}.

We calculate AUC at both the opinion and individual levels without incorporating survey weights, as a preliminary step to understanding the model's baseline predictive performance. We assess which respondents are inherently more predictable at the individual level and which survey questions are more consistently predictable across the population at the opinion level. By modeling predictability as a function of individual traits or opinion characteristics, we systematically explore the factors that contribute to the predictability of individuals and opinions. Conducting this analysis without survey weights ensures that our findings are driven purely by the model's internal dynamics, providing a clean foundation for subsequent weighted analyses that account for sampling probabilities.

\subsubsection*{Public opinion prediction}

For evaluating the accuracy of public opinion predictions, we employ three complementary metrics. First, we measure \textit{Pearson correlations} between observed and predicted mean responses across all available years. For each opinion $i$:
\[
r_{i} = \frac{\sum_{t=1}^{T} (y_{i,t} - \bar{y}_{i})(\hat{y}_{i,t} - \bar{\hat{y}}_{i})}{\sqrt{\sum_{t=1}^{T} (y_{i,t} - \bar{y}_{i})^2 \sum_{t=1}^{T} (\hat{y}_{i,t} - \bar{\hat{y}}_{i})^2}}
\]
where $y_{i,t}$ and $\hat{y}_{i,t}$ are the observed and predicted mean responses for opinion $i$ at time $t$, $\bar{y}_{i}$ and $\bar{\hat{y}}_{i}$ are their averages over all $T$ time points. Higher correlations indicate stronger linear association between predicted and observed proportions.

Second, we calculate the \textit{mean absolute error} (MAE) to quantify the average deviation of predictions from observed targets on an absolute scale:
\[
\text{MAE}_{i} = \frac{1}{T} \sum_{t=1}^{T} \lvert y_{i,t} - \hat{y}_{i,t} \rvert.
\]
MAE highlights how closely predicted proportions align with the actual responses and is particularly useful for assessing practical reliability.

Third, we report \textit{percent prediction accuracy}: the proportion of survey opinions for which the predicted mean response falls within a predefined margin of error of the observed mean response (margins from 1\% to 10\%). At a 5\% margin of error, a prediction is considered accurate if $\lvert \hat{y}_{i,t} - y_{i,t} \rvert \leq 0.05$. This metric identifies the range of tolerances within which the model provides reliable predictions, offering practical insights into its applicability for different research contexts.

\clearpage
\section*{Appendix D. MCAR, MAR, and MNAR simulation}\label{appendix:mcar_mar_mnar}

First, MCAR assumes that $M$ is completely unrelated to the data $X$ -- unrelated to both $X_{\text{observed}}$ and $X_{\text{unobserved}}$. In other words, $p(M = 1 \mid X, \xi) = p(M = 1 \mid \xi)$. To simulate missing data based on MCAR, we randomly selected $10\%$ of observed values and removed them. Second, MAR assumes that $M$ depends on the observed values of other variables but not on unobserved variables. Therefore, $p(M = 1 \mid X, \xi) = p(M = 1 \mid X_{\text{observed}}, \xi)$. To simulate missing data based on MAR, we adopt the method proposed by~\citet{senguptaSparseDataReconstruction2023} that fits a logistic regression model predicting $m_{i, j}$ using the values of other observed variables. Specifically, we used the observed variables that rarely have missing values (less than $10\%$) to predict $m_{i, j}$. Then, we remove the $10\%$ of values with the highest probability of being missing based on the regression. Third, MNAR assumes that $M$ depends on both observed and unobserved data. Thus, $p(M = 1 \mid X, \xi) = p(M = 1 \mid X_{\text{observed}}, X_{\text{unobserved}}, \xi)$. To simulate missing data based on MNAR, we fit a logistic regression model that predicts $m_{i, j}$ using demographic variables (i.e., age, cohort, gender, race, education, income, and religion). Since demographic variables have not been used by the missing data imputation model in our main results, we assume that these variables are unobserved and use these variables to generate missing values based on MNAR.

\clearpage
\section*{Appendix E. Alternative Benchmarks}\label{appendix:alt_benchmarks}

\subsection*{Matrix factorization (MF)}

To establish a benchmark model for comparison, we train a matrix factorization model, which is a well-established technique for solving matrix completion problems by estimating missing values based on available elements. This involves decomposing a matrix with known elements into two lower-dimensional matrices. By optimizing the dot products of these matrices to align with the known elements of the original matrix, accurate predictions for the missing values can be obtained. It achieves high accuracy comparable to deep learning models, making it a popular choice in recommender system development \citep{koren_matrix_2009}. Matrix factorization also outperforms traditional imputation methods, such as Amelia and MICE, in filling missing data in sparse survey datasets \citep{senguptaSparseDataReconstruction2023}.

First, we create a matrix \(S\  \in \ R^{68,846 \times 3,699}\) of survey responses. Each row represents an individual, each column represents a survey question, and each element indicates the individual's binarized response (0 = negative and 1 = positive, see \cref{tab:binary_transformation}). Using this matrix \(S\), we train two lower-dimensional matrices: \(I\  \in \ R^{68,846 \times 50}\) and \(Q\  \in \ R^{50 \times 3,699}\). The objective is to optimize \(I\) and \(Q\) such that the dot product of the corresponding rows and columns closely matches the available values in \(S\). Specifically, we minimize the squared difference \((I_{i \cdot}Q_{\cdot q}-S_{iq})^{2}\) for the available values in  \(S\). To optimize \(I\) and \(Q\), we employ the alternating least squares (ALS) method, following the previous study on the GSS 2014 data \citep{senguptaSparseDataReconstruction2023}. We conduct ALS for 15 iterations, applying a regularization penalty of \(\lambda=10\) to prevent overfitting. Using the learned latent factors from the matrix factorization model optimized throughout this process, we predict the missing response of an individual \(i\) for a given variable \(q\) by taking the dot product of the corresponding vectors: \(I_{i \cdot}\) and \(Q_{\cdot q}\).
\subsection*{Multivariate Imputation by Chained Equations (MICE)}%

We implemented MICE following van Buuren's logistic-regression imputation algorithm for binary targets \citep{vanbuuren2018flexible}, where each missing cell is filled in by a logistic regression on the other (observed) variables and the missing values are then iteratively re-imputed across columns until the imputations stabilize. For each target variable, we used the top-$40$ most correlated predictors (selected once on the observed data and held fixed across $15$ chained-equations iterations), drew $m=5$ multiple imputations per missing cell \citep{rubin1987multiple}, and averaged across imputations. The per-column logistic regressions were fit only on training cells, so no validation labels entered the regression coefficients.
\subsection*{Text-aware matrix factorization (MF + TF-IDF, MF + SentenceBERT)}

\input{Snippets/snippet_text_aware_mf_explanation.tex}

\subsection*{Prompted out-of-shelf LLMs (GPT-4o, Gemini-2.5-flash)}

We benchmark our model against GPT-4o and Gemini-2.5-flash on the 2018 GSS data under three prompting strategies, following \citep{argyleOutOneMany2023}: (i) a \emph{demographics-only} prompt that includes 15 respondent characteristics (year, ideology, party affiliation, race, sex, family income, age, birth year, education, region, urbanicity, ethnic background, work status, marital status, religion); (ii) a \emph{random-$k{=}10$ context} prompt that adds the respondent's binarized responses to 10 randomly selected GSS items on top of (i); and (iii) a \emph{correlation-top-$k{=}10$ context} prompt that instead adds the 10 items whose responses correlate most strongly with the target variable in the training data. We set $k=10$ to match the context size typically used in prior work and to keep inference costs practical. This benchmark establishes whether feature-based transfer learning offers any advantage over prompting a more capable LLM with the same input.%
\subsection*{Per-variable time-series regression (population-level)}

For public opinion prediction we add a different family: simple per-variable regressions of the population mean against time. The intuition is that for a question asked across multiple years, one can fit a line to its mean responses (i.e., agreement rate) over time and predict the missing year. For each GSS variable $v$, let $\mathcal{T}_v$ be the set of training years available for $v$ in a given fold, and let $p_{v,t}\in[0,1]$ be the observed year-level mean response in year $t$. We fit a regression to $\{(t,p_{v,t}):t\in\mathcal{T}_v\}$ and predict $\hat{p}_{v,t^{*}}$ for each held-out year $t^{*}\notin\mathcal{T}_v$. With $\sigma(z)=1/(1+e^{-z})$ denoting the standard logistic function, we consider four variants of increasing flexibility:
\begin{itemize}
  \item \textit{Linear OLS} --- the simplest possible trend line: $\hat{p}_{v,t^{*}}=\hat{\beta}_{0,v}+\hat{\beta}_{1,v}\,t^{*}$, fit by ordinary least squares.
  \item \textit{Logistic GLM} --- the same trend line, but bounded: a Binomial GLM with logit link, $\mathrm{logit}(p_{v,t})=\beta_{0,v}+\beta_{1,v}\,t$, with predictions recovered on $[0,1]$ as $\hat{p}_{v,t^{*}}=\sigma(\hat{\beta}_{0,v}+\hat{\beta}_{1,v}\,t^{*})$. Enforces the $[0,1]$ bound that Linear OLS can violate near the edges.
  \item \textit{Polynomial OLS} --- a curved trajectory: $\hat{p}_{v,t^{*}}=\hat{\beta}_{0,v}+\hat{\beta}_{1,v}\,t^{*}+\hat{\beta}_{2,v}\,(t^{*})^{2}$, fit by ordinary least squares with a quadratic year term so the trajectory can capture non-monotonic change.
  \item \textit{Logistic polynomial GLM} --- both at once: a Binomial GLM with logit link applied to the same quadratic predictor, $\hat{p}_{v,t^{*}}=\sigma(\hat{\beta}_{0,v}+\hat{\beta}_{1,v}\,t^{*}+\hat{\beta}_{2,v}\,(t^{*})^{2})$. Combines the $[0,1]$ bound with curvature.
\end{itemize}
When a variable has fewer than two observed training years in a fold, the linear and logistic variants are not identifiable; we therefore fall back to an intercept-only model (i.e., variable's mean). The polynomial variants additionally require at least three training years; with fewer, the polynomial fit reverts to the linear fit, and with fewer than two we fall back to the intercept-only model (i.e., training-year mean). These baselines enable us to contextualize our model's performance, particularly in scenarios where data sparsity or complex temporal trends challenge public opinion prediction.

\clearpage
\section*{Appendix F. Sensitivity analyses and detailed heterogeneity results}\label{appendix:sensitivity}
\label{sec:sensitivity_heterogeneity}

\subsection*{Methodology for between-/within-group variance evaluation}

\noindent Recent research highlights a critical limitation of LLMs in their role as synthetic survey respondents: the tendency to underestimate variation across demographic and political groups~\citep{bisbee2024synthetic}. While LLMs may be effective at predicting overall mean responses, they often fail to capture the broad spectrum of human diversity. This limitation is especially problematic for social science research because its goal is often to understand how different groups and individuals think about various social issues differently. Capturing these variations---both between groups and within groups---is essential for generating insights that reflect the complexities of human opinion. Recognizing the importance of this challenge, we evaluate our model's capacity to accurately represent \textit{between-group} and \textit{within-group} variance. Specifically, we compare the predicted variance produced by our model against the observed variance derived from the GSS dataset. This assessment allows us to examine the extent to which the model captures the nuanced differences in opinions across and within demographic groups, a critical measure of its utility in social science applications.

To assess between-group variance, we estimate the standard deviation of predicted and observed group means across key demographic categories: gender (Male, Female), race (White, Black, Other), age groups (18--29, 30--44, 45--59, 60--74, 75--89), education levels ($<$HS, HS Graduates, Some College, BS, MS+), and political stance (Extremely Liberal, Liberal, Slightly Liberal, Neither Liberal Nor Conservative, Slightly Conservative, Conservative, Extremely Conservative). For each category, we calculate the average predicted response for each subgroup. The standard deviation across these group-specific means serves as the measure of between-group variance (e.g., white vs.\ black vs.\ others), where higher values suggest greater divergence among groups and lower values indicate more similarity. To evaluate within-group variance, we calculate the standard deviations of individual responses within each demographic subgroup (e.g., within white). Higher within-group standard deviations indicate greater diversity of opinions within the subgroup, reflecting heterogeneous perspectives, while lower values suggest a more homogenous outlook within the group.

Our results indicate that our model could successfully capture between-group variations, particularly for political stance and education (\cref{fig:between_group_variance}), while the within-group analysis shows relatively strong correspondence across demographic groupings as well as an overall (``All'') reference panel (\cref{fig:within_group_variance}). \cref{fig:between_group_variance,fig:within_group_variance} report the Spearman correlations $\rho$ and mean absolute errors (MAEs) summarizing agreement between predicted and observed subgroup dispersion for Age, Gender, Race, Political stance, and Education. The within-group comparison uses binarized predictions (threshold 0.5) to match the scale of the observed binary responses.

These findings indicate that our model can effectively predict between-group variance for key social categories, particularly political stance and education. However, it underestimates within-group variance across most demographic categories~\citep{bisbee2024synthetic}. This evaluation highlights our model's strengths in capturing systematic group-level differences while suggesting areas for improvement in modeling more nuanced, within-group variations.

\subsection*{Detailed individual-level heterogeneity}

To assess between-group gaps in individual-level AUC, we estimate OLS regressions with HC2 robust standard errors. The predictors are sex, race, education (degree), region, income quintile, age group, party identification (7-point), and survey year (continuous, scaled per $+10$ years). To absorb the changing composition of survey items across waves, we control for two variables: $n_q$, the number of items each respondent answered, and $\overline{|p_v - 0.5|}$, the mean class imbalance of items being asked in the years, where $p_v$ is the marginal binarized rate of variable $v$.

Absolute predictive accuracy is consistently lower for racial minorities, lower-SES respondents, and non-partisans, mirroring their underrepresentation in the GSS sample and in large-scale language-model pretraining corpora. Our Alpaca-7B model nevertheless improves AUC over the matrix factorization baseline for every subgroup on both the missing data imputation and retrodiction tasks, with similar magnitudes of improvement across groups. The absolute-accuracy gap is therefore largely a property of the prediction problem itself---fewer training cases for marginalized groups or lower correlations among variables within certain groups---rather than a property of our method relative to alternatives, and the framework should be understood as improving representativeness for these groups compared to existing imputation methods, even where absolute performance for them remains lower than for majority groups.

\subsection*{Predictability and ideological correlation}

\paragraph{Method.} For every binarized GSS variable $v$, we compute the per-variable Spearman correlation $\rho_v(\text{polviews})$ between the binarized response and the 7-point political-ideology scale (\texttt{polviews}: 1 = extremely liberal, 4 = moderate, 7 = extremely conservative), pooled across all respondents who answered both items. We use the absolute value $|\rho_v(\text{polviews})|$ as the variable-level measure of ``ideological loading''---larger values indicate that the response splits more strongly along the liberal--conservative axis. We then assess how this ideological loading relates to predictability by computing a Spearman rank correlation, across all binarized GSS variables, between $|\rho_v(\text{polviews})|$ and our Alpaca-7B retrodiction AUC for that variable.

\paragraph{Results.} The Spearman rank correlation between $|\rho_v(\text{polviews})|$ and Alpaca-7B retrodiction AUC is $0.225$, meaning that an item being more correlated with political ideology is associated with better predictability, but the relationship is somewhat modest---only about $5\%$ of the variance in AUC is explained by ideological loading alone. Substantively interesting counter-examples sit on both sides. There are variables with $|\rho_v(\text{polviews})| < 0.10$ that the model nevertheless predicts well (AUC $\geq 0.85$): items about religious practice (frequency of private prayer, desire to be closer to God, the belief that God forgives me), traditional gender roles (women as caretakers of the home), tolerance of disliked outgroups (allowing a militarist or socialist to teach), and beliefs about suicide. Conversely, there are variables with $|\rho_v(\text{polviews})| \geq 0.30$ that the model predicts \emph{poorly} (AUC $< 0.65$), including questions about whether police treat racial groups differently, a self-placement scale on left--right politics, attitudes about defunding the police, feelings about wealth inequality, and trust in news media---items that are highly ideologically loaded but apparently rely on respondent dispositions that are not well captured by the rest of the GSS battery. \cref{tab:polviews_rank} lists the top-40 most ideologically correlated and bottom-40 least correlated GSS variables, each paired with their retrodiction AUC.

\paragraph{Interpretation.} The model is partly learning partisan sorting---ideologically polarized items are easier on average---but it is also recovering coherent non-political structure: religious commitment, traditional vs.\ progressive gender attitudes, tolerance of outgroups, and other latent dispositions are organized enough to be predictable from a respondent's answers to other GSS items even when those dispositions do not project onto the liberal--conservative axis. Conversely, items that \emph{are} ideologically loaded but require very specific group experiences or volatile attentional triggers (e.g., trust in the news media) remain difficult to predict. In short, the method may work better for opinion domains with coherent latent structure---ideological, religious, value-based, or otherwise---while its performance may be more limited for highly idiosyncratic, low-salience, or experience-specific items.

\subsection*{Detailed opinion-level heterogeneity}

One might question whether our models exhibited a different performance in 2021 when the survey response rate was 17\%, the lowest in the history of GSS (the average response rate in the GSS was 72\%), and the survey mode was altered due to the COVID-19 pandemic. However, the opinion-level regression shows that the 2021 indicator is significantly positive only for missing data imputation; retrodiction and unasked opinion prediction show no significant difference.

It may not be surprising that the larger sample size (i.e., the number of respondents who answer a specific survey question in the training data) is associated with the larger AUC, though this sample size effect is negligible in unasked opinion prediction. Opinions with higher response rates show higher AUCs, though they are not significant for unasked opinion prediction. Surveys that use split ballot designs show better performance, encouraging more active use of split ballot designs in surveys. One might question whether predicting opinions with multiple response options, given our binarization method, is harder than those with just two options. However, we find minor effects, only in the case of retrodiction models. The proportion of non-responses, such as refusals or inapplicable responses, is associated with the lower AUCs, highlighting the challenge of predicting opinions with fewer responses due to systematic non-responses.

Additionally, one may be concerned that our binarization procedure might disadvantage survey responses with an odd number of categories, as it requires assigning middle categories to either the positive or negative group. To address this, we examined the relationship between the number of response categories and predictive accuracy, analyzing items with both even and odd numbers of categories. \cref{fig:mean_auc_response_categories} provides a detailed breakdown of the mean AUC by the number of response categories. Our analysis found no systematic differences in predictive accuracy between these two groups. While predictive accuracy generally decreases as the number of response categories increases with the exception of items with the highest number of response categories (i.e., \textit{relactiv}, \textit{trlegis}, \textit{trmedia}, \textit{trbusind}, \textit{trresrch}, and \textit{prayfreq}). This suggests that while the model captures broad patterns of agreement or disagreement effectively, it is less precise in differentiating between gradations of attitudes.

On the other hand, an AUC is smaller when the survey response shows a larger variance, indicating that it is harder to predict a controversial opinion. To further explore the model's performance on rare opinions---situations where a small percentage of respondents hold a particular view---we analyzed the mean AUC across different ranges of positive response rates. \cref{fig:auc_positive_responses} illustrates this relationship, showing the mean AUC at varying percentages of positive responses along with error bars representing standard errors. Our analysis reveals a non-linear relationship between the AUC and the proportion of positive responses. Interestingly, the model's AUC is higher at the extreme ends of the opinion spectrum, specifically when the positive response rate is relatively low or high. This indicates that the model performs better when predicting opinions that are either widely accepted or largely rejected by the population, possibly due to greater predictive clarity or distinct feature patterns associated with these extremes. In the intermediate ranges of positive responses, the AUC slightly decreases or stabilizes, suggesting a modest reduction in predictive accuracy when opinions are more polarized.

Additionally, in the retrodiction task the model shows better predictive performance for opinions that are closely located to others in the embedding space, as measured by the average cosine similarity with all other opinions, even though the effect size is small. This suggests that the model can accurately predict opinions not only when they are semantically similar but also when they deviate from other opinions in the sentence embedding space. These findings indicate that the model performs better at predicting opinions that are more central within the embedding space. These centrally located opinions are surrounded by thematically or structurally related beliefs, forming dense networks of interconnected beliefs across various domains. In contrast, opinions on the periphery of the embedding space lack such connections, offering fewer cues for the model and leading to lower predictive accuracy. This underscores the importance of cultural belief systems and their interconnectedness for understanding which opinions are more predictable.

\clearpage
\section*{Appendix G. Practical Considerations}

\input{Snippets/snippet_practical_considerations.tex}

\clearpage

\section*{Appendix H. Embedding Representations and Temporal Stability}
\label{sec:embedding_stability}

A recent line of work raises concerns about the stability and interpretability of cosine similarity in language-model embedding spaces, especially across contexts and over time~\citep{kindel2024cosine}. We engage with the critique on two levels: the specific concern raised about cosine similarity, and the broader question of embedding stability across decades.

\paragraph{The scope of the cosine-similarity critique.}
\citet{kindel2024cosine} focuses on a specific measurement strategy in which two latent concepts are each represented as sets of embedding vectors (typically derived from keyword lists), and their association is inferred from pairwise cosine similarities between vectors across the two sets, often aggregated into a mean cosine similarity. The paper shows that such between-set cosine similarities are constrained by within-set anisotropy induced by keyword selection. However, our model does not use cosine similarity between embeddings as the basis for producing predictions, nor is it a linear or cosine-similarity-based model. Instead, our Alpaca-7B question embedding (passed through a trainable linear projection), along with respondent and period embeddings, are used as inputs to a deep neural architecture trained end-to-end. These representations are optimized for predictive performance via the training objective (i.e., accurately predicting individuals' responses to survey questions in particular years) and are processed through cross layers and feed-forward layers that learn higher-order interactions among them to produce the predicted response. If our model were a simple linear regression that directly relied on distances or cosine similarities in embedding space, then the stability and interpretability concerns raised by \citet{kindel2024cosine} would directly apply. Because predictions are produced by a deep, non-linear function of the three embeddings, the model does not rely on cosine similarity in the predictive pipeline. The embeddings are therefore best understood as \emph{functionally useful representations for prediction}. While all embeddings and downstream components are trained for predictive accuracy, cosine similarities computed on the learned representations should still be interpreted with the caution emphasized by \citet{kindel2024cosine}.

\paragraph{Representation of meaning across decades.}
A separate concern is whether question representations remain meaningful across decades given that social meanings evolve. We aimed to address this concern when designing the architecture of the model. When the model predicts a response to a question such as ``Should same-sex couples have the right to marry?'' in a given year, it does not rely on a fixed, time-invariant embedding of that question. Instead, it combines the question embedding with a respondent embedding and a year-specific \emph{period embedding}, and cross-layer interactions jointly condition the prediction on all three components. As a result, the same question posed in different years (e.g., 1990 versus 2010) is processed through different period embeddings, yielding context-dependent representations that allow the model to capture temporal shifts in meaning and population attitudes.

\clearpage
\section*{Appendix I. Fine-tuning of the LLM Backbone}
\label{sec:app:lora-ft}
\input{Snippets/snippet_lora_appendix.tex}

\clearpage
\section*{Appendix Tables}

\input{Tables/table_binary_transformation.tex}
\clearpage
\input{Tables/table_question_distribution.tex}
\clearpage

\input{Tables/table_alpaca_text_examples.tex}
\clearpage

\input{Tables/table_similar_questions.tex}
\clearpage

\input{Tables/table_top20_modules.tex}
\clearpage

\input{Tables/table_modelcomparison.tex}
\clearpage
\input{Tables/table_retro_interp_forecast.tex}
\clearpage
\input{Tables/table_retro_distance_metrics.tex}
\clearpage
\input{Tables/table_featureimportance.tex}
\clearpage
\input{Tables/table_accuracy_thresholds.tex}
\clearpage
\input{Tables/table_samesex_50.tex}
\clearpage
\input{Tables/table_counterfactual_roper_surveys.tex}
\clearpage
\input{Tables/table_samesex_framing.tex}
\clearpage
\input{Tables/table_hybrid_mf.tex}
\clearpage
\input{Tables/table_frontier_llm.tex}
\clearpage
\input{Tables/table_perfbygroup.tex}
\clearpage
\input{Tables/table_polviews_rank.tex}
\clearpage
\input{Tables/table_subgroup_delta_imputation.tex}
\clearpage
\input{Tables/table_subgroup_delta_retrodiction.tex}
\clearpage

\section*{Appendix Figures}

\input{Figures/figure_a_variable_selection_block.tex}
\clearpage
\input{Figures/figure_a_embeddings_block.tex}
\clearpage
\input{Figures/figure_a_architecture_detail_block.tex}
\clearpage
\input{Figures/figure_a_demographic_cv_block.tex}
\clearpage
\input{Figures/figure_a_cv_scheme_block.tex}
\clearpage
\input{Figures/figure_a_regime_schematic_block.tex}
\clearpage
\input{Figures/figure_a_missing_mech_block.tex}
\clearpage
\input{Figures/figure_a_missing_prop_block.tex}
\clearpage
\input{Figures/figure_a_retro_interp_forecast_block.tex}
\clearpage
\input{Figures/figure_a_distance_nearest_year_dummy_block.tex}
\clearpage
\input{Figures/figure_a_module_removed_block.tex}
\clearpage
\input{Figures/figure_a_samesex_exclude_block.tex}
\clearpage
\input{Figures/figure_a_groupvar_between_block.tex}
\clearpage
\input{Figures/figure_a_groupvar_within_block.tex}
\clearpage
\input{Figures/figure_a_polviews_rank_block.tex}
\clearpage
\input{Figures/figure_a_response_category_block.tex}
\clearpage
\input{Figures/figure_a_auc_by_agreement_block.tex}
\clearpage
\input{Figures/figure_a_n_questions_block.tex}
\clearpage
\input{Figures/fig_figure_years_vs_auc_block.tex}
\clearpage
\input{Figures/figure_a_individualauc_compare_block.tex}
\clearpage
\input{Figures/figure_a_ideology_corr_time_block.tex}
\clearpage
\input{Figures/figure_ft_effect_block.tex}
\clearpage

%% file: Snippets/snippet_text_aware_mf_explanation.tex
We extend the matrix-factorization baseline to incorporate question text using either TF-IDF features or SentenceBERT (all-MiniLM-L6-v2) embeddings. Specifically, we implement the collective matrix factorization model of \citet{singh2008collective}, which jointly factorizes (i) the respondent-by-question response matrix $X \in \mathbb{R}^{m \times n}$ and (ii) the question-by-text feature matrix $Y \in \mathbb{R}^{n \times d}$ by sharing a common latent representation for questions, $V \in \mathbb{R}^{n \times k}$. Intuitively, this allows information from question text to influence prediction: questions with similar wording are encouraged to have similar latent representations, so that responses to one question can help predict responses to another.

Here, $m$ denotes the number of respondents, $n$ the number of questions, $d$ the dimensionality of textual features, and $k$ the latent dimension (we set $k = 50$). Each entry $X_{ij} \in \{0,1\}$ indicates respondent $i$'s binarized response to question $j$ when observed, and is masked otherwise. Each row $Y_{j,:}$ represents the textual features of question $j$: under the TF-IDF variant, this is a bag-of-words representation over the vocabulary; under the SentenceBERT variant, it is a 384-dimensional sentence embedding from all-MiniLM-L6-v2.

The model assumes $X \approx UV^\top$ and $Y \approx VZ^\top$, where $U \in \mathbb{R}^{m \times k}$ contains respondent embeddings, $V \in \mathbb{R}^{n \times k}$ contains question embeddings, and $Z \in \mathbb{R}^{d \times k}$ maps textual features into the shared latent space. The same question representation $V$ is shared across both the response matrix and the text features, so that similar question texts induce similar latent factors, which in turn affect predicted responses $UV^\top$. We optimize the joint objective
\begin{equation}
\mathcal{L} \;=\; \alpha \,\bigl\| W \odot (X - UV^\top) \bigr\|_F^2 \;+\; (1-\alpha)\, \bigl\| Y - VZ^\top \bigr\|_F^2 \;+\; \lambda \bigl(\|U\|_F^2 + \|V\|_F^2 + \|Z\|_F^2\bigr),
\end{equation}
where $W$ masks unobserved entries in $X$, and we set $\alpha = 0.5$ to weight the two reconstruction terms equally.

%% file: Snippets/snippet_practical_considerations.tex
\noindent\textit{(1) Time and hardware required to train.} A single training run for one scenario (missing-data imputation, retrodiction, or unasked-opinion prediction) requires roughly 12 GPU-hours on a single NVIDIA T4 (16~GB). Inference can be performed without a GPU on a typical consumer desktop or cloud platforms such as Google Colab, taking only a few minutes per survey year.\\

\noindent\textit{(2) Approximate cost.} Using the Azure on-demand rate for a \texttt{Standard\_NC4as\_T4\_v3} instance (one NVIDIA T4 GPU, 4 vCPU, 28~GiB RAM; $\sim$\$0.53/hr in the US-East region at the time of writing), a single training run costs approximately \$6--7. Inference and downstream analysis on CPU incur negligible additional cost.\\

\noindent\textit{(3) When this approach adds the most value.}
\begin{itemize}
  \item \textbf{Response-level missing data:} For a moderately complete matrix with scattered missing cells, conventional methods such as MICE or matrix factorization (MF) are adequate and familiar.
  \item \textbf{Year-level missing data (retrodiction):} For predicting a variable's responses in years when the GSS did not field it, Alpaca-7B clearly outperforms MICE and MF. Within this setting, time-series regression baselines perform comparably to Alpaca-7B when the variable has many observed training years, a low-volatility trajectory, and the held-out year is close to a training year. For sparsely observed, highly volatile variables, or held-out years far ($\geq 3$ years) from the nearest training year, Alpaca-7B has a clear advantage.
  \item \textbf{Unasked opinion prediction:} For variables never measured by the GSS, only our method is applicable, as MICE, MF, and per-variable time-series regressions all require some prior observations of the target variable.
\end{itemize}

%% file: Snippets/snippet_lora_appendix.tex

To assess whether updating LLM parameters would meaningfully change our results, we conducted a preliminary analysis in which the LLM backbone was \emph{not} held frozen during training. Following popular parameter-efficient fine-tuning practice, we used LoRA (Low-Rank Adaptation) \citep{hu2022lora} in two configurations: (i) \emph{LoRA $q,v$ fine-tuning}, which fine-tunes the query and value projection matrices of transformer self-attention blocks, following the original implementation proposed by \citet{hu2022lora}; and (ii) \emph{LoRA all-linear fine-tuning}, which fine-tunes every linear module within the transformer blocks (i.e., the query, key, value, output, and feed-forward projections), as recommended by subsequent work \citep{dettmers2023qlora}.

Specifically, we trained our model as described in the \emph{Methods: Model Training} section, but without freezing the LLM weights throughout the entire training process. Instead, after two initial epochs in which the LLM weights were frozen, we unfroze the LLM backbone and jointly updated it together with the rest of the model over 200{,}000 training rows (respondent, question, year, response) using LoRA techniques. The LLM was then refrozen for the remainder of the training. This unfreeze-then-refreeze technique follows prior fine-tuning practice designed to guard against overfitting \citep{gordonJuryLearningIntegrating2022}.

Across missing data imputation, retrodiction, and unasked opinion prediction, both LoRA configurations yielded marginally higher (or near-identical) AUC and accuracy than the fully frozen baseline, but consistently lower F-1 scores (Figure~\ref{fig:ft-effect}). Specifically, baseline AUC values of $0.8681$, $0.8287$, and $0.6707$ for the three tasks shifted to $0.8688$, $0.8288$, and $0.6718$ under \emph{LoRA $q,v$} and to $0.8686$, $0.8286$, and $0.6775$ under \emph{LoRA all-linear}. Baseline accuracy values of $0.7812$, $0.7394$, and $0.6196$ shifted to $0.7811$, $0.7394$, and $0.6254$ under \emph{LoRA $q,v$} and to $0.7819$, $0.7384$, and $0.6272$ under \emph{LoRA all-linear}. Finally, baseline F-1 scores of $0.7789$, $0.7382$, and $0.6417$ shifted to $0.7776$, $0.7330$, and $0.6401$ under \emph{LoRA $q,v$} and to $0.7814$, $0.7300$, and $0.6214$ under \emph{LoRA all-linear}, respectively---representing a maximum F-1 decline of $0.020$ in unasked opinion prediction under \emph{LoRA all-linear}.

These results suggest that feature-based transfer learning already performs comparably to models in which the LLM backbone is jointly updated. However, we leave a more systematic comparison---including alternative fine-tuning strategies and larger adaptation budgets---for future work.

%% file: Tables/table_binary_transformation.tex
\begin{longtable}{r p{9cm} r p{3cm}}
\caption{Binary Transformation of Survey Response Options: Top 50 responses.} \\
\toprule
\textbf{Rank} & \textbf{Response Options} & \textbf{N} & \textbf{Binarized} \\
\midrule
\endfirsthead
\multicolumn{4}{c}{\textit{(continued from previous page)}} \\
\toprule
\textbf{Rank} & \textbf{Response Options} & \textbf{N} & \textbf{Binarized} \\
\midrule
\endhead
\midrule
\multicolumn{4}{r}{\textit{(continued on next page)}} \\
\endfoot
\bottomrule
\endlastfoot
1 & yes, no & 748 & 1, 0 \\
2 & strongly agree, agree, neither agree nor disagree, disagree, strongly disagree & 211 & 1, 1, 0, 0, 0 \\
3 & strongly agree, agree, disagree, strongly disagree & 106 & 1, 1, 0, 0 \\
4 & too little, about right, too much & 32 & 1, 0, 0 \\
5 & true, false & 31 & 1, 0 \\
6 & strongly agree, agree, not agree/dsagre, disagree, strong disagree & 29 & 1, 1, 0, 0, 0 \\
7 & strongly agree, agree, neither, disagree, strongly disagree & 29 & 1, 1, 0, 0, 0 \\
8 & agree, disagree & 29 & 1, 0 \\
9 & agree strongly, agree, neither agree nor disagree, disagree, disagree strongly & 26 & 1, 1, 0, 0, 0 \\
10 & strongly favor, favor, neither favor nor oppose, oppose, strongly oppose & 25 & 1, 1, 0, 0, 0 \\
11 & strongly agree, agree, neither agree nor disagree, disagree, strongly disagree & 24 & 0, 0, 0, 1, 1 \\
12 & not at all, 1 or 2 times, 3-5 times, 6 or more times & 23 & 0, 1, 1, 1 \\
13 & never, 1-2 times, 3-5 times, more than 5 times & 20 & 0, 1, 1, 1 \\
14 & definitely allowed, probably allowed, prob not allowed, defint. not allowed & 20 & 1, 1, 0, 0 \\
15 & very likely, somewhat likely, not very likely, not at all likely & 20 & 1, 1, 0, 0 \\
16 & definitely should, probably should, probably should not, definitely should not & 18 & 1, 1, 0, 0 \\
17 & unimportant, , , , , , very important & 18 & 0, 0, 0, 0, 1, 1, 1 \\
18 & very true, somewhat true, not too true, not at all true & 18 & 1, 1, 0, 0 \\
19 & like very much, like it, mixed feelings, dislike it, dislike very much & 18 & 1, 1, 0, 0, 0 \\
20 & very likely, somewhat likely, somewhat unlikely, very unlikely & 18 & 1, 1, 0, 0 \\
21 & spend much more, spend more, spend same, spend less, spend much less & 18 & 1, 1, 0, 0, 0 \\
22 & essential, very important, fairly important, not very important, not important at all & 17 & 1, 1, 1, 0, 0 \\
23 & very likely, somewhat likely, mixed, somewht unlikely, very unlikely & 17 & 1, 1, 0, 0, 0 \\
24 & not at all important, , , , , , very important & 17 & 0, 0, 0, 0, 1, 1, 1 \\
25 & often, sometimes, rarely, never & 16 & 1, 1, 1, 0 \\
26 & very scientific, pretty scientific, not too scientific, not scientific at all & 16 & 1, 1, 0, 0 \\
27 & should, should not & 15 & 1, 0 \\
28 & strongly agree, somewhat agree, somewht disagree, strngly disagree & 15 & 1, 1, 0, 0 \\
29 & what is best for the country, , , , own narrow interests & 15 & 1, 1, 0, 0, 0 \\
30 & very well, , , , not at all & 15 & 1, 1, 0, 0, 0 \\
31 & definitely expect, probably expect, probably not expect, definitely not expect & 15 & 1, 1, 0, 0 \\
32 & major reason, minor reason, not a reason & 15 & 1, 1, 0 \\
33 & did, didnt & 15 & 1, 0 \\
34 & no, yes, respondent, yes, someone respondent knows, yes, both respondent and someone respondent knows & 14 & 0, 1, 1, 1 \\
35 & very important, important, somewhat important, not at all important & 14 & 1, 1, 0, 0 \\
36 & no, yes & 14 & 0, 1 \\
37 & very likely, somewhat likely, not too likely, not likely at all & 13 & 1, 1, 0, 0 \\
38 & most impt, 2nd most impt, 3rd most impt, not chosen & 13 & 1, 1, 1, 0 \\
39 & 1 not at all effective, 2, 3, 4, 5 extremely effective & 13 & 0, 0, 0, 1, 1 \\
40 & a great deal, only some, hardly any & 13 & 1, 0, 0 \\
41 & definitely true, probably true, probably not true, definitely not true & 13 & 1, 1, 0, 0 \\
42 & allowed, not allowed & 12 & 1, 0 \\
43 & strongly agree, agree somewhat, disagree somewhat, strongly disagree & 12 & 1, 1, 0, 0 \\
44 & extremely likely, somewhat likely, not too likely, not likely at all & 12 & 1, 1, 0, 0 \\
45 & more often, at the same rate, less often, r not attending these activities & 12 & 1, 0, 0, 0 \\
46 & strongly agree, agree, uncertain, disagree, strong disagree & 12 & 1, 1, 0, 0, 0 \\
47 & more than once a week, once a week, once a month, at least 2 or 3 times in the past year, once in the past year, not at all in the past year & 11 & 1, 1, 1, 1, 1, 0 \\
48 & strongly agree, agree, disagree, strongly disagree & 11 & 0, 0, 1, 1 \\
49 & stongly agree, agree, not agree/dsagre, disagree, strong disagree & 11 & 1, 1, 0, 0, 0 \\
50 & excellent, very good, good, fair, poor & 11 & 1, 1, 1, 0, 0 \\
  \label{tab:binary_transformation}
\end{longtable}

%% file: Tables/table_question_distribution.tex
\begin{table}[!htbp]
\centering
\caption{Distribution of Question Types}
\label{tab:question_distribution}
\begin{tabular}{lrr}
\toprule
\textbf{Variable Category} & \textbf{Frequency} & \textbf{Proportion} \\
\midrule
Attitudinal/Opinion Questions         &  2521 &  68.15\% \\
Behavioral Questions                  &   993 &  26.85\% \\
Objective/Knowledge/Factual Questions &   185 &   5.00\% \\
\midrule
\textbf{Total Questions}                       &  3699 & 100.00\% \\
\bottomrule
\end{tabular}
\end{table}

%% file: Tables/table_alpaca_text_examples.tex
\begin{table}[!h]
    \caption{\textbf{Examples of text responses generated by the Alpaca-7B model.}}
    \label{tab:alpaca_text_examples}
    \renewcommand{\arraystretch}{1.5}
    \begin{center}
        \begin{tabular}{p{0.15\textwidth} p{0.375\textwidth} p{0.375\textwidth}}
            \toprule
            \textbf{Variables} & \textbf{Survey questions} & \textbf{Text responses} \\
            \midrule
homosex & What about sexual relations between two adults of the same sex--do you think it is always wrong, almost always wrong, wrong only sometimes, or not wrong at all? & I believe that sexual relations between two adults of the same sex is not wrong at all. \\
marhomo1 & Do you agree or disagree with the following statement?
Homosexual couples have the right to marry one another. & I agree that homosexual couples should have the right to marry one another. \\
busing & A. In general, do you favor or oppose the busing of (Negro/Black/African-American) and white school children from one school district to another? & I favor the busing of (Negro/Black/African-American) and white school children from one school district to another. I believe that busing is an effective way to create integrated schools and promote racial harmony. \\
nomeat & C. And how often do you refuse to eat meat for moral or environmental reasons? & I rarely refuse to eat meat for moral or environmental reasons. \\
           \bottomrule
        \end{tabular}
\end{center}
\end{table}

%% file: Tables/table_similar_questions.tex
\begin{longtable}{
p{5cm}@{}
p{0.3cm}@{}
>{\raggedright}p{8cm}@{}
p{0.3cm}@{}
p{2cm}@{}
}
\caption{\textbf{Most similar survey questions.}}
\label{tab:similar_questions}
\\
 \toprule
\textbf{Survey questions} & & \textbf{Most similar survey questions} & & \textbf{Cosine \linebreak similarity} \\ \midrule
\endfirsthead
\\
 \toprule
\textbf{Survey questions} & & \textbf{Most similar survey questions} & & \textbf{Cosine \linebreak similarity} \\ \midrule
\endhead
\multirow{3}{5cm}{(homosex) What about sexual relations between two adults of the same sex--do you think it is always wrong, almost always wrong, wrong only sometimes, or not wrong at all?} & & (homosex1) C. And what about sexual relations between two adults of the same sex, is it. . . & & 0.932 \\
 & & (premarsx) There's been a lot of discussion about the way morals and attitudes about sex are changing in this country. If a man and woman have sex relations before marriage, do you think it is always wrong, almost always wrong, wrong only
sometimes, or not wrong at all? & & 0.918 \\
 & & (grass) Do you think the use of marijuana should be made legal or not? & & 0.890 \\
 \hline
\multirow{3}{5cm}{(marhomo1) Do you agree or disagree with the following statement?
Homosexual couples have the right to marry one another.} & & (marhomo) Do you agree or disagree?
J. Homosexual couples should have the right to marry one another. & & 0.990 \\
 & & (ssfchild) To what extent do you agree or disagree with the following statements?
A. A same sex female couple can bring up a child as well as a male-female couple. & & 0.941 \\
 & & (ssmchild) To what extent do you agree or disagree with the following statements?
B. A same sex male couple can bring up a child as well as a male-female couple. & & 0.939 \\
 \hline
\multirow{3}{5cm}{(busing) A. In general, do you favor or oppose the busing of (Negro/Black/African-American) and white school children from one school district to another?} & & (engoff1) Do you favor a law making English the official language of the United States, or do you oppose such a law? & & 0.906 \\
 & & (racsubs) Some religious and business groups have set up programs to encourage (Negro/Black) people to buy houses in white suburbs. Do you favor or oppose these voluntary programs to integrate white suburbs? & & 0.903 \\
 & & (busing10) B. Now, thinking about ten years ago, that is in 1972, did you then favor or oppose the busing of (Negro/Black) and white school children from one school district to another? & & 0.903 \\
 \hline
\multirow{3}{5cm}{(nomeat) C. And how often do you refuse to eat meat for moral or environmental reasons?} & & (drivless) D. And how often do you cut back on driving a car for environmental reasons? & & 0.935 \\
 & & (nobuygrn) K3. And how often do you avoid buying certain products for environmental reasons? & & 0.926 \\
 & & (chemfree) B. And how often do you make a special effort to buy fruits and vegetables grown without pesticides or chemicals? & & 0.917 \\
\bottomrule
\end{longtable}

%% file: Tables/table_top20_modules.tex
\begin{longtable}{p{4cm}p{8cm}r}
\caption{Top 20 Modules by Number of Variables in the Module with Representative Variables}\label{tab:top20_modules_updated}\\
\hline
\textbf{Module} & \textbf{Representative Variables} & \textbf{\# Variables} \\
\hline
\endfirsthead
\hline
\textbf{Module} & \textbf{Representative Variables} & \textbf{\# Variables} \\
\hline
\endhead
\hline \multicolumn{3}{r}{{Continued on next page}} \\
\endfoot
\hline
\endlastfoot
Science Knowledge \& Attitudes & scispec (science is too concerned with theory and speculation),; leadsci (shld pay attntn only to theory accepted by leading scientists),; whichsci (industry scientists less reliable than univer'sity scientists) & 195 \\
Information Society & compuse (r use computer),; webtv (r has internet via webtv),; webmob (r uses home internet through mobile device) & 189 \\
ISSP Role of Gov't & prespop (approve of pres handling job),; volactyr (since last yr any volunteering),; volacty2 (done other types of volunteering for child's school or youth org) & 159 \\
Work Orientation & stress (how often does r find work stressful),; rintjob (r's job is interesting),; rwrkindp (r can work independently) & 150 \\
Work Organizations & genejob (should employers give genetic tests),; genehire (employers' right to hire by genetic tests),; genecanx (employer test cancer or safeworkplace) & 100 \\
ISSP Family \& Gender Roles & kidsuffr (preschooler will suffer if mom works),; famsuffr (family life suffers if mom works f-t),; hapifwrk (woman and family happier if she works) & 100 \\
Mental Health & psycmed1 (psychiatric medicine is harmful to the body),; psycmed2 (people should stop taking if symptoms gone),; psycmed3 (taking these medicine interferes w daily act) & 94 \\
ISSP Environment & obeythnk (should chldrn be obedient or think for selves),; privent (private enterprise will solve u.s. problems),; scifaith (believe too much in science, not enough faith) & 85 \\
Intergroup Relations & farejews (self supporting - live off welfare),; fareblks (self supporting - live off welfare),; farehsps (self supporting - live off welfare) & 82 \\
ISSP Religion & hapunhap (happy or unhappy with life today),; stiffpun (lawbreakers should get stiffer sentences),; deathpen (murderers should get death penalty) & 77 \\
ISSP National Identity & clsenei (how close do you feel to your neighborhood),; clsetown (how close do you feel to your town or city),; clsestat (how close do you feel to your state) & 75 \\
Quality of Working Life & safetywk (worker safety priority at work),; safefrst (no shortcuts on worker safety),; teamsafe (mgt and employees work together re safety) & 73 \\
Culture & readfict (read novels poems or plays),; popmusic (went to a live performance of popular music),; drama (went to a live drama) & 72 \\
Markets & carprivt (have you purchased a used car past 5 yrs?),; warrntyc (did the seller provide a warranty?),; cardealr (purchased a car from a dealership past 5 yrs?) & 64 \\
Emotions & lonely (felt lonely?),; noplan (there's no sense planning a lot),; badbrks (most of my problems are due to bad breaks.) & 62 \\
ISSP Citizenship & voteelec (how important always to vote in elections),; paytaxes (how important never to try to evade taxes),; obeylaws (how important always to obey laws) & 62 \\
ISSP Social Inequality & opwlth (need wealthy family to get ahead),; oppared (need educated parents to get ahead),; opeduc (need good education to get ahead) & 58 \\
National Security & sectech (should govt maintain secrecy over technology),; secdocs (does govt classify too many documents as secrets),; rptcowrk (report coworkers threatening secrecy of work) & 57 \\
Mental Health II & badchar (situation caused by own bad character),; chembal (situation by a chemical imbalance in the brain),; stressfl (situation caused by stressful circumstances) & 54 \\
ISSP Social Networks \& Support Systems & localgvt (how likely can r do sth for improvmnt in community),; macall (contact with mother),; pacall (contact with father) & 54 \\
\end{longtable}

%% file: Tables/table_modelcomparison.tex
\begin{table}
  \caption{Prediction performance across five models and three scenarios.}
  \label{tab:modelcomparison}
  \centering
  \begin{tabular}{@{}p{2.3cm} p{1.6cm} >{\centering\arraybackslash}p{1.9cm} >{\centering\arraybackslash}p{1.9cm} >{\centering\arraybackslash}p{1.9cm} >{\centering\arraybackslash}p{1.9cm} >{\centering\arraybackslash}p{1.9cm}@{}}
    \toprule
    & & \multicolumn{5}{c}{Models} \\ \cmidrule(l){3-7}
     &  & Alpaca-7B & GPT-J-6b & RoBERTa-large & \shortstack{Matrix\\Factorization} & MICE \\ \midrule
    \multirow{3}{2.3cm}{Missing data imputation} & AUC & \textbf{0.868} & 0.866 & 0.860 & 0.856 & 0.780 \\
     & Accuracy & 0.783 & 0.778 & 0.774 & \textbf{0.784} & 0.751 \\
     & F1-score & 0.789 & \textbf{0.791} & 0.780 & 0.785 & 0.756 \\ \midrule
    \multirow{3}{2.3cm}{Retrodiction} & AUC & \textbf{0.862} & 0.861 & 0.854 & 0.807 & 0.727 \\
     & Accuracy & \textbf{0.776} & 0.773 & 0.768 & 0.734 & 0.690 \\
     & F1-score & 0.783 & \textbf{0.786} & 0.773 & 0.704 & 0.694 \\ \midrule
    \multirow{3}{2.3cm}{Unasked opinion prediction} & AUC & \textbf{0.701} & 0.661 & 0.573 &  &  \\
     & Accuracy & \textbf{0.648} & 0.616 & 0.553 &  &  \\
     & F1-score & \textbf{0.659} & 0.641 & 0.574 &  &  \\ \bottomrule
  \end{tabular}

  \medskip
  \parbox{\linewidth}{\footnotesize \textit{Notes.} The best-performing models are highlighted in bold. AUC measures the probability of ranking a randomly selected positive response above a randomly selected negative response. Accuracy is (TP + TN) / (TP + FP + TN + FN), and F1 is $2 \cdot (\mathrm{precision} \cdot \mathrm{recall}) / (\mathrm{precision} + \mathrm{recall})$. Matrix factorization and MICE cannot be applied to unasked-opinion prediction.}
\end{table}

%% file: Tables/table_retro_interp_forecast.tex
\begin{table}[!htbp]
\caption{Alpaca individual-level performance, by retrodiction scenario.}
\label{tab:retro_interp_forecast}
\centering
\small
\begin{tabular}{lrrrr}
\toprule
Scenario & $N$ & AUC & Accuracy & F1 \\
\midrule
Backcasting & 1,660,211 & 0.842 & 0.757 & 0.780 \\
Interpolation & 8,550,887 & 0.868 & 0.782 & 0.784 \\
Forecasting & 1,816,822 & 0.852 & 0.766 & 0.782 \\
\bottomrule
\end{tabular}

\medskip
\parbox{\linewidth}{\footnotesize \textit{Notes.} Each held-out validation cell $(v, y)$ is classified by where $y$ sits relative to the variable's other observation years: backcasting if $y$ is before the earliest, forecasting if $y$ is after the latest, otherwise interpolation. $N$ is the number of held-out (respondent, variable, year) cells pooled across the retrodiction-task CV folds.}
\end{table}

%% file: Tables/table_retro_distance_metrics.tex
\begin{table}[!htbp]
\caption{Alpaca individual-level performance, by retrodiction scenario and distance to the nearest training year.}
\label{tab:retro_distance_metrics}
\centering
\small
\begin{tabular}{llrrrr}
\toprule
Scenario & Distance to nearest training year & $N$ & AUC & Accuracy & F1 \\
\midrule
\multicolumn{6}{l}{\textit{A. Backcasting}} \\
 & 1 & 328,901 & 0.855 & 0.770 & 0.777 \\
 & 2-3 & 382,191 & 0.843 & 0.763 & 0.797 \\
 & 4-5 & 296,285 & 0.858 & 0.775 & 0.801 \\
 & 6-7 & 112,917 & 0.869 & 0.783 & 0.816 \\
 & $8+$ & 471,546 & 0.835 & 0.752 & 0.765 \\
\midrule
\multicolumn{6}{l}{\textit{B. Interpolation}} \\
 & 1 & 3,303,364 & 0.873 & 0.787 & 0.777 \\
 & 2-3 & 4,449,777 & 0.865 & 0.778 & 0.784 \\
 & 4-5 & 360,677 & 0.880 & 0.795 & 0.825 \\
 & 6-7 & 147,530 & 0.862 & 0.778 & 0.804 \\
 & $8+$ & 285,980 & 0.838 & 0.759 & 0.803 \\
\midrule
\multicolumn{6}{l}{\textit{C. Forecasting}} \\
 & 1 & 88,056 & 0.845 & 0.761 & 0.790 \\
 & 2-3 & 764,806 & 0.863 & 0.779 & 0.800 \\
 & 4-5 & 243,915 & 0.859 & 0.775 & 0.795 \\
 & 6-7 & 48,794 & 0.819 & 0.742 & 0.777 \\
 & $8+$ & 600,481 & 0.855 & 0.769 & 0.770 \\
\bottomrule
\end{tabular}

\medskip
\parbox{\linewidth}{\footnotesize \textit{Notes.} Each held-out validation cell is classified by scenario (\emph{backcasting}: held-out year before the variable's earliest training year; \emph{forecasting}: after the latest; \emph{interpolation}: in between) and by its distance, in years, to the variable's nearest training year in the same fold. AUC, Accuracy, and F1 are computed over individual (respondent, variable, year) cells pooled across the 10 retrodiction-task CV folds.}
\end{table}

%% file: Tables/table_featureimportance.tex
\begin{table}[!htbp]
\caption{Feature importance from permutation experiments.}
\label{tab:metrics}
\centering
\begin{tabular}{lcccc}
\toprule
\textbf{Metric} &
\shortstack{\textbf{Retrodiction}\\\textbf{(Original)}} &
\shortstack{\textbf{Shuffling}\\\textbf{Question}\\\textbf{Embedding}} &
\shortstack{\textbf{Shuffling}\\\textbf{Respondent}\\\textbf{Embedding}} &
\shortstack{\textbf{Shuffling}\\\textbf{Period}\\\textbf{Embedding}} \\
\midrule
AUC        & 0.857 & 0.509 & 0.720 & 0.849 \\
Accuracy   & 0.766 & 0.508 & 0.660 & 0.761 \\
F1-score   & 0.784 & 0.544 & 0.684 & 0.779 \\
\bottomrule
\end{tabular}

\medskip
\parbox{\linewidth}{\footnotesize \textit{Notes.} We randomly shuffle one embedding at a time while keeping all others unchanged, and report the resulting drop in retrodiction performance (AUC, Accuracy, F1). The larger the drop, the more important the embedding.}
\end{table}

%% file: Tables/table_accuracy_thresholds.tex
\begin{table}[!htbp]
    \centering
    \caption{Percentage of opinions predicted by different models within varying accuracy thresholds}
    \label{tab:model_prediction_accuracy}
    \begin{tabular}{lccc}
        \hline
        \textbf{Prediction intervals} & \textbf{Imputation (\%)} & \textbf{Retrodiction (\%)} & \textbf{Unasked prediction (\%)} \\
        \hline
        within 1\% & 29.7 & 17.6 & 3.3 \\
        within 2\% & 55.5 & 35.4 & 7.1 \\
        within 3\% & 74.0 & 50.6 & 10.8 \\
        within 4\% & 85.7 & 63.6 & 14.2 \\
        within 5\% & 91.9 & 73.6 & 17.5 \\
        within 6\% & 95.0 & 80.9 & 21.0 \\
        within 7\% & 96.8 & 86.2 & 24.5 \\
        within 8\% & 97.9 & 89.9 & 27.8 \\
        within 9\% & 98.7 & 92.7 & 30.9 \\
        within 10\% & 99.3 & 94.6 & 34.3 \\
        \hline
    \end{tabular}
\end{table}

%% file: Tables/table_samesex_50.tex
\begin{longtable}{clp{7cm}c}
\caption{List of the 50 Survey Questions Most Closely Correlated with Same-Sex Marriage Opinions.}\label{tab:samesex_50} \\ \hline
\textbf{Rank} & \textbf{Variable} & \textbf{Description} & \textbf{Abs(Correlation)} \\ \hline
1 & homosex1 & Is homosexual sex wrong? & 0.645 \\
2 & homosex & Homosexual sex relations & 0.628 \\
3 & ssfchild & Same sex female couple raise child as well as male-female couple & 0.580 \\
4 & ssmchild & Same sex male couple raise child as well as male-female couple & 0.556 \\
5 & negjob5 & In last 5 yrs, r been denied promo or recvd bad evaluation bc employr belivd r w & 0.500 \\
6 & dwell5 & In last 5 yrs, r has been prevented from moving bc landlord believed r was gay/l & 0.500 \\
7 & cohabok & Living together as an acceptable option & 0.498 \\
8 & attractd & At what age were you first sexually attracted to same sex & 0.430 \\
9 & abfelegl & Women only: women should be able to have legal abortions & 0.395 \\
10 & premarsx & Sex before marriage & 0.392 \\
11 & evolved & Sci knowledge: human beings developed from animals & 0.372 \\
12 & abmelegl & Men only: women should be able to have legal abortions & 0.357 \\
13 & premars1 & Is premarital sex wrong? & 0.355 \\
14 & abany & Abortion if woman wants for any reason & 0.354 \\
15 & godmeans & Life meaningful because god exists & 0.354 \\
16 & amchrstn & How important to be a christian & 0.350 \\
17 & abnomore & Married--wants no more children & 0.344 \\
18 & libhomo & Allow homosexuals book in library & 0.344 \\
19 & abhelp1 & R would help with arrangements for abortion & 0.339 \\
20 & abpoorw & Wrong for woman to get abortion if low income? & 0.339 \\
21 & grass & Should marijuana be made legal & 0.338 \\
22 & suicide1 & Suicide if incurable disease & 0.338 \\
23 & absingle & Not married & 0.334 \\
24 & openrel2 & R and leader ever become an openly acknowledged couple & 0.333 \\
25 & prespop & Approve of pres handling job & 0.329 \\
26 & abpoor & Low income--cant afford more children & 0.328 \\
27 & marlegit & Those wanting kids should get married & 0.326 \\
28 & toldsxor & At what age you first told someone that you were gay/les/bi & 0.319 \\
29 & abmoral & R has a moral opposition to abortion & 0.318 \\
30 & colhomo & Allow homosexual to teach & 0.310 \\
31 & abinspay & Should people be able to use health insurance for abortion - ver y & 0.308 \\
32 & pillok & Birth control to teenagers 14-16 & 0.305 \\
33 & scresrch & Government should fund stem cell research & 0.304 \\
34 & letdie1 & Allow incurable patients to die & 0.303 \\
35 & spkhomo & Allow homosexual to speak & 0.301 \\
36 & fefam & Better for man to work, woman tend home & 0.298 \\
37 & bigbang & Sci knowledge: the universe began with a huge explosion & 0.298 \\
38 & kidlived & Children under 18 have ever lived with r under r's care & 0.297 \\
39 & savesoul & Tried to convince others to accept jesus & 0.294 \\
40 & letdie1y & Allow incurable patients to die (form 2) & 0.293 \\
41 & abdefctw & Wrong for woman to get abortion for birth defects & 0.291 \\
42 & abrape & Pregnant as result of rape & 0.289 \\
43 & polviews2 &  & 0.284 \\
44 & prayer & Bible prayer in public schools & 0.281 \\
45 & feelrel & How religious is r & 0.280 \\
46 & polviews1 &  & 0.278 \\
47 & attend & How often r attends religious services & 0.278 \\
48 & theism & God concerned with human beings personally & 0.275 \\
49 & reborn & Has r ever had a 'born again' experience & 0.272 \\
50 & pilloky & Birth control to teenagers 14-16 (form 2) & 0.271 \\ \hline
  \end{longtable}

%% file: Tables/table_counterfactual_roper_surveys.tex
\clearpage
\begin{longtable}{rlp{7.0cm}p{4.0cm}}
\caption{Roper Center surveys used as external-validation ground truth for the counterfactual-trend example variables (panels A--D of Figure~6).}
\label{tab:counterfactual_roper_surveys} \\
\toprule
Year & GSS variable & Roper study title & Source \\
\midrule
\endfirsthead

\multicolumn{4}{l}{\textit{(continued from previous page)}} \\
\toprule
Year & GSS variable & Roper study title & Source \\
\midrule
\endhead

\midrule
\multicolumn{4}{r}{\textit{(continued on next page)}} \\
\endfoot

\bottomrule
\multicolumn{4}{p{15cm}}{\footnotesize \textit{Notes.} Matches are filtered to US national-adult samples, LLM-verified same-construct pairs with confidence $\geq 0.85$, simple binary response mapping, and years in which the GSS did not field the corresponding question. ``Source'' is the surveying / sponsoring organization listed in Roper's study metadata.} \\
\endlastfoot

\multicolumn{4}{l}{\textit{A. homosexual couples have the right to marry}} \\
1992 & marhomo1 & Yankelovich/Time Magazine/CNN Poll: 1992 Presidential Election/Abortion/Family Values/Bab… & Time Magazine/Cable News Network (CNN) \\
1993 & marhomo1 & Yankelovich/Time/CNN Poll: Politics, Religion, and Morality & Time Magazine/Cable News Network (CNN) \\
1994 & marhomo1 & ABC News/Washington Post Poll:  Role of Government/Democrats/Republicans & ABC News/Washington Post \\
1996 & marhomo1 & Associated Press Poll \# 1996-924L: Homosexuality & Associated Press \\
1996 & marhomo1 & Gallup/CNN/USA Today Poll \# 1996-9603008: Politics/1996 Election & Cable News Network (CNN)/USA Today \\
1996 & marhomo1 & Post-Modernity Project at the University of Virginia Poll: January 1996 & Post-Modernity Project at the University of Virgi… \\
1996 & marhomo1 & Wirthlin Worldwide Poll: December 1996 & Wirthlin Worldwide \\
1998 & marhomo1 & Yankelovich/Time Magazine/CNN Poll: Politics/Homosexuality & Time Magazine/Cable News Network (CNN) \\
1999 & marhomo1 & Gallup News Service Poll \#9902009:  Impeachment Handling/Past Presidents & Gallup Organization \\
2000 & marhomo1 & Gallup/CNN/USA Today Poll:  Election 2000/Guns/Taxes/Defense & Cable News Network (CNN)/USA Today \\
2000 & marhomo1 & Los Angeles Times Poll \# 2000-442: Abortion/Gay Rights & Los Angeles Times \\
2002 & marhomo1 & Gallup Organization Poll: May 2002 & Gallup Organization \\
2003 & marhomo1 & ABC News Poll: September 2003 & ABC News \\
2003 & marhomo1 & CBS News/New York Times Poll \# 2003-12A: George W. Bush/Medicare/2004 Presidential Electi… & CBS News/New York Times \\
2003 & marhomo1 & CBS News/New York Times Poll:  Race Relations/Bush Administration/Family Values & CBS News/New York Times \\
2003 & marhomo1 & Gallup/CNN/USA Today Poll:  2004 Presidential Election/Iraq/Gay Marriage & Cable News Network (CNN)/USA Today \\
2003 & marhomo1 & Gallup/CNN/USA Today Poll:  Iraq/New Tax Credit Law/Medicare Policy/Major League Baseball & Cable News Network (CNN)/USA Today \\
2003 & marhomo1 & Gallup/CNN/USA Today Poll:  Presidential Election/Iraq/Abortion/Investments/Medicare & Cable News Network (CNN)/USA Today \\
2003 & marhomo1 & Harris Interactive/Time Magazine/CNN Poll \# 2003-06: 2004 Presidential Election/Foreign P… & Time Magazine/Cable News Network (CNN) \\
2003 & marhomo1 & Time Magazine/CNN Poll \# 2003-09: 2004 Presidential Election/George W. Bush Job Performan… & Time Magazine/Cable News Network (CNN) \\
2004 & marhomo1 & ABC News/Washington Post Poll: Pre-State of the Union Address & ABC News/Washington Post \\
2004 & marhomo1 & CBS News Poll \# 2004-02C: 2004 Presidential Election/War in Iraq/Same-Sex Marriages & CBS News \\
2004 & marhomo1 & Gallup News Service Poll: 2004 Presidential Election/Gay Marriage/Superbowl Halftime Show & Gallup Organization \\
2004 & marhomo1 & Gallup Organization Poll: May 2004 & Gallup Organization \\
2004 & marhomo1 & Gallup/CNN/USA Today Poll:  2004 Presidential Election/Gay Marriage/"The Passion of the C… & Cable News Network (CNN)/USA Today \\
2004 & marhomo1 & Gallup/CNN/USA Today Poll:  2004 Presidential Election/Patriot Act/Gay Marriage & Cable News Network (CNN)/USA Today \\
2004 & marhomo1 & Gallup/CNN/USA Today Poll:  2004 Presidential Election/Terrorism/Same-Sex Marriages & Cable News Network (CNN)/USA Today \\
2004 & marhomo1 & Gallup/CNN/USA Today Poll:  Election Benchmark II & Cable News Network (CNN)/USA Today \\
2004 & marhomo1 & Pew Internet \& American Life Project Poll: Selective Exposure Survey & Pew Internet \& American Life Project \\
2004 & marhomo1 & SRBI/Time Magazine Poll \# 2004-3333: Election 2004 & Time Magazine \\
2004 & marhomo1 & Time Magazine/CNN Poll \# 2004-02: 2004 Presidential Election/George W. Bush/Homosexual Ma… & Time Magazine/Cable News Network (CNN) \\
2004 & marhomo1 & Time Magazine/CNN Poll \# 2004-05: 2004 Presidential Election/War in Iraq/Abuse of Iraqi P… & Time Magazine/Cable News Network (CNN) \\
2005 & marhomo1 & Gallup News Service Poll \# 2005-39: Economy/Finances & Gallup Organization \\
2005 & marhomo1 & Gallup/CNN/USA Today Poll \# 2005-14: Social Security/Same-Sex Marriage/Terri Schiavo & Cable News Network (CNN)/USA Today \\
2005 & marhomo1 & Gallup/CNN/USA Today Poll \# 2005-21:  Social Security/Iraq/Filibuster Procedures/Child Se… & Cable News Network (CNN)/USA Today \\
2007 & marhomo1 & CNN/ORC Poll \# 2007-008: War in Iraq/Bridges/2008 Presidential Election & Cable News Network (CNN) \\
2007 & marhomo1 & CNN/ORC Poll \# 2007-010: Immigration/Iran/2008 Presidential Election & Cable News Network (CNN) \\
2009 & marhomo1 & CNN/ORC Poll: Recession/Muslim World/H1N1/Swine Flu/Supreme Court & Cable News Network (CNN) \\
2009 & marhomo1 & Pew Research Center Poll: 2009 Religion \& Public Life & Pew Research Center for the People \& the Press/Pe… \\
2011 & marhomo1 & Pew Research Center Poll: September 2011 Generations Survey & Pew Research Center for the People \& the Press \\
2013 & marhomo1 & CBS News Poll: Congress & CBS News \\
2015 & marhomo1 & CBS News/New York Times Poll: 2016 Presidential Campaign/Republican Party Nominees/Gun La… & CBS News/New York Times \\
2015 & marhomo1 & CBS News/New York Times Poll: 2016 Presidential Campaign/Republican and Democratic Party … & CBS News/New York Times \\
2015 & marhomo1 & CBS News/New York Times Poll: 2016 Presidential Campaign/Republican and Democratic Party … & CBS News/New York Times \\
\midrule
\multicolumn{4}{l}{\textit{B. favor busing of Black and white schoolchildren}} \\
2019 & busing & Gallup Poll & Gallup Organization \\
\midrule
\multicolumn{4}{l}{\textit{C. confidence in U.S. Congress}} \\
1992 & concong & Los Angeles Times Poll \#268:  National Politics and the State of the Union & Los Angeles Times \\
1992 & concong & Los Angeles Times Poll \#275:  National Politics and the Congressional Scandal & Los Angeles Times \\
1992 & concong & Louis Harris \& Associates Poll: February 1992 & Louis Harris \& Associates \\
1993 & concong & Los Angeles Times Poll:  Clinton Inauguration & Los Angeles Times \\
1993 & concong & Los Angeles Times Poll:  National Economy and Political Coalitions & Los Angeles Times \\
1993 & concong & Los Angeles Times Poll:  National Survey - Year Roundup & Los Angeles Times \\
1993 & concong & Louis Harris \& Associates Poll: January 1993 & Louis Harris \& Associates \\
1994 & concong & Los Angeles Times Poll:  National Issues, April 1994 & Los Angeles Times \\
1994 & concong & Los Angeles Times Poll: National Politics & Los Angeles Times \\
1994 & concong & Louis Harris \& Associates Poll: February 1994 & Louis Harris \& Associates \\
1995 & concong & Los Angeles Times Poll \# 1995-356:  National Issues & Los Angeles Times \\
1995 & concong & Los Angeles Times Poll \# 1995-369:  Federal Government/National Issues & Los Angeles Times \\
1995 & concong & Los Angeles Times Poll:  National Issues, January 1995 & Los Angeles Times \\
1995 & concong & Louis Harris \& Associates Poll: February 1995 & Louis Harris \& Associates \\
1996 & concong & Louis Harris \& Associates Poll: January 1996 & Louis Harris \& Associates \\
1999 & concong & Los Angeles Times Poll \# 1999-422: Impeachment Trial & Los Angeles Times \\
2000 & concong & Harris Interactive Poll: January 2000 & Harris Interactive \\
2001 & concong & Harris Interactive Poll: January 2001 & Harris Interactive \\
2002 & concong & Harris Interactive Poll: December 2002 & Harris Interactive \\
2002 & concong & Harris Interactive Poll: January 2002 & Harris Interactive \\
2004 & concong & Harris Interactive Poll: February 2004 & Harris Interactive \\
2005 & concong & Harris Interactive Poll: February 2005 & Harris Interactive \\
2005 & concong & Hart-McInturff/NBC/WSJ Poll \# 2005-6054: Economy/Social Security/Iraq & NBC News/Wall Street Journal \\
2006 & concong & Harris Interactive Poll: February 2006 & Harris Interactive \\
2007 & concong & Harris Interactive Poll: February 2007 & Harris Interactive \\
2009 & concong & Harris Interactive Poll: February 2009 & Harris Interactive \\
2010 & concong & Harris Interactive Poll: February 2010 & Harris Interactive \\
2013 & concong & CBS News Poll: Congress & CBS News \\
2017 & concong & McClatchy-Marist Poll: March 2017 & McClatchy \\
2017 & concong & NPR/PBS NewsHour/Marist Poll: June 2017 & PBS NewsHour/NPR \\
2017 & concong & The Chicago Council on Global Affairs Poll: 2017 Biannual Survey & The Chicago Council on Global Affairs \\
2019 & concong & NPR/PBS NewsHour/Marist Poll: October 2019 & PBS NewsHour/NPR \\
\midrule
\multicolumn{4}{l}{\textit{D. lived with spouse before marriage}} \\
1995 & cohabit & CBS News Poll: Marriage/World War II/Vietnam/Oklahoma City Bombing & CBS News \\
1996 & cohabit & Gallup News Service Poll:  Para-Normal Beliefs & Gallup Organization \\
2002 & cohabit & Gallup Organization Poll: July 2002 & Gallup Organization \\
\end{longtable}

%% file: Tables/table_samesex_framing.tex
\begin{table}[!htbp]
\centering
\caption{Effect of the word ``should'' on predicted agreement with same-sex marriage rights.}
\label{tab:samesex_framing}
\begin{tabular}{p{9cm} r}
\toprule
Survey question & Mean (\%) \\
\midrule
``Do you agree or disagree with the statement that homosexual couples have the right to marry one another?'' & 37.35 \\
\addlinespace
``Do you agree or disagree that homosexual couples \emph{should} have the right to marry one another?'' & 27.93 \\
\midrule
\textit{Paired difference}, Wilcoxon signed-rank test: $z = -194.82$, $p$ $< .001$, $N = 68,846$ & $-9.42$ \\
\bottomrule
\end{tabular}
\par\vspace{0.5em}
\begin{minipage}{\linewidth}
\footnotesize
\emph{Note.} Two GSS items differ only by the inclusion of the word ``should'' (one item asks about a descriptive right, the other about a normative right). The two items were never fielded in the same wave, so for every respondent in the analytic sample we generate the model's predicted probability for both items by holding the respondent and period embeddings constant and swapping the question embedding. Each respondent therefore contributes a paired prediction; the difference is tested by a Wilcoxon signed-rank test.
\end{minipage}
\end{table}

%% file: Tables/table_hybrid_mf.tex
\begin{table}[!htbp]
\caption{Text-aware matrix-factorization baselines.}
\label{tab:hybrid_mf}
\centering
\small
\begin{tabular}{llrr}
\toprule
Scenario & Model & AUC & Accuracy \\
\midrule
Missing data imputation & \textbf{Alpaca-7B (ours)} & \textbf{0.868} & \textbf{0.782} \\
 & Matrix factorization & 0.856 & 0.784 \\
 & MF + TF-IDF & 0.853 & 0.782 \\
 & MF + embedding (SentenceBERT) & 0.854 & 0.782 \\
\midrule
Retrodiction & \textbf{Alpaca-7B (ours)} & \textbf{0.829} & \textbf{0.739} \\
 & Matrix factorization & 0.728 & 0.686 \\
 & MF + TF-IDF & 0.723 & 0.669 \\
 & MF + embedding (SentenceBERT) & 0.723 & 0.669 \\
\bottomrule
\end{tabular}

\medskip
\parbox{\linewidth}{\footnotesize \textit{Notes.} \emph{MF + TF-IDF} augments the user--item rating matrix with raw TF-IDF question-text features and factorizes jointly; \emph{MF + embedding} uses SentenceBERT sentence embeddings in the same place. This follows the hybrid / collective-factorization design of Singh and Gordon (2008) and the feature-augmented variants discussed in Koren et al.\ (2009).}
\end{table}

%% file: Tables/table_frontier_llm.tex
\begin{table}[!htbp]
\caption{Our models based on Alpaca-7B vs.\ prompted out-of-shelf LLMs on the GSS 2018 validation data.}
\label{tab:frontier_llm}
\centering
\small
\begin{tabular}{llrr}
\toprule
Model & Strategy & Accuracy & F1 \\
\midrule
\multirow{3}{*}{Alpaca-7B}
 & Imputation   & \textbf{0.787}  & \textbf{0.813}  \\
 & Retrodiction & \textbf{0.723} & \textbf{0.764} \\
 & Unasked      & \textbf{0.672} & \textbf{0.725} \\
\midrule
\multirow{3}{*}{GPT-4o}
 & Demographics & 0.638 & 0.673 \\
 & Random-$k{=}10$ & 0.644 & 0.649 \\
 & Correlation-top-$k{=}10$ & 0.703 & 0.714 \\
\midrule
\multirow{3}{*}{Gemini-2.5-flash}
 & Demographics & 0.655 & 0.659 \\
 & Random-$k{=}10$ & 0.656 & 0.661 \\
 & Correlation-top-$k{=}10$ & 0.701 & 0.712 \\
\bottomrule
\end{tabular}

\medskip
\parbox{\linewidth}{\footnotesize \textit{Notes.} Alpaca-7B entries (in bold) are the baseline evaluated on the three scenarios (imputation, retrodiction, unasked). Out-of-shelf LLM entries show three prompting strategies (Demographics, Random-$k{=}10$ context, Correlation-top-$k{=}10$ context) evaluated on the GSS 2018 data.}
\end{table}

%% file: Tables/table_perfbygroup.tex
\clearpage
\begin{sidewaystable}
  \caption{Performance by sparsity, temporal distance, and volatility.}
  \label{tab:perfbygroup}
  \centering
  \begin{tabular}{lrrcccccc}
    \toprule
    \textbf{Scenario} & $N_{\text{var, year}}$ & $N_{\text{var}}$ & \textbf{Our Method} & \textbf{Regression} & \textbf{Logistic} & \textbf{Reg+Sq} & \textbf{Logistic+Sq} & \textbf{MF} \\
     &  &  & ($\rho$ / MAE) & ($\rho$ / MAE) & ($\rho$ / MAE) & ($\rho$ / MAE) & ($\rho$ / MAE) & ($\rho$ / MAE) \\
    \midrule
    All & 8858 & 1317 & \textbf{.982/.038} & .981/.034 & .981/.033 & .973/.040 & .972/.037 & .831/.108 \\
    \midrule
    \multicolumn{9}{l}{\textit{A. Sparsity}} \\
    year = 1 & 1050 & 525 & \textbf{.968/.051} &  &  &  &  & .809/.215 \\
    year = 2 & 654 & 218 & \textbf{.965/.047} & .944/.059 & .945/.055 &  &  & .767/.183 \\
    year $>$ 2 & 7154 & 574 & .985/.035 & .986/.030 & \textbf{.986/.029} & .977/.036 & .975/.032 & .895/.085 \\
    \midrule
    \multicolumn{9}{l}{\textit{B. Volatility Level}} \\
    High volatility & 3749 & 419 & \textbf{.960/.045} & .958/.044 & .959/.042 & .953/.050 & .951/.042 & .737/.109 \\
    Low volatility & 5109 & 898 & .988/.033 & .993/.022 & \textbf{.993/.021} & .988/.025 & .987/.024 & .874/.107 \\
    \midrule
    \multicolumn{9}{l}{\textit{C. Temporal Distance to the Nearest Year}} \\
    Distance $\geq$ 3 & 2222 & 929 & \textbf{.971/.047} & .958/.054 & .959/.050 & .903/.098 & .894/.071 & .796/.169 \\
    Distance $<$ 3 & 6636 & 680 & .985/.035 & .988/.027 & .989/.026 & \textbf{.989/.026} & .989/.026 & .883/.087 \\
    \bottomrule
  \end{tabular}

  \medskip
  \parbox{\linewidth}{\footnotesize \textit{Notes.} Scenario describes the condition under which survey questions are missing. $N_{\text{var, year}}$ is the number of variable--year entries; $N_{\text{var}}$ is the number of distinct GSS variables in each group. Each cell reports $\rho$ / MAE, where $\rho$ is the Spearman correlation between predicted and observed population averages and MAE is the mean absolute error. ``Regression'' is linear time-series regression (OLS, $\text{obs\_wavg} \sim \text{year}$); ``Logistic'' is the same regression with a Binomial logit link; ``Reg+Sq'' adds a quadratic year term; ``Logistic+Sq'' is the quadratic regression with a Binomial logit link; ``MF'' is the matrix-factorization baseline. \textbf{Bold} indicates the highest~$\rho$ in each row.}
\end{sidewaystable}
\clearpage

%% file: Tables/table_polviews_rank.tex
{\small
\begin{longtable}{rlp{8cm}rr}
\caption{GSS variables ranked by absolute Spearman correlation with the 7-point political-ideology scale (polviews).}\label{tab:polviews_rank} \\
\toprule
Rank & Variable & Description & $\rho$(polviews) & Alpaca AUC \\
\midrule
\endfirsthead
\multicolumn{5}{l}{\textit{(continued from previous page)}} \\
\toprule
Rank & Variable & Description & $\rho$(polviews) & Alpaca AUC \\
\midrule
\endhead
\multicolumn{5}{l}{\textit{Top-40 most ideology-correlated}} \\
1 & abpoorg & low income--cant afford more children (grid on web) & -0.502 & 0.941 \\
2 & strvbias & priority for criminal justice system & +0.501 & 0.674 \\
3 & absingleg & not married (grid on web) & -0.497 & 0.931 \\
4 & abnomoreg & married--wants no more children (grid on web) & -0.484 & 0.925 \\
5 & poltrthsp & police treats whites better than latinos or latinos better than whites & -0.478 & 0.568 \\
6 & abanyg & abortion if woman wants for any reason (grid on web) & -0.477 & 0.907 \\
7 & leftrght & how left or right in politics & +0.470 & 0.510 \\
8 & defund & move funds for police to social service & -0.465 & 0.646 \\
9 & prochoic & i consider myself pro-choice. & -0.464 & 0.797 \\
10 & racesurv17 & how much, if at all, do you think the legacy of slavery affects the position of & -0.457 & 0.752 \\
11 & poltrtblk & Police treats whites better than Black people & -0.457 & 0.713 \\
12 & ineqmad & feeling about differences in wealth & -0.433 & 0.511 \\
13 & immlimit & amerca should limit immigration to protect way of life & +0.425 & 0.691 \\
14 & prolife & i consider myself pro-life. & +0.421 & 0.388 \\
15 & prespop & approve of pres handling job & -0.417 & 0.443 \\
16 & govfnanc & govt should finance projects to create new jobs & -0.399 & 0.830 \\
17 & clmtcaus & climate has been changing due to human activity & -0.397 & 0.814 \\
18 & govfnaid & govt should give financial assistance to college students from low-income famili & -0.387 & 0.827 \\
19 & religinf & the u.s. would be a better country if religion had less influence & -0.382 & 0.844 \\
20 & prayerv & bible prayer in public schools (with volunteered response on web) & -0.379 & 0.820 \\
21 & buseqinc & pay difference in companies & -0.378 & 0.790 \\
22 & govchrst & the federal government should advocate christian values. & +0.371 & 0.799 \\
23 & tempgen1 & greenhouse effect danger to envir & -0.368 & 0.864 \\
24 & fairdist & how fair is the income distribution in america & +0.363 & 0.599 \\
25 & biblenv & feelings about the bible (no volunteered response on web) & +0.361 & 0.723 \\
26 & abdefectg & strong chance of serious defect (grid on web) & -0.360 & 0.917 \\
27 & trmedia & how much do you trust the news media? & -0.359 & 0.523 \\
28 & upwages & the government should ensure that the wages of low-paying jobs increase as the e & -0.357 & 0.782 \\
29 & grnexagg & environmental threats exaggerated & +0.350 & 0.823 \\
30 & marhomo1 & homosexuals have right to marry & -0.346 & 0.851 \\
31 & grncon & concerned about environment & -0.346 & 0.844 \\
32 & trdunio1 & workers need strong trade unions to protect their interests & -0.344 & 0.640 \\
33 & marhomo & homosexuals should have right to marry & -0.340 & 0.811 \\
34 & abfelegl & women only: women should be able to have legal abortions & -0.338 & 0.881 \\
35 & abrapeg & pregnant as result of rape (grid on web) & -0.336 & 0.917 \\
36 & trresrch & how much do you trust university research centers? & -0.331 & 0.765 \\
37 & biblev & feelings about the bible (with volunteered response on web) & +0.326 & 0.711 \\
38 & smallgap & for a society to be fair, differences in people's standard of living should be s & -0.326 & 0.658 \\
39 & powrorgs & intl orgs take away much power from american govt & +0.319 & 0.764 \\
40 & godusa & the success of the us is part of god's plan & +0.318 & 0.855 \\
\midrule
\multicolumn{5}{l}{\textit{Bottom-40 least ideology-correlated}} \\
1 & spvtrfair & supervisor is fair & -0.000 & 0.883 \\
2 & wlthhsps & rich - poor & +0.000 & 0.693 \\
3 & didlessp & health prevent doing desired activities past 4 & -0.000 & 0.575 \\
4 & poleff13 & have a pretty good understanding of issues & +0.000 & 0.703 \\
5 & nonurse & unable to care for a sick child-relative? & -0.000 & 0.648 \\
6 & pres08 & vote obama or mccain & -0.000 & 0.608 \\
7 & spyenemy & punishment for selling secrets to hostile foreign gov & -0.000 & 0.591 \\
8 & caninf11 & would go to govt site for information on candid & -0.000 & 0.653 \\
9 & rank & r's self ranking of social position & +0.000 & 0.659 \\
10 & governor & did r correctly name governor & +0.000 & 0.667 \\
11 & folk & like or dislike folk music & -0.000 & 0.628 \\
12 & anomia7 & officials not interested in average man & +0.000 & 0.725 \\
13 & investgn & r invested money past year & +0.000 & 0.751 \\
14 & farewhts & self supporting - live off welfare & +0.000 & 0.718 \\
15 & mhhired & x should be hired like others & -0.000 & 0.528 \\
16 & trainops & r have the training opportunities & +0.000 & 0.791 \\
17 & hsrespct & school subjects: respect for authority & -0.000 & 0.493 \\
18 & kidout & govt intervene: child stays out late & +0.000 & 0.604 \\
19 & voteelec & how important always to vote in elections & -0.000 & 0.807 \\
20 & grace & had powerful religious experience & -0.000 & 0.680 \\
21 & finan2 & having a car etc. repossessed & -0.000 & 0.852 \\
22 & draftdef & defense occupations exempt from draft? & +0.000 & 0.563 \\
23 & trust5 & people can or cannot be trusted & -0.001 & 0.573 \\
24 & litcntrl & i have little control over the bad things & +0.001 & 0.442 \\
25 & eff911h & how effective r thinks reducing use of public transportation & +0.001 & 0.741 \\
26 & conurban & farmer's and city people in conflict? & +0.001 & 0.683 \\
27 & swayvote & has r tried to influence votes of others & -0.001 & 0.655 \\
28 & eff911m & how effective r thinks avoiding national landmarks & +0.001 & 0.850 \\
29 & wrkindp & importance of independent work in a job & -0.001 & 0.640 \\
30 & bookmark & how often r uses bookmarks & +0.001 & 0.491 \\
31 & hope6 & i am meeting my current goals & -0.001 & 0.760 \\
32 & chldeduc & education improving? & -0.001 & 0.568 \\
33 & health & condition of health & -0.001 & 0.740 \\
34 & alike1 & how alike: orange \& banana & -0.001 & 0.718 \\
35 & uswar & expect u.s. in war within 10 years & +0.001 & 0.643 \\
36 & firstjob & type of first job held by r & -0.001 & 0.631 \\
37 & painarms & r had pain in the arms in the past 12 months & +0.001 & 0.675 \\
38 & poleff5 & people like me have much to say re govt & +0.001 & 0.663 \\
39 & conghrm1 & how often congregation make too much demand & -0.001 & 0.615 \\
40 & successy & tries hard to succeed-version y & -0.001 & 0.481 \\
\bottomrule
\multicolumn{5}{p{15cm}}{\footnotesize \textit{Notes.} $\rho$ is the signed Spearman correlation with polviews (positive: conservative respondents are more likely to endorse). Alpaca AUC is the out-of-fold retrodiction task AUC per variable.} \\
\end{longtable}
}

%% file: Tables/table_subgroup_delta_imputation.tex
\begin{table}[!htbp]
  \caption{Per-subgroup group-level AUC, Alpaca-7B vs.\ matrix factorization, on the missing-data imputation task.}
  \label{tab:subgroup_groupauc_imputation}
  \centering
  \small
  \renewcommand{\arraystretch}{0.92}
  \begin{tabular}{llrrrr}
    \toprule
    \textbf{Group} & \textbf{Subgroup} & $N$ & \textbf{Alpaca-7B} & \textbf{MF} & $\mathbf{\Delta}$ \textbf{(Alpaca $-$ MF)} \\
     &  &  & (AUC) & (AUC) & (AUC) \\
    \midrule
    \multirow{2}{*}{\textbf{Sex}} & Male & 29,923 & 0.865 & 0.853 & \textbf{+0.012} \\
     & Female & 37,749 & 0.871 & 0.858 & \textbf{+0.013} \\
    \midrule
    \multirow{5}{*}{\textbf{Age}} & 18--29 & 13,694 & 0.859 & 0.846 & \textbf{+0.013} \\
     & 30--44 & 20,647 & 0.870 & 0.858 & \textbf{+0.012} \\
     & 45--59 & 16,241 & 0.870 & 0.859 & \textbf{+0.011} \\
     & 60--74 & 12,035 & 0.873 & 0.860 & \textbf{+0.013} \\
     & 75--89 & 5,055 & 0.868 & 0.853 & \textbf{+0.015} \\
    \midrule
    \multirow{5}{*}{\textbf{Period}} & 1970s & 10,499 & 0.871 & 0.854 & \textbf{+0.017} \\
     & 1980s & 14,102 & 0.868 & 0.854 & \textbf{+0.015} \\
     & 1990s & 13,082 & 0.867 & 0.855 & \textbf{+0.012} \\
     & 2000s & 14,755 & 0.864 & 0.855 & \textbf{+0.008} \\
     & 2010s & 15,234 & 0.872 & 0.859 & \textbf{+0.013} \\
    \midrule
    \multirow{3}{*}{\textbf{Race}} & White & 54,395 & 0.872 & 0.860 & \textbf{+0.013} \\
     & Black & 9,413 & 0.854 & 0.843 & \textbf{+0.011} \\
     & Other & 3,864 & 0.849 & 0.838 & \textbf{+0.011} \\
    \midrule
    \multirow{4}{*}{\textbf{Region}} & Northeast & 12,756 & 0.868 & 0.856 & \textbf{+0.013} \\
     & Midwest & 17,314 & 0.871 & 0.858 & \textbf{+0.013} \\
     & South & 24,103 & 0.867 & 0.855 & \textbf{+0.012} \\
     & West & 13,499 & 0.868 & 0.856 & \textbf{+0.012} \\
    \midrule
    \multirow{5}{*}{\textbf{Education}} & Less than high school & 13,629 & 0.853 & 0.839 & \textbf{+0.014} \\
     & High school graduate & 34,365 & 0.865 & 0.853 & \textbf{+0.012} \\
     & Some college & 3,969 & 0.869 & 0.858 & \textbf{+0.011} \\
     & Bachelor's & 10,342 & 0.881 & 0.869 & \textbf{+0.012} \\
     & Graduate & 5,367 & 0.888 & 0.876 & \textbf{+0.012} \\
    \midrule
    \multirow{6}{*}{\textbf{Income}} & Q1 (lowest) & 13,750 & 0.855 & 0.842 & \textbf{+0.013} \\
     & Q2 & 12,796 & 0.863 & 0.851 & \textbf{+0.012} \\
     & Q3 & 13,319 & 0.870 & 0.858 & \textbf{+0.012} \\
     & Q4 & 11,768 & 0.875 & 0.863 & \textbf{+0.012} \\
     & Q5 (highest) & 9,509 & 0.881 & 0.869 & \textbf{+0.012} \\
     & Missing & 6,530 & 0.865 & 0.852 & \textbf{+0.013} \\
    \midrule
    \multirow{8}{*}{\textbf{Party ID}} & Independent & 10,510 & 0.860 & 0.849 & \textbf{+0.012} \\
     & Strong Democrat & 11,067 & 0.868 & 0.856 & \textbf{+0.012} \\
     & Weak Democrat & 13,731 & 0.867 & 0.855 & \textbf{+0.013} \\
     & Lean Democrat & 8,203 & 0.868 & 0.855 & \textbf{+0.013} \\
     & Lean Republican & 5,988 & 0.868 & 0.856 & \textbf{+0.013} \\
     & Weak Republican & 10,251 & 0.873 & 0.861 & \textbf{+0.012} \\
     & Strong Republican & 6,759 & 0.876 & 0.863 & \textbf{+0.013} \\
     & Something else & 1,163 & 0.874 & 0.861 & \textbf{+0.013} \\
    \bottomrule
  \end{tabular}

  \medskip
  \parbox{\linewidth}{\footnotesize \textit{Notes.} $\Delta$ (Alpaca $-$ MF) is the per-subgroup difference in group-level AUC, where group-level AUC is computed by pooling held-out predictions within each subgroup. Bolded $\Delta$ values highlight where the proposed approach improves over the matrix-factorization baseline.}
\end{table}

%% file: Tables/table_subgroup_delta_retrodiction.tex
\begin{table}[!htbp]
  \caption{Per-subgroup group-level AUC, Alpaca-7B vs.\ matrix factorization, on the retrodiction task.}
  \label{tab:subgroup_groupauc_retrodiction}
  \centering
  \small
  \renewcommand{\arraystretch}{0.92}
  \begin{tabular}{llrrrr}
    \toprule
    \textbf{Group} & \textbf{Subgroup} & $N$ & \textbf{Alpaca-7B} & \textbf{MF} & $\mathbf{\Delta}$ \textbf{(Alpaca $-$ MF)} \\
     &  &  & (AUC) & (AUC) & (AUC) \\
    \midrule
    \multirow{2}{*}{\textbf{Sex}} & Male & 29,923 & 0.828 & 0.732 & \textbf{+0.097} \\
     & Female & 37,749 & 0.835 & 0.735 & \textbf{+0.100} \\
    \midrule
    \multirow{5}{*}{\textbf{Age}} & 18--29 & 13,694 & 0.824 & 0.730 & \textbf{+0.094} \\
     & 30--44 & 20,647 & 0.833 & 0.734 & \textbf{+0.099} \\
     & 45--59 & 16,241 & 0.834 & 0.738 & \textbf{+0.096} \\
     & 60--74 & 12,035 & 0.837 & 0.734 & \textbf{+0.103} \\
     & 75--89 & 5,055 & 0.834 & 0.725 & \textbf{+0.109} \\
    \midrule
    \multirow{5}{*}{\textbf{Period}} & 1970s & 10,499 & 0.863 & 0.822 & \textbf{+0.040} \\
     & 1980s & 14,102 & 0.838 & 0.734 & \textbf{+0.103} \\
     & 1990s & 13,082 & 0.818 & 0.715 & \textbf{+0.103} \\
     & 2000s & 14,755 & 0.818 & 0.708 & \textbf{+0.110} \\
     & 2010s & 15,234 & 0.841 & 0.749 & \textbf{+0.092} \\
    \midrule
    \multirow{3}{*}{\textbf{Race}} & White & 54,395 & 0.836 & 0.738 & \textbf{+0.098} \\
     & Black & 9,413 & 0.819 & 0.718 & \textbf{+0.101} \\
     & Other & 3,864 & 0.814 & 0.716 & \textbf{+0.097} \\
    \midrule
    \multirow{4}{*}{\textbf{Region}} & Northeast & 12,756 & 0.833 & 0.735 & \textbf{+0.098} \\
     & Midwest & 17,314 & 0.834 & 0.736 & \textbf{+0.099} \\
     & South & 24,103 & 0.831 & 0.731 & \textbf{+0.099} \\
     & West & 13,499 & 0.832 & 0.734 & \textbf{+0.098} \\
    \midrule
    \multirow{5}{*}{\textbf{Education}} & Less than high school & 13,629 & 0.823 & 0.727 & \textbf{+0.096} \\
     & High school graduate & 34,365 & 0.828 & 0.731 & \textbf{+0.097} \\
     & Some college & 3,969 & 0.830 & 0.730 & \textbf{+0.100} \\
     & Bachelor's & 10,342 & 0.843 & 0.743 & \textbf{+0.101} \\
     & Graduate & 5,367 & 0.850 & 0.747 & \textbf{+0.104} \\
    \midrule
    \multirow{6}{*}{\textbf{Income}} & Q1 (lowest) & 13,750 & 0.821 & 0.719 & \textbf{+0.102} \\
     & Q2 & 12,796 & 0.827 & 0.730 & \textbf{+0.097} \\
     & Q3 & 13,319 & 0.834 & 0.737 & \textbf{+0.097} \\
     & Q4 & 11,768 & 0.837 & 0.739 & \textbf{+0.098} \\
     & Q5 (highest) & 9,509 & 0.845 & 0.747 & \textbf{+0.098} \\
     & Missing & 6,530 & 0.829 & 0.731 & \textbf{+0.098} \\
    \midrule
    \multirow{8}{*}{\textbf{Party ID}} & Independent & 10,510 & 0.825 & 0.730 & \textbf{+0.095} \\
     & Strong Democrat & 11,067 & 0.833 & 0.729 & \textbf{+0.104} \\
     & Weak Democrat & 13,731 & 0.833 & 0.734 & \textbf{+0.098} \\
     & Lean Democrat & 8,203 & 0.833 & 0.735 & \textbf{+0.098} \\
     & Lean Republican & 5,988 & 0.831 & 0.735 & \textbf{+0.096} \\
     & Weak Republican & 10,251 & 0.835 & 0.738 & \textbf{+0.097} \\
     & Strong Republican & 6,759 & 0.836 & 0.735 & \textbf{+0.102} \\
     & Something else & 1,163 & 0.837 & 0.742 & \textbf{+0.095} \\
    \bottomrule
  \end{tabular}

  \medskip
  \parbox{\linewidth}{\footnotesize \textit{Notes.} $\Delta$ (Alpaca $-$ MF) is the per-subgroup difference in group-level AUC, where group-level AUC is computed by pooling held-out predictions within each subgroup. Bolded $\Delta$ values highlight where the proposed approach improves over the matrix-factorization baseline.}
\end{table}

%% file: Figures/figure_a_variable_selection_block.tex
\begin{figure}[ht]
  \begin{center}
      \centering
     \includegraphics[width=1\columnwidth]{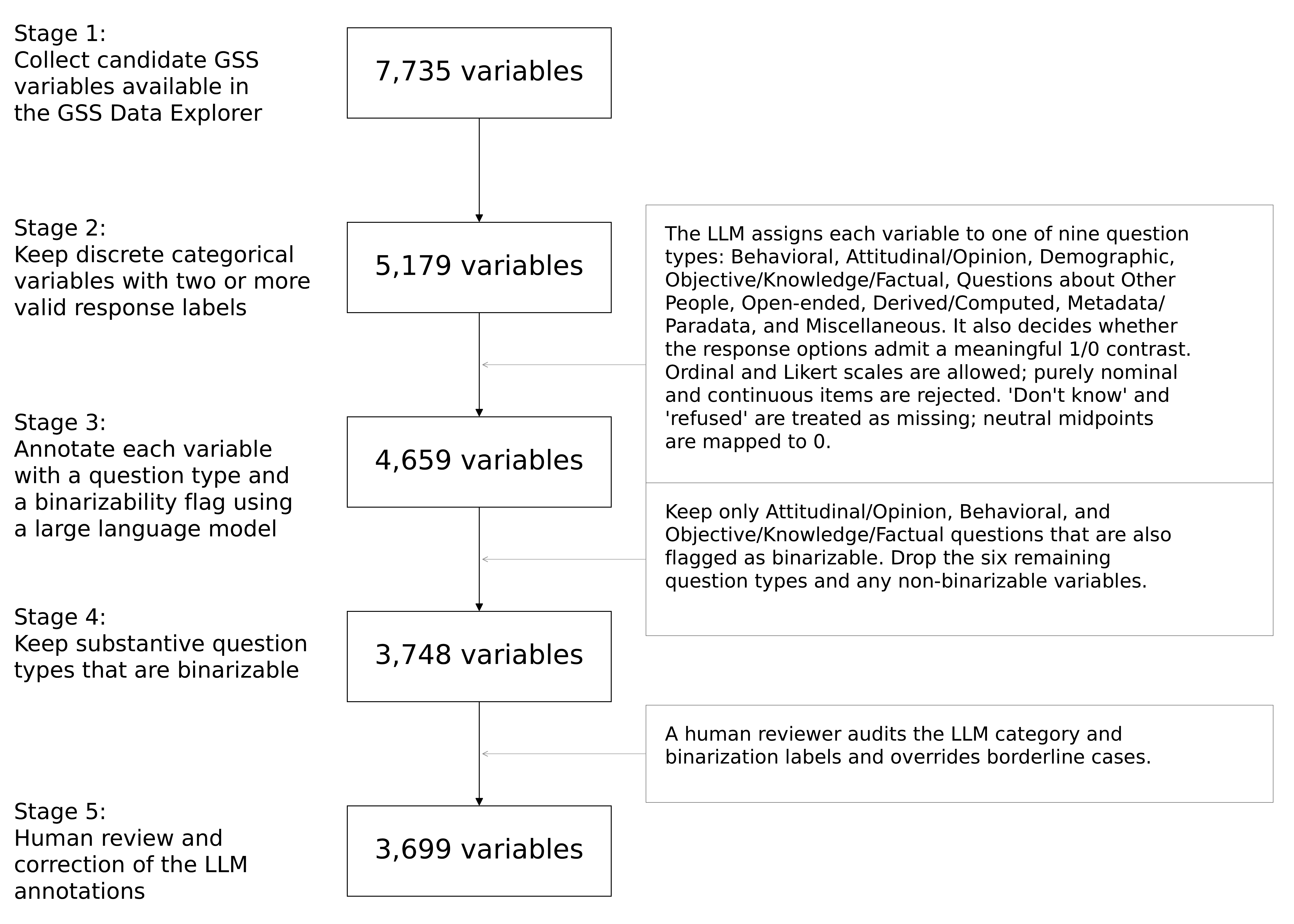}
  \end{center}
  \caption{\textbf{Variable selection process.}}
  \label{fig:figure_a_variable_selection}
\end{figure}

%% file: Figures/figure_a_embeddings_block.tex
\begin{figure}[ht]
  \begin{center}
     \includegraphics[width=1\columnwidth]{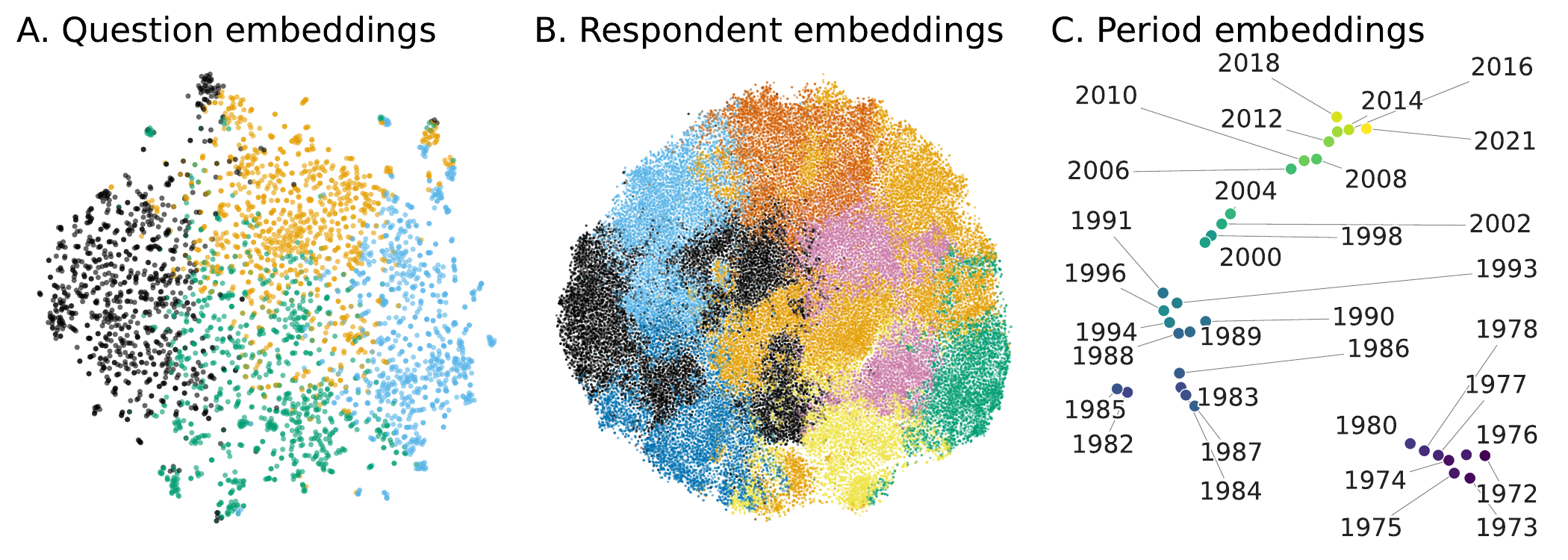}
  \end{center}
    \caption{\textbf{Visualization of the learned question, respondent, and period embeddings.} Panel A shows a two-dimensional t-SNE projection of the fine-tuned 50-dimensional question embeddings for all. Panel B shows the t-SNE projection of the 68{,}846 respondent embeddings (one point per respondent), with points coloured by $k$-means cluster ($k = 10$). Panel C shows the t-SNE projection of the 33 period embeddings, one per General Social Survey wave from 1972 to 2021; each point is labeled with its survey year.}
  \label{fig:figure3}
\end{figure}

%% file: Figures/figure_a_architecture_detail_block.tex
\begin{figure}[!t]
  \begin{center}
      \centering
     \includegraphics[width=0.7\columnwidth]{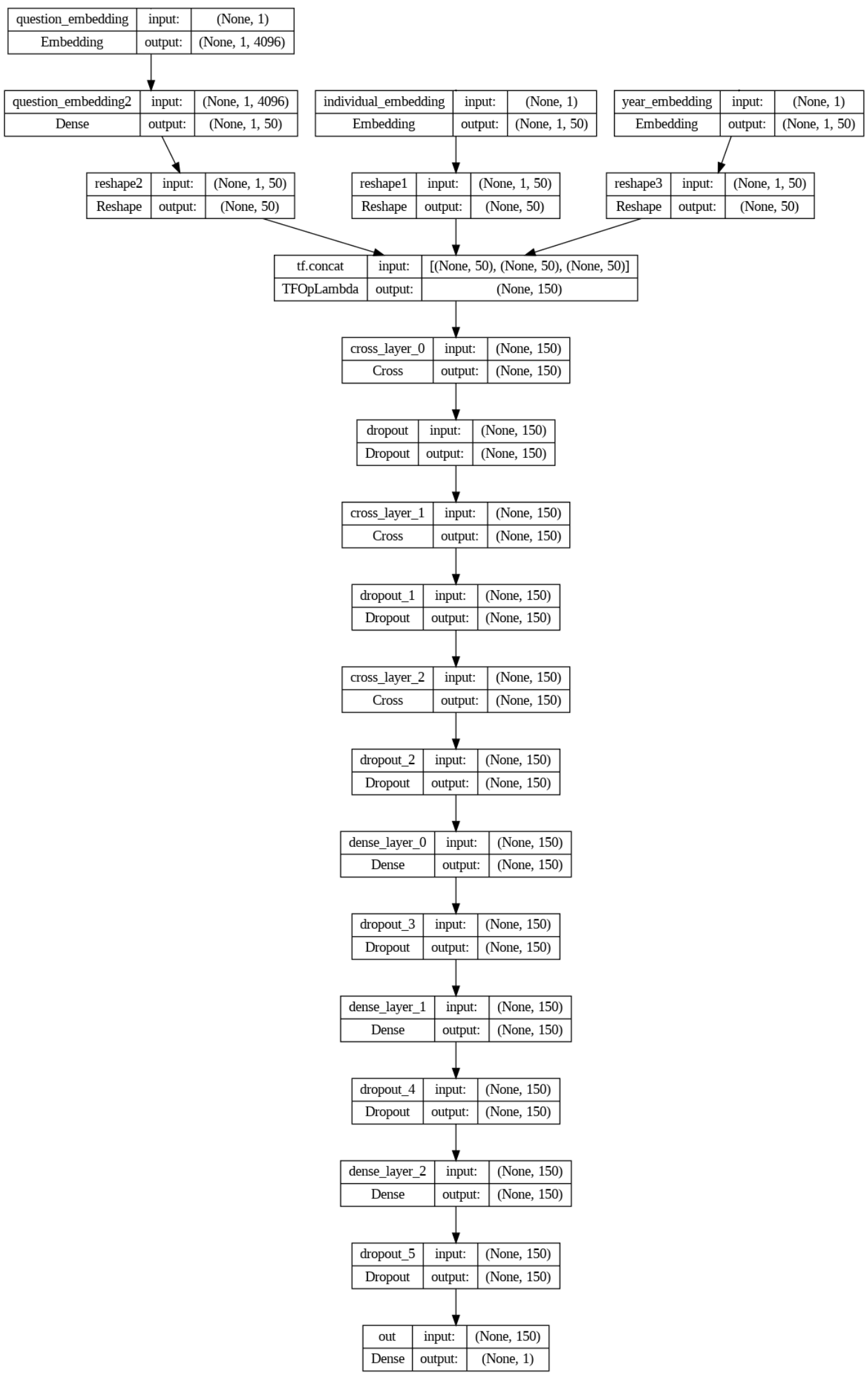}
  \end{center}
  \caption{\textbf{Model Architecture.} Here, we present input and output dimensions for each layer: question\_embedding = survey question embedding, individual\_embedding = respondent embedding, period\_embedding = period embedding, Cross= cross layers, Dense= feed-forward dense layers.}
  \label{fig:figure_a_architecture_detail}
\end{figure}

%% file: Figures/figure_a_demographic_cv_block.tex
\begin{figure}[!t]
  \begin{center}
      \centering
     \includegraphics[width=1\columnwidth]{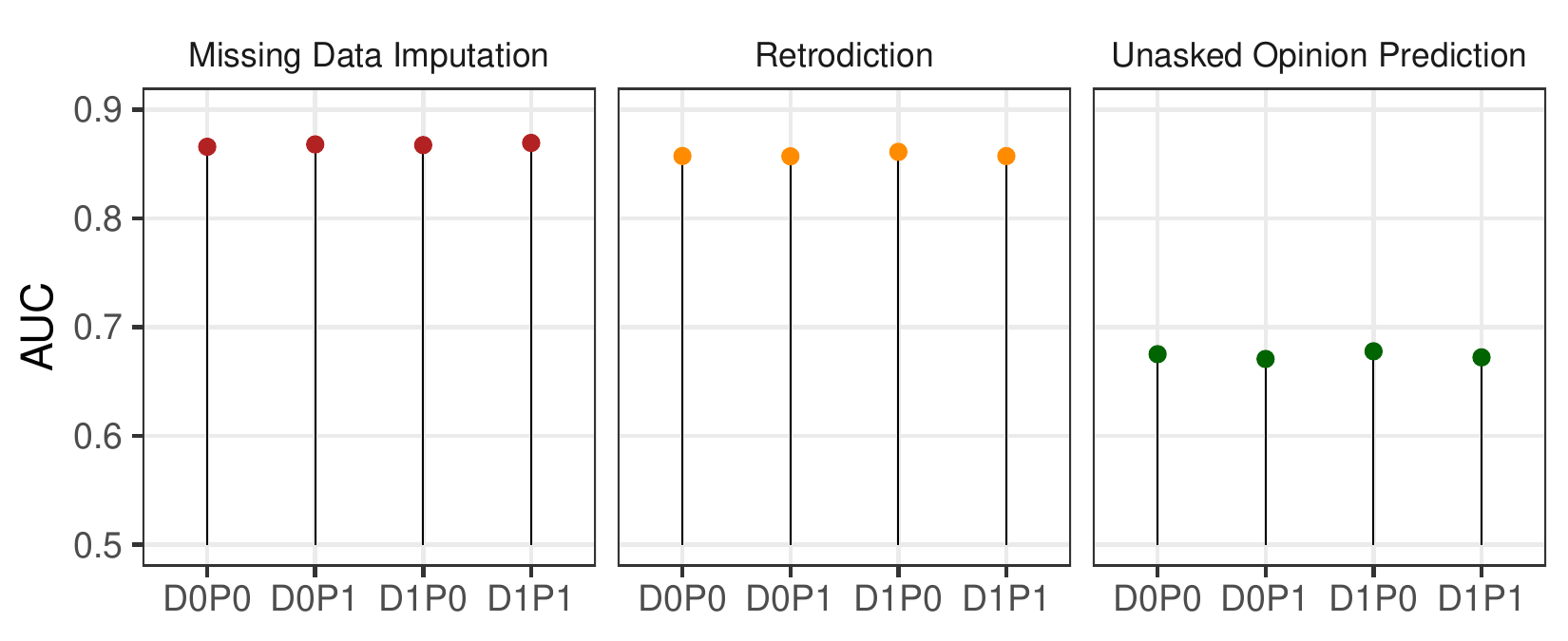}
  \end{center}
  \caption{\textbf{Model performance with or without demographics or partisanship information.} We use the personalized LLMs based on Alpaca-7b to measure the AUC of models for missing data imputation, retrodiction, and unasked opinion prediction using one of the ten folds in the 10-fold cross-validation scheme. The notation D1P1 indicates that demographic information (i.e., age, cohort, gender, race, education, income, and religion) and partisanship information (i.e., political ideology, party affiliation) are used as training data. D1P0 indicates that demographic information is used, but partisanship information is not. D0P1 indicates that partisanship information is used, but demographic information is not. Finally, D0P0 indicates that neither demographic nor partisanship information is used in the model training.}
  \label{fig:figure_a_demographic_cv}
\end{figure}

%% file: Figures/figure_a_cv_scheme_block.tex
\begin{figure}[!t]
  \begin{center}
      \centering
     \includegraphics[width=1\columnwidth]{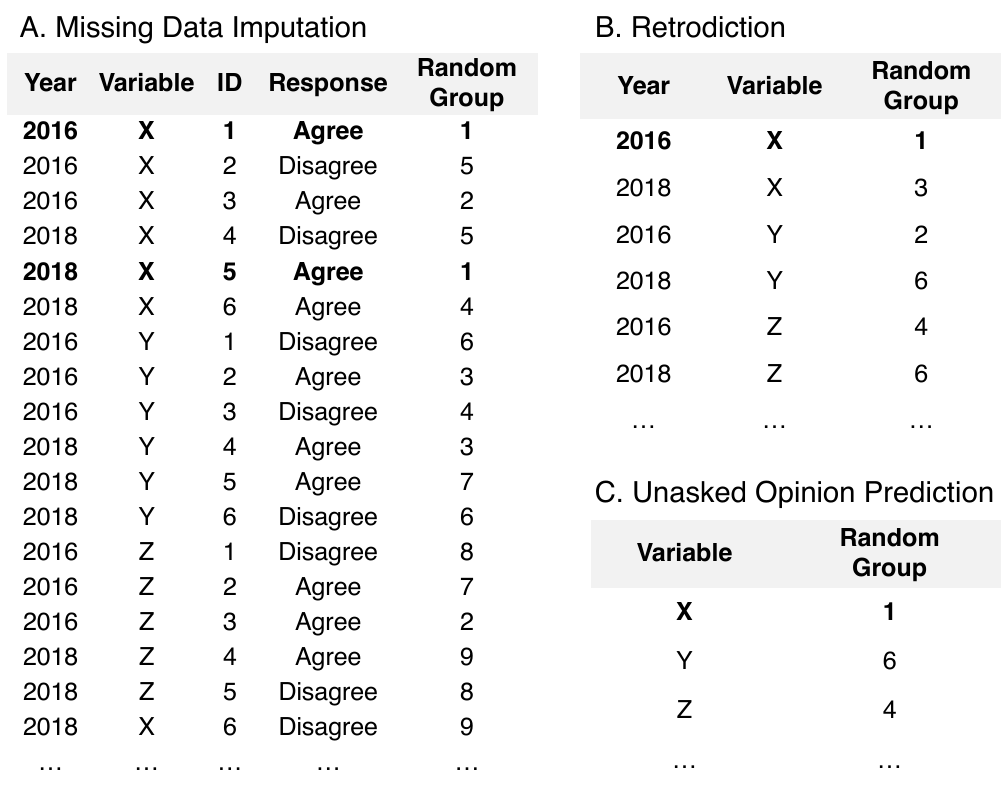}
  \end{center}
  
  \caption{\textbf{Examples of 10-fold cross-validation scheme.} In Panel A, when predicting response-level missing opinions, we randomly allocate the combinations of survey years, variables, and individual IDs into ten groups, which are held out in each round of cross-validation. In Panel B, when retrodicting year-level missing opinions, we randomly assign pairs of survey years and variables into ten groups, which are held out in each round of cross-validation. In Panel C, when predicting completely missing opinions, we randomly assign variables into ten groups, which are held out in each round of cross-validation. Data in bold cells are excluded in the first round of cross-validation. By iterating these processes ten times, we can fill in all observed as well as unobserved responses with predicted values that arises when those responses are missing because of different reasons.}
  \label{fig:cv_scheme_a}

\end{figure}

%% file: Figures/figure_a_regime_schematic_block.tex
\begin{figure}[!htbp]
  \begin{center}
    \includegraphics[width=0.9\columnwidth]{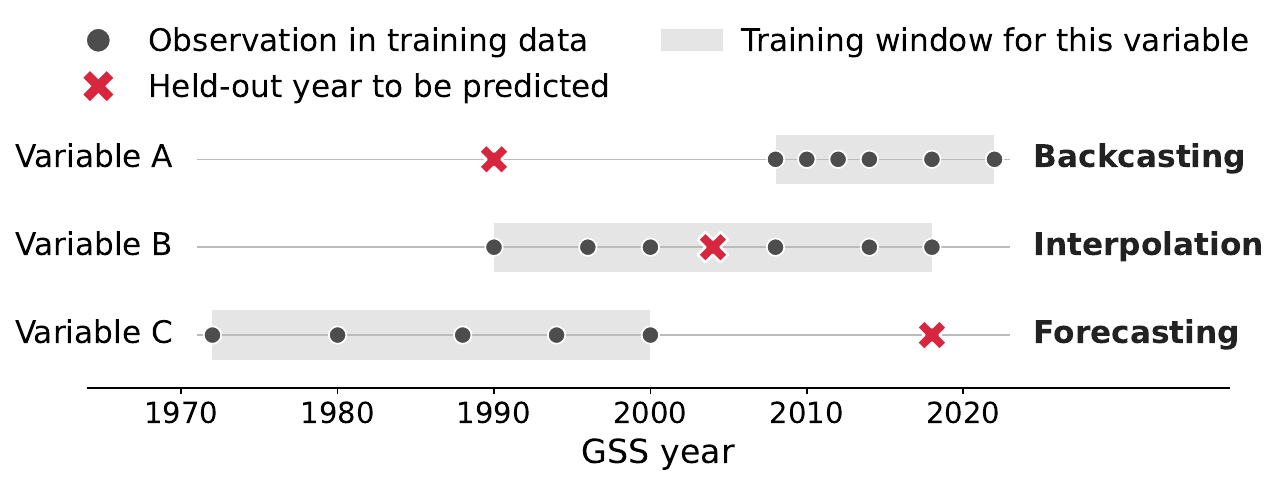}
    \caption{\textbf{Schematic of the three retrodiction scenarios.} Each row is one example variable. Gray dots are observations that remain in the training data; the red marker is the held-out year for that variable, to be predicted in the validation data. The shaded band spans the training window (from the earliest to the latest remaining training year for that variable). \emph{Backcasting}: the held-out year falls before the training window. \emph{Interpolation}: it falls inside. \emph{Forecasting}: it falls after.}
    \label{fig:regime_schematic}
  \end{center}
\end{figure}

%% file: Figures/figure_a_missing_mech_block.tex
\begin{figure}[!t]
  \begin{center}
      \centering
     \includegraphics[width=1\columnwidth]{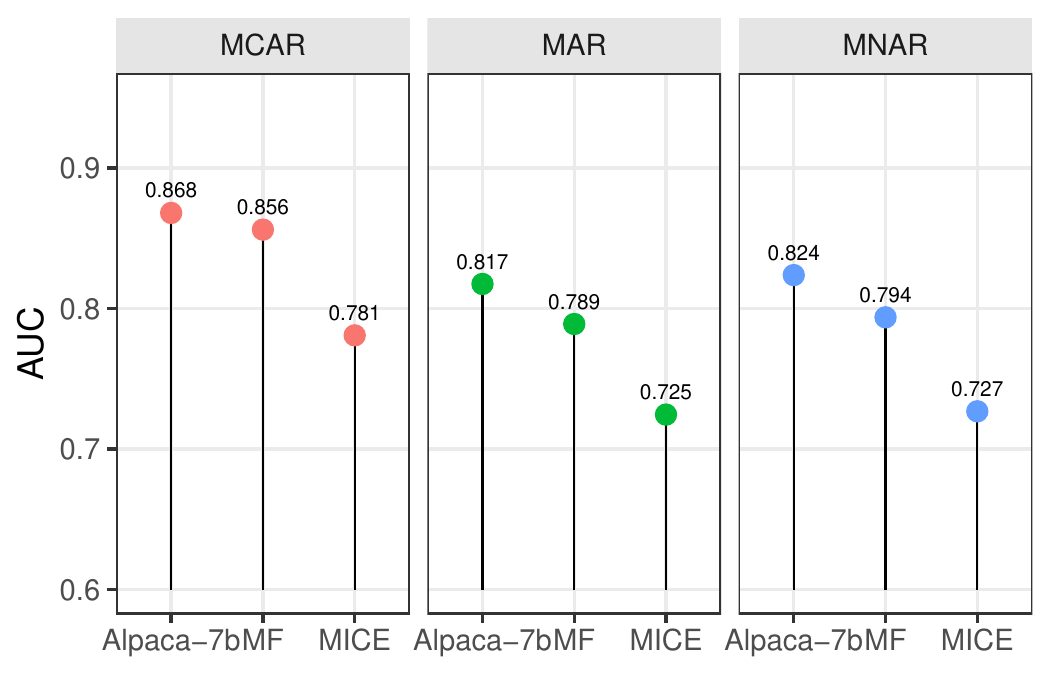}
  \end{center}
  \caption{\textbf{Performance of Alpaca-7b, matrix factorization, and MICE models for missing data imputation by different missing mechanisms.} AUC (Area Under Curve) measures the performance of each model in predicting data that are MCAR (missing completely at random), MAR (missing at random), and MNAR (missing not at random), as indicated by each set of bars. Under MCAR, our Alpaca-7B model reaches AUC $=0.868$, compared with $0.856$ for matrix factorization and $0.781$ for MICE. Under MAR, all three approaches degrade, but the ranking is preserved: Alpaca-7B reaches AUC $=0.817$, matrix factorization $0.789$, and MICE $0.725$. The same ordering holds under MNAR: Alpaca-7B $=0.824$, matrix factorization $0.794$, and MICE $0.727$. MF: Matrix factorization. MICE: Multiple Imputation by Chained Equations.}
  \label{fig:figure_a_missing_mech}
\end{figure}

%% file: Figures/figure_a_missing_prop_block.tex
\begin{figure}[!t]
  \begin{center}
      \centering
     \includegraphics[width=1\columnwidth]{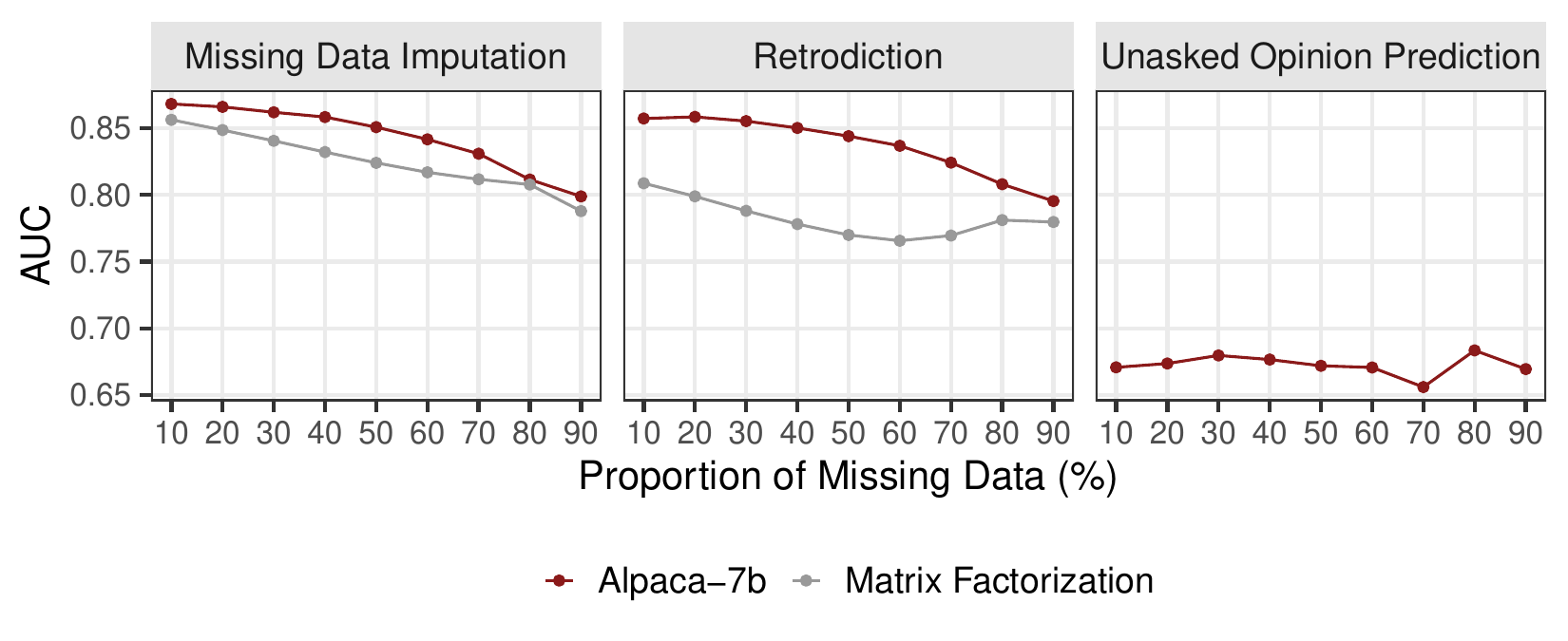}
  \end{center}
  \caption{\textbf{Model performance by the proportion of missing data in training data (10\% to 90\%).} X-axis indicates the proportion of missing data in the training data. For instance, 10\% indicates that only 10\% of the existing data has been used to train the model. Y-axis indicates the AUC values for missing data imputation, retrodiction, and unasked opinion prediction which are estimated using one of the ten folds in the 10-fold cross-validation scheme.}
  \label{fig:figure_a_missing_prop}
\end{figure}

%% file: Figures/figure_a_retro_interp_forecast_block.tex
\begin{figure}[!htbp]
  \begin{center}
    \includegraphics[width=0.9\columnwidth]{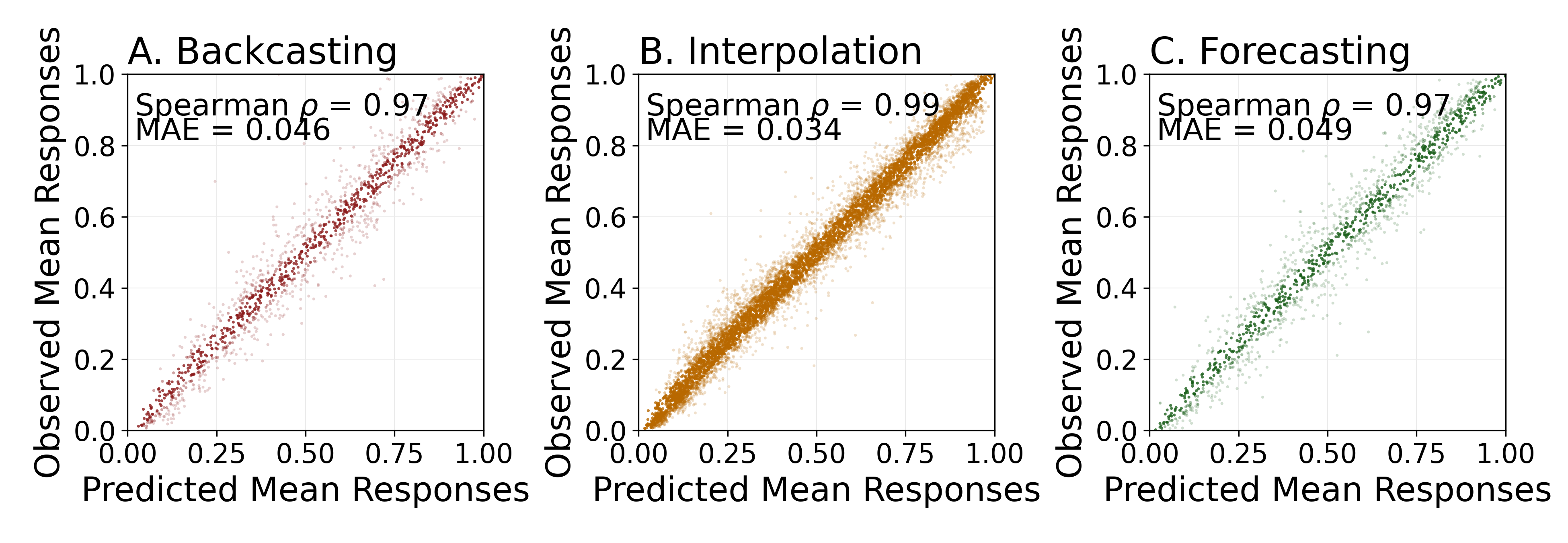}
\caption{\textbf{Aggregate-level performance across three temporal prediction scenarios: backcasting, interpolation, and forecasting.} Each held-out $(\text{variable}, \text{year})$ pair is classified into one of three scenarios based on the training years available for that variable. \emph{Interpolation} refers to predicting responses for years between observed survey years, \emph{backcasting} refers to predicting responses for years before the earliest observed survey year, and \emph{forecasting} refers to predicting responses for years after the latest observed survey year. Panels report the relationship between observed and predicted mean responses for each year per each variable.}    \label{fig:retro_interp_forecast}
  \end{center}
\end{figure}

%% file: Figures/figure_a_distance_nearest_year_dummy_block.tex
\begin{figure}[!htbp]
  \begin{center}
    \includegraphics[width=0.9\columnwidth]{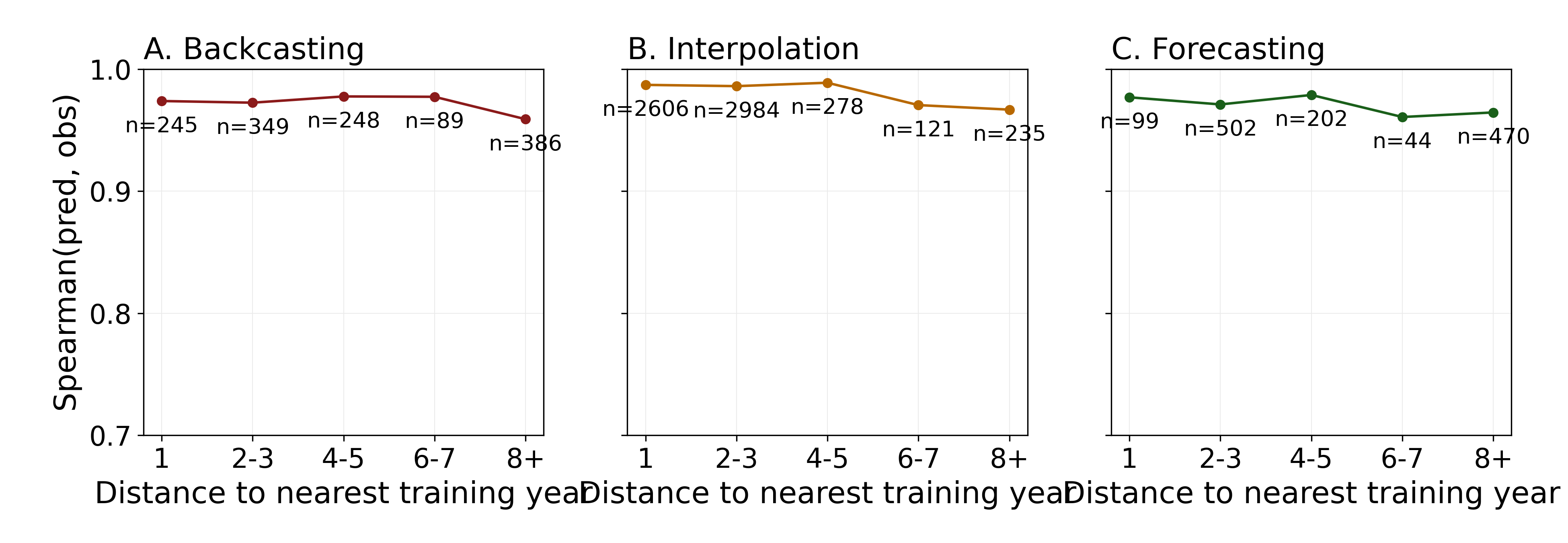}
    \caption{\textbf{Spearman correlation between Alpaca-predicted and GSS-observed mean responses, by distance to the nearest training year.} Held-out (variable, year) pairs are grouped by their distance, in years, to the nearest year in which the GSS fielded that variable in the training data (bins: 1, 2--3, 4--5, 6--7, 8+). Within each subgroup, the Spearman correlation between Alpaca-7B-predicted and GSS-observed mean responses is reported separately for the three retrodiction regimes (backcasting, interpolation, forecasting). The correlation remains above $0.95$ across all distance bins for all three regimes, indicating that predictability is approximately preserved even far from the training window.}
    \label{fig:distance_nearest_year_dummy}
  \end{center}
\end{figure}

%% file: Figures/figure_a_module_removed_block.tex
\begin{figure}[!h]
  \begin{center}
      \centering
     \includegraphics[width=0.86\columnwidth]{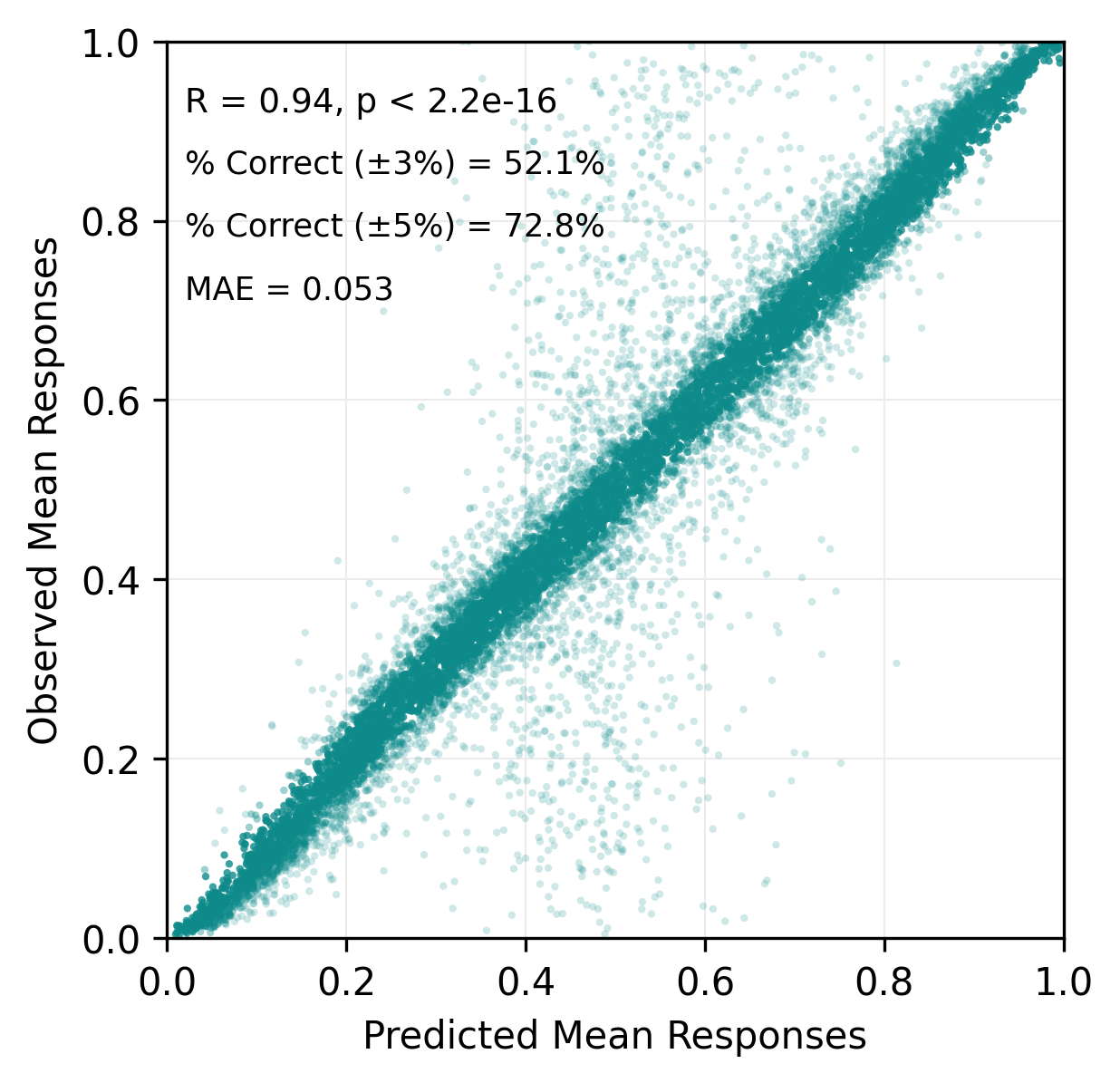}
  \caption{\textbf{Model performance for predicting missing responses at aggregate levels after module removal.} Scatter plot of observed mean responses against the Alpaca-7b prediction under the module-grouped 10-fold cross-validation procedure, restricted to multi-year modules. Each point is a (variable, year) cell; point transparency reflects the magnitude of the prediction error. Annotations report the Pearson correlation $R$, the percentage of cells predicted within $\pm 3\%$ and $\pm 5\%$ of the observed mean, and the mean absolute error (MAE).}
  \label{fig:figure_a_module_removed}
  \end{center}
\end{figure}

%% file: Figures/figure_a_samesex_exclude_block.tex
\begin{figure}[!t]
  \begin{center}
      \centering
     \includegraphics[width=1\columnwidth]{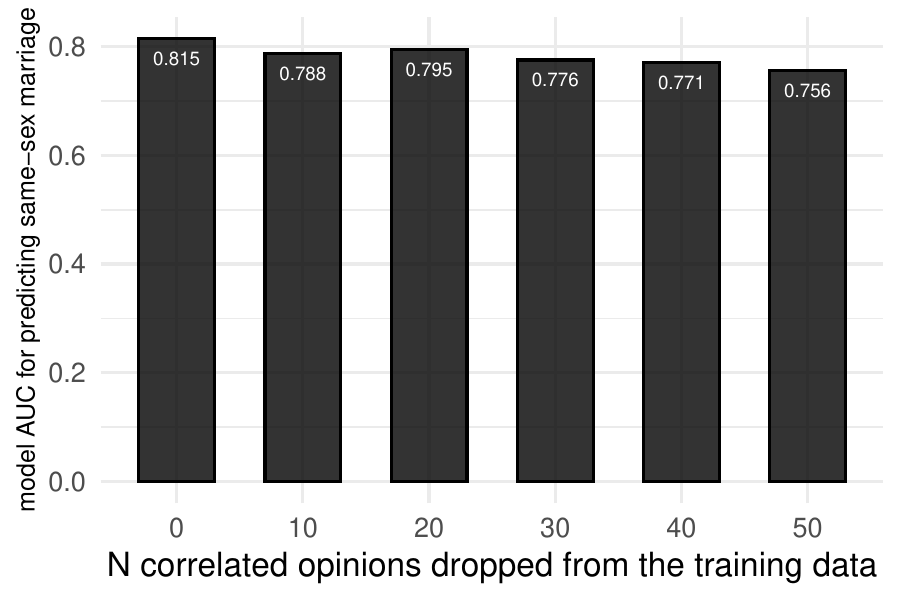}
  \end{center}
    \caption{\textbf{Impact of Excluding Highly Correlated Variables on Model Performance Predicting Opinion on Same-Sex Marriage.} The figure illustrates the Area Under the Curve (AUC) of the Alpaca-7b retrodiction model after systematically excluding the top-$N$ survey questions most correlated with the target variable (same-sex marriage opinion; marhomo1) from the training data. The x-axis indicates the number of excluded top-correlated variables, and the y-axis indicates the resulting AUC. Even after removing the most strongly correlated variables, the model's AUC degrades only marginally, indicating that predictions do not hinge on a small set of direct proxies for the target opinion.}
  \label{fig:figure_a_samesex_exclude}
\end{figure}

%% file: Figures/figure_a_groupvar_between_block.tex
\begin{figure}[ht]
\centering
\includegraphics[width=\linewidth]{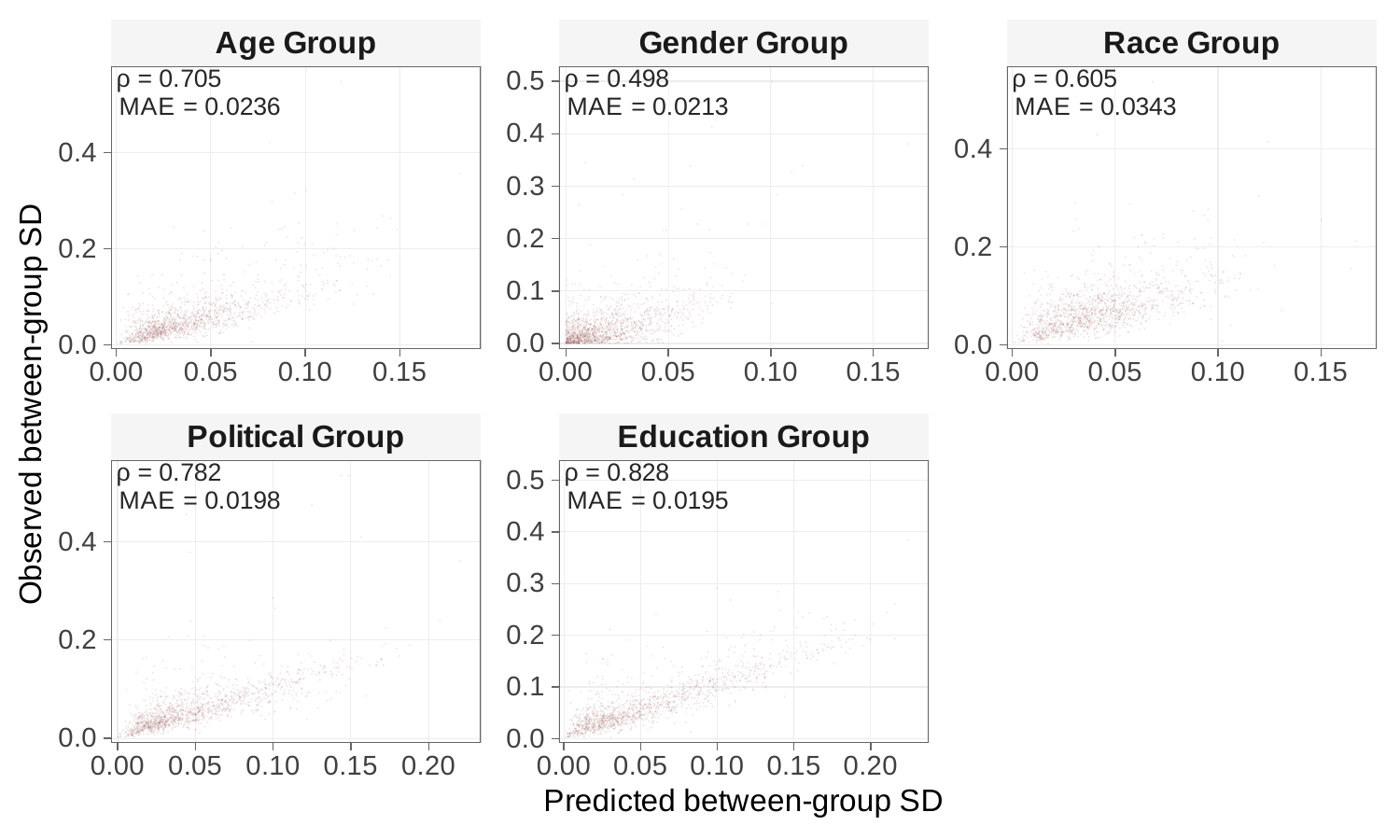}
\caption{\textbf{Between-group variance results across key demographic categories.} Each panel plots the standard deviation of Alpaca-7b predicted group means (x-axis) against the standard deviation of observed group means (y-axis) for one demographic grouping (Age Group, Gender, Race, Political, Education). Each point is one GSS variable; Panels report the Spearman correlation $\rho$ and the mean absolute error (MAE) summarising agreement between predicted and observed between-group spreads.}
\label{fig:between_group_variance}
\end{figure}

%% file: Figures/figure_a_groupvar_within_block.tex
\begin{figure}[ht]
\centering
\includegraphics[width=\linewidth]{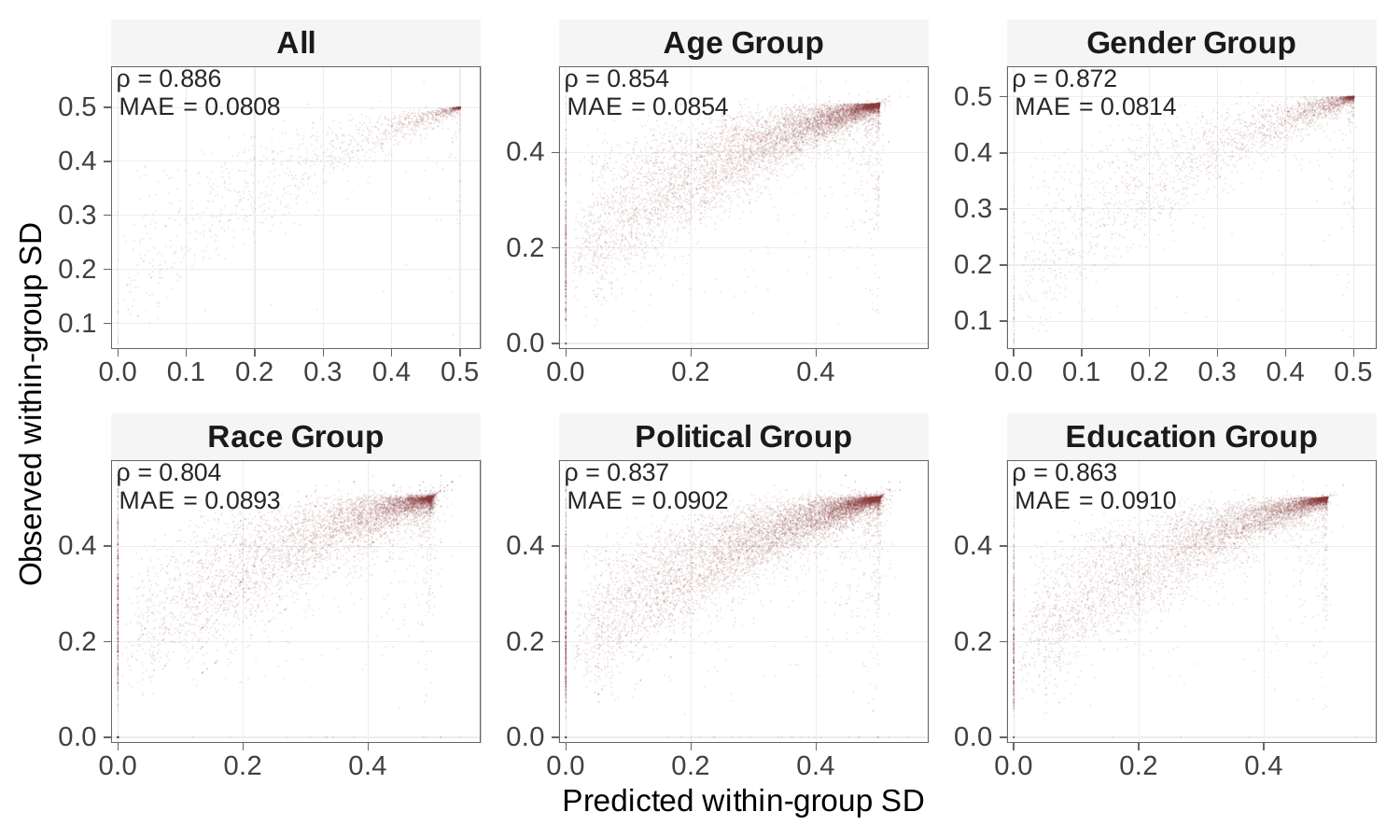}
\caption{\textbf{Within-group variance results across key demographic categories.} Each panel plots the standard deviation of Alpaca-7B predicted responses within demographic subgroups (x-axis) against the standard deviation of observed responses within the same subgroups (y-axis), for each demographic grouping (Age Group, Gender, Race, Political, Education) plus an overall ``All'' panel that uses the full sample. Each point is a (variable~$\times$~subgroup) cell; within-group standard deviations are computed unweighted. Panels report the Spearman correlation $\rho$ and the mean absolute error (MAE).}
\label{fig:within_group_variance}
\end{figure}

%% file: Figures/figure_a_polviews_rank_block.tex
\begin{figure}[!htbp]
  \begin{center}
    \includegraphics[width=0.7\columnwidth]{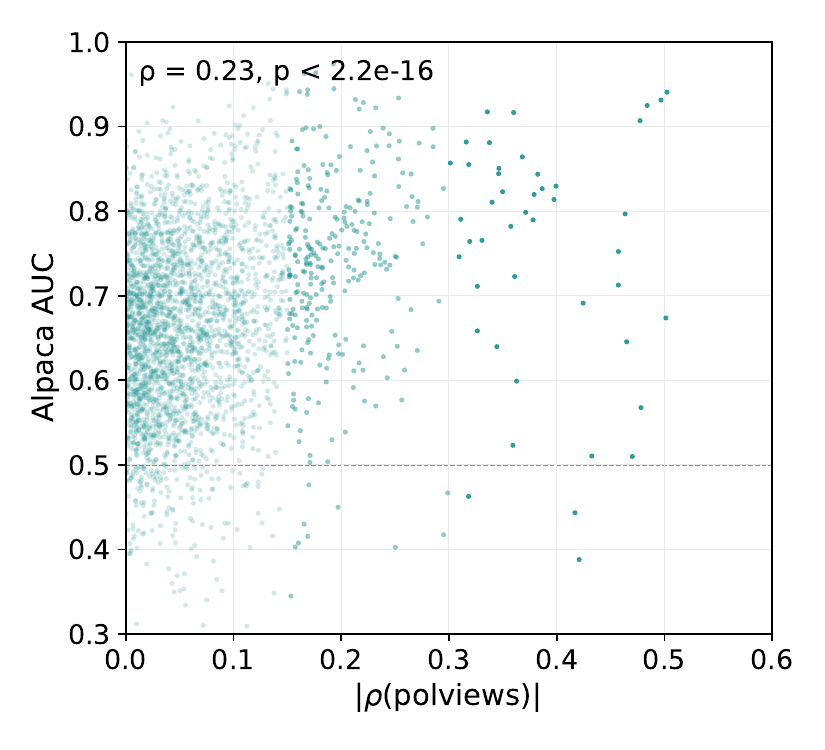}
    \caption{\textbf{Ideological correlation vs.\ Alpaca-7B retrodiction AUC.} Each point is one GSS variable. The $x$-axis is the absolute Spearman correlation of the binarized response with the 7-point \texttt{polviews} ideology scale; the $y$-axis is Alpaca retrodiction AUC. The Spearman rank correlation between $|\rho|$ and AUC is $0.225$.}
    \label{fig:polviews_rank_scatter}
  \end{center}
\end{figure}

%% file: Figures/figure_a_response_category_block.tex
\begin{figure}[htp]
    \centering
    \includegraphics[width=\textwidth]{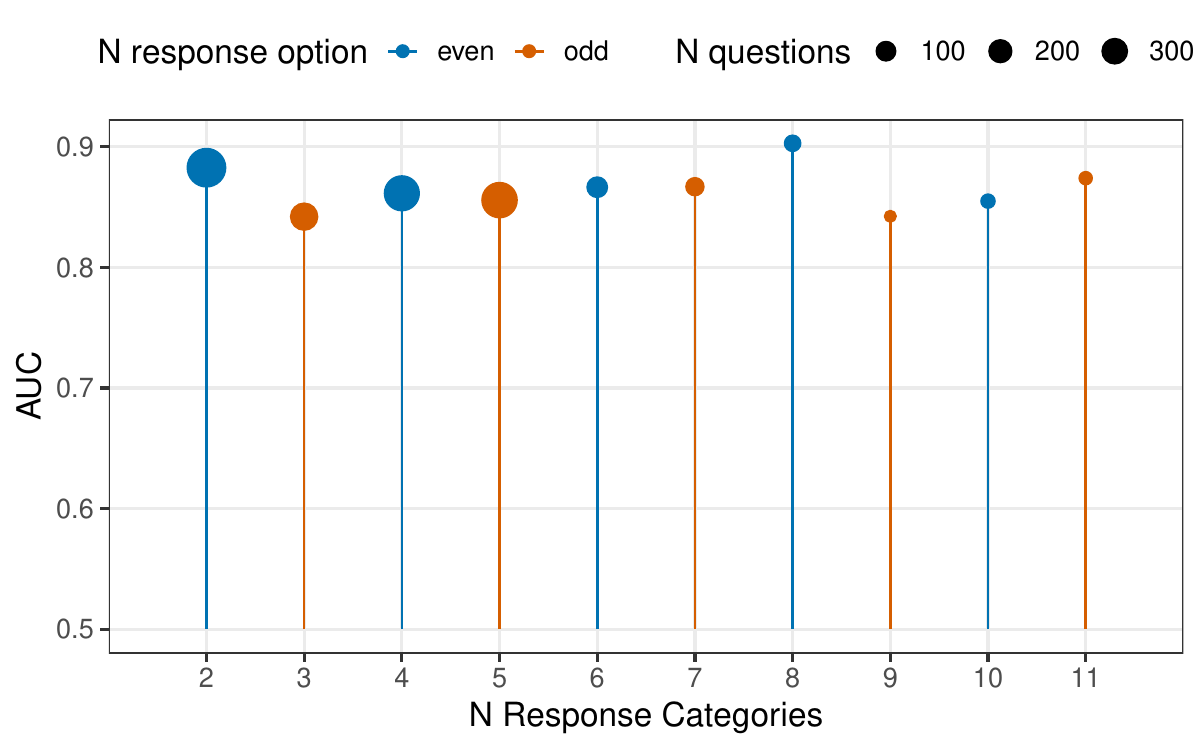}
    \caption{\textbf{Mean AUC by number of response categories and parity (even vs. odd)}.
    The shape and color indicate whether the number of response options is even (blue) or odd (orange). Each point's vertical line extends down to the chance baseline (0.50), and the circle at the top represents the mean AUC for that combination of conditions. The size of each circle indicates the number of questions utilized in that scenario, with larger circles representing more questions (e.g., 100, 200, or 300). Higher points correspond to better predictability (higher AUC) of the model or measurement method under the given conditions.}
    \label{fig:mean_auc_response_categories}
    
\end{figure}

%% file: Figures/figure_a_auc_by_agreement_block.tex
\begin{figure}[htp]
    \centering
    \includegraphics[width=0.6\textwidth]{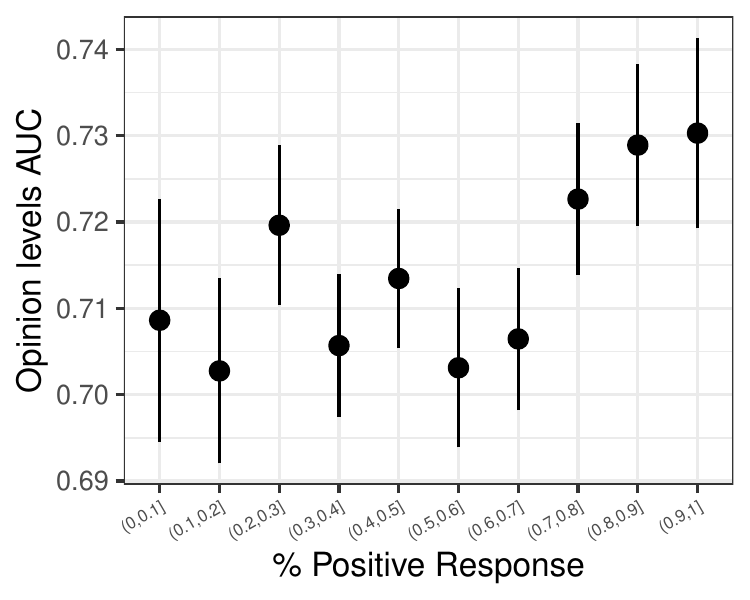}
    
    \caption{\textbf{Mean AUC by Percentage of Positive Responses.}
    This figure shows how the model's predictive performance (AUC) varies as a function of the proportion of respondents giving a positive response to a particular opinion. Each point represents the mean AUC within a given range of the positive response rate, and vertical lines denote the 95\% confidence intervals.}
    \label{fig:auc_positive_responses}
\end{figure}

%% file: Figures/figure_a_n_questions_block.tex
\begin{figure}[!t]
  \begin{center}
      \centering
     \includegraphics[width=1\columnwidth]{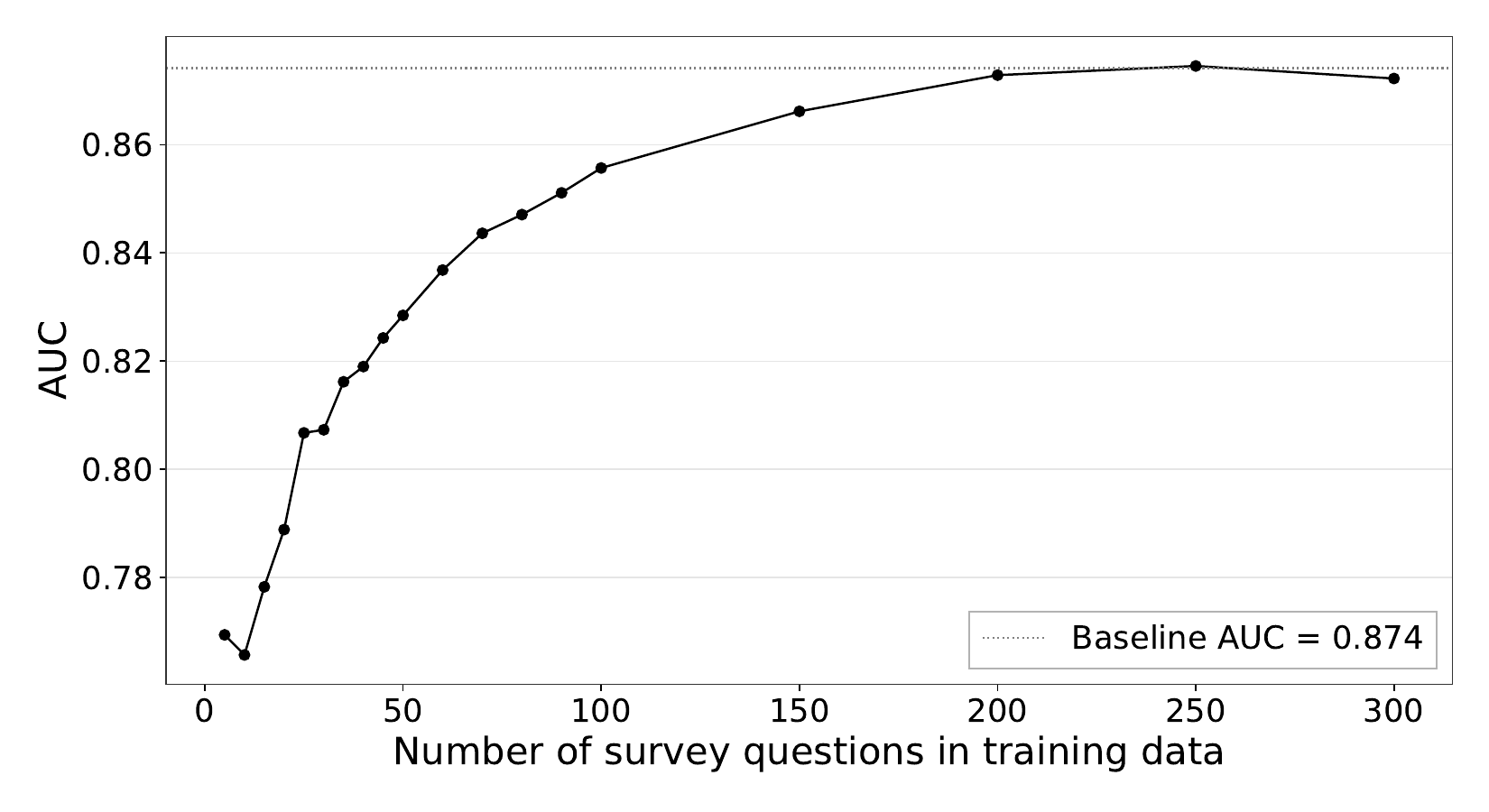}
  \end{center}
  \caption{\textbf{Model performance by the number of survey questions in the training data.} The Y-axis of the graph represents the AUC score, which is an indicator of how well the model can accurately fill in response-level missing opinion data. The X-axis displays the number of survey questions in the training data. We only use survey participants in 2016, 2018, and 2021 for this analysis.}
  \label{fig:figure_a_n_questions}
\end{figure}

%% file: Figures/fig_figure_years_vs_auc_block.tex
\begin{figure*}[!t]
  \begin{center}
      \centering
     \includegraphics[width=1\columnwidth]{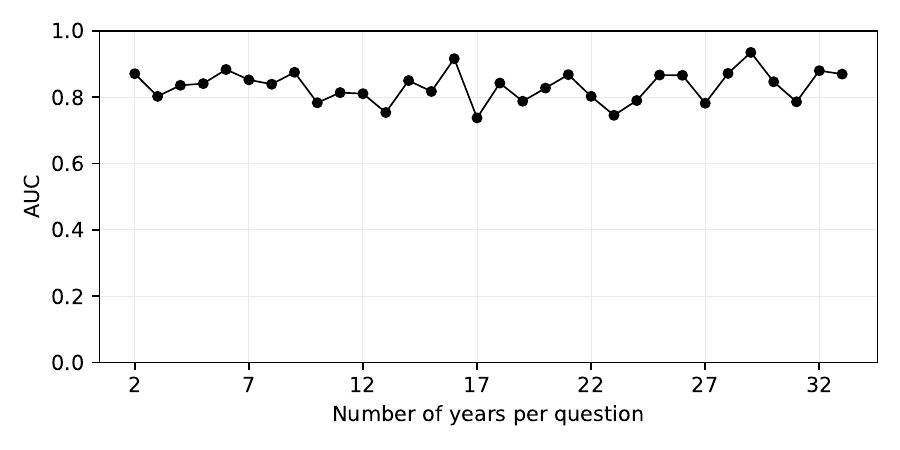}
  \end{center}
  \caption{\textbf{Number of years per survey question in training data and AUC in retrodiction.} X-axis indicates the number of years per survey question in training data. For instance, the value of 2 indicates the case when a survey question is asked only twice. Y-axis indicates the AUC values for year-level missing opinions.}
  \label{fig:fig_figure_years_vs_auc}
\end{figure*}

%% file: Figures/figure_a_individualauc_compare_block.tex
\begin{figure}[!t]
  \begin{center}
      \centering
     \includegraphics[width=1\columnwidth]{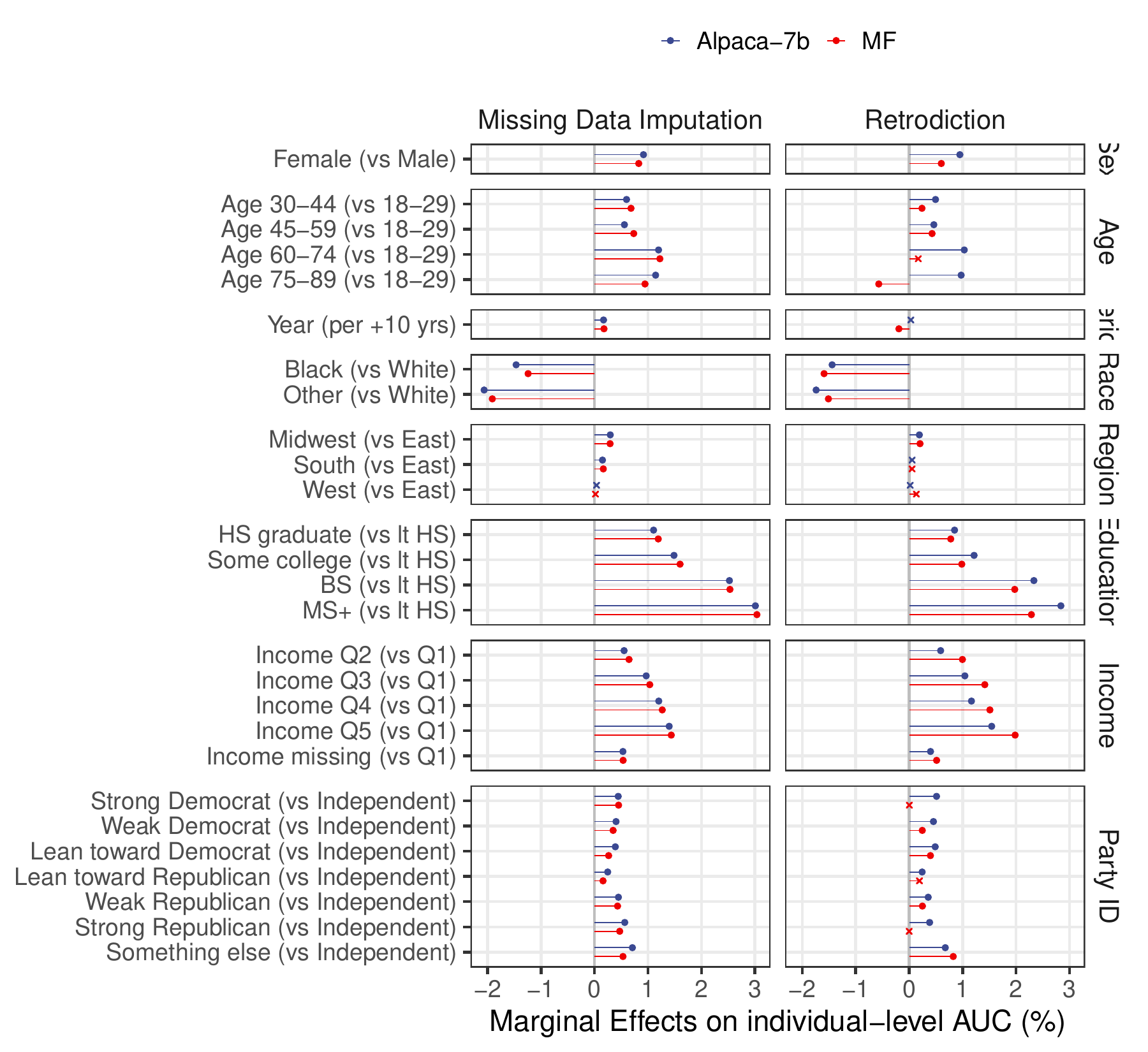}
  \end{center}
  \caption{\textbf{Comparison of results from OLS regression models predicting individual-level AUC across two different types of missing data imputation between Alpaca and matrix factorization models.} A higher AUC value indicates greater model accuracy for individuals. Here, each dot represents the expected difference of AUC (i.e., average marginal effects) against the reference group within each subgroup. Here, a filled dot refers to a statistically significant difference, and an X refers to a statistically insignificant difference based on robust standard errors (p \textless{} 0.05).}
  \label{fig:alpaca_vs_mf_individualauc}
\end{figure}

%% file: Figures/figure_a_ideology_corr_time_block.tex
\begin{figure}[htp]
    \centering
    \includegraphics[width=\textwidth]{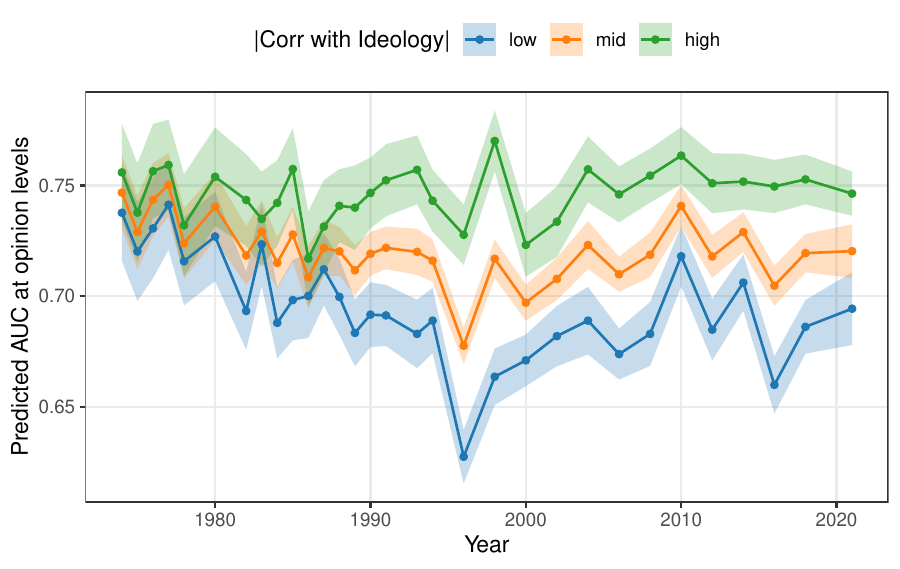}
    \caption{\textbf{Model performance over time by ideological correlation level.} This figure illustrates how the predicted AUC varies across years for opinions that differ in their correlation with political ideology. Predicted AUC values are obtained from a regression of per-(variable, year) AUC on the z-scored absolute correlation with political ideology ($|$Corr with Ideology$|$) and survey-year dummies, $\text{auc} \sim |\text{Corr with Ideology}| \times \text{factor}(\text{year})$. The three lines correspond to opinions at low, medium, and high levels of $|$Corr with Ideology$|$, defined as $z = -1$, $0$, and $+1$ on the z-scored scale (i.e., one standard deviation below the mean, the mean, and one standard deviation above the mean of $|$Corr with Ideology$|$ across all (variable, year) cells). Shaded bands are 95\% confidence intervals from the regression.}
    \label{fig:ideology_corr_time}
\end{figure}

%% file: Figures/figure_ft_effect_block.tex
\begin{figure}[ht]
  \begin{center}
      \centering
     \includegraphics[width=0.90\columnwidth]{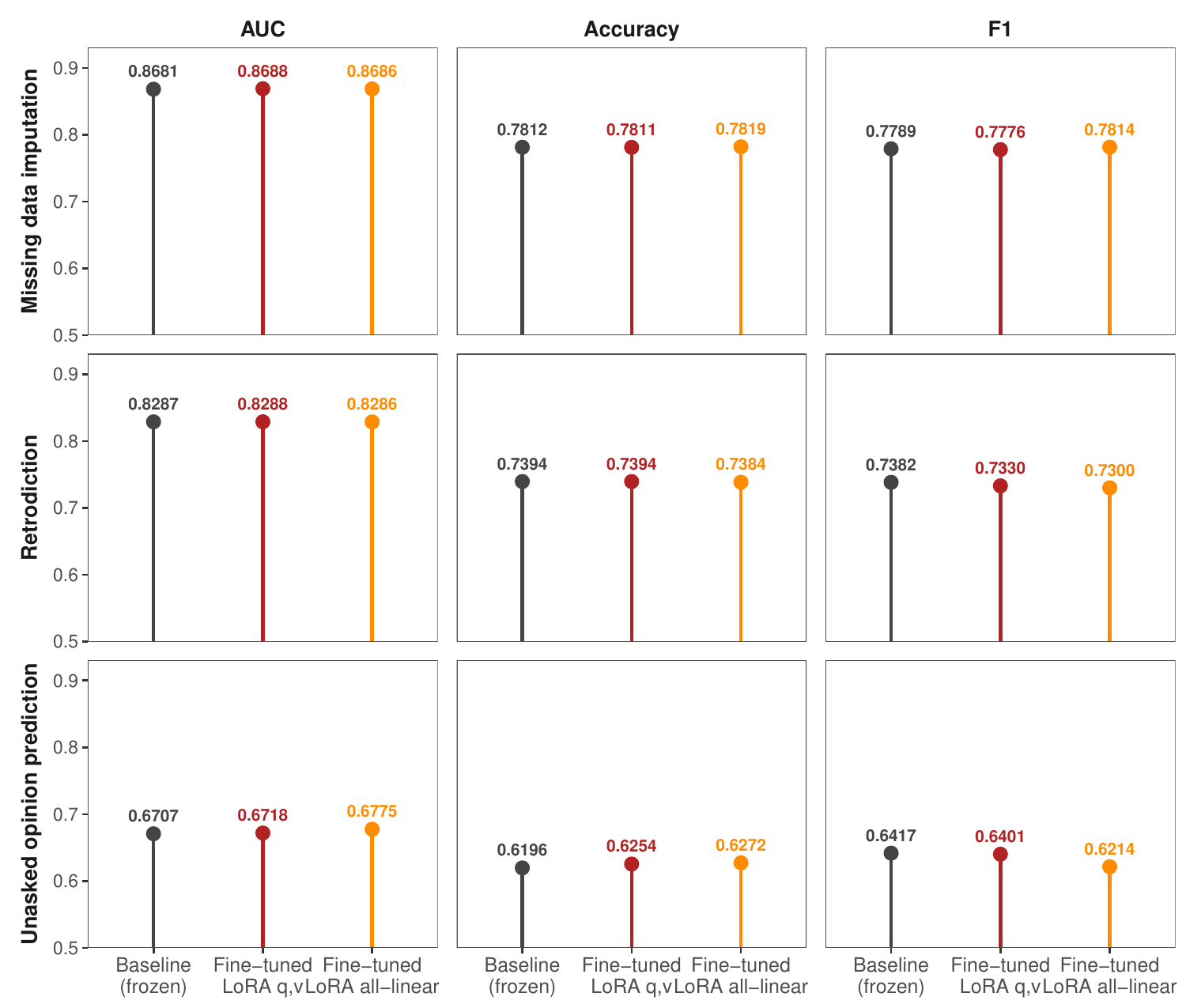}
  \caption{\textbf{Effect of fine-tuning LLM parameters on downstream prediction.} Rows correspond to the three prediction tasks (missing data imputation, retrodiction, unasked opinion prediction); columns correspond to AUC, accuracy, and F-1. Each panel compares the fully frozen baseline (gray) against two LoRA fine-tuning configurations: \emph{LoRA $q,v$}, adapting the query and value projection matrices of self-attention blocks (red), and \emph{LoRA all-linear}, adapting every linear module within the transformer blocks (orange). For each task we hold the fold-0 evaluation cells fixed and identical to the cells used by the main analyses ($n = 1{,}451{,}906$ for missing data imputation; $n = 1{,}428{,}512$ for retrodiction; $n = 1{,}616{,}175$ for unasked opinion prediction). Both LoRA configurations yield marginally higher (or near-identical) AUC and accuracy than the fully frozen baseline, but slightly lower F-1 scores, with the largest decline in F-1 ($-0.020$) appearing in unasked opinion prediction under \emph{LoRA all-linear}.}
  \label{fig:ft-effect}
  \end{center}
\end{figure}